\def\isarxiv{1} %%% for icml submission version, we comment this line

\ifdefined\isarxiv
\documentclass[11pt]{article}

\usepackage[numbers]{natbib}

\else
\documentclass[twoside]{article}
\usepackage[authoryear]{natbib}
\usepackage{aistats2024}
\fi

\usepackage{amsmath}
\usepackage{amsthm}
\usepackage{amssymb}
\usepackage{algorithm}
\usepackage{subfig}
\usepackage{algpseudocode}
\usepackage{graphicx}
\usepackage{grffile}
\usepackage{wrapfig,epsfig}
\usepackage{url}
\usepackage{xcolor}
\usepackage{epstopdf}

\usepackage{bbm}
\usepackage{dsfont}

\allowdisplaybreaks

\ifdefined\isarxiv

\usepackage{tikz}
\usepackage{hyperref}  %%% arxiv don't allow this.
\hypersetup{colorlinks=true,citecolor=blue,linkcolor=blue} %%% Zhao : maybe we should comment this in submission.
\usetikzlibrary{arrows}
\usepackage[margin=1in]{geometry}

\else

\usepackage{microtype}
\usepackage{hyperref}
\definecolor{mydarkblue}{rgb}{0,0.08,0.45}
\hypersetup{colorlinks=true, citecolor=mydarkblue,linkcolor=mydarkblue}

\fi
\graphicspath{{./figs/}}

\newtheorem{theorem}{Theorem}[section]
\newtheorem{lemma}[theorem]{Lemma}
\newtheorem{definition}[theorem]{Definition}

\newtheorem{fact}[theorem]{Fact}

\newtheorem{claim}[theorem]{Claim}

\newcommand{\R}{\mathbb{R}}

\DeclareMathOperator*{\E}{{\mathbb{E}}}

\DeclareMathOperator{\A}{\mathsf{A}}

\DeclareMathOperator{\poly}{poly}

\DeclareMathOperator{\diag}{diag}

\DeclareMathOperator{\vect}{vec}

\DeclareMathOperator{\lin}{lin}

\makeatletter
\newcommand*{\RN}[1]{\expandafter\@slowromancap\romannumeral #1@}
\makeatother
\usepackage{lineno}

\begin{document}

\ifdefined\isarxiv

\date{}

\title{Why Softmax Attention Outperforms Linear Attention}
\author{
Yichuan Deng\thanks{\texttt{ycdeng@cs.washington.edu}. The University of Washington.} 
\and
Zhao Song\thanks{\texttt{zsong@adobe.com}. Adobe Research.}
\and 
Kaijun Yuan\thanks{\texttt{yybrs0929@gmail.com}. University of Bologna, Italy. }
\and 
Tianyi Zhou\thanks{\texttt{tzhou029@usc.edu}. University of Southern California.}
}

\else

\twocolumn[
\aistatstitle{Superiority of Softmax: Unveiling the Performance Edge Over Linear Attention}
\aistatsauthor{Yichuan Deng \And Zhao Song \And Tianyi Zhou}
\aistatsaddress{Adobe Research \And ??? \And ???}
]

\fi

\ifdefined\isarxiv
\begin{titlepage}
  \maketitle
  \begin{abstract}
Large transformer models have achieved state-of-the-art results in numerous natural language processing tasks. Among the pivotal components of the transformer architecture, the attention mechanism plays a crucial role in capturing token interactions within sequences through the utilization of softmax function.

Conversely, linear attention presents a more computationally efficient alternative by approximating the softmax operation with linear complexity. However, it exhibits substantial performance degradation when compared to the traditional softmax attention mechanism.

In this paper, we  bridge the gap in our theoretical understanding of the reasons behind the practical performance gap between softmax and linear attention. By conducting a comprehensive comparative analysis of these two attention mechanisms, we shed light on the underlying reasons for why softmax attention outperforms linear attention in most scenarios.

  \end{abstract}
  \thispagestyle{empty}
\end{titlepage}

{
%\hypersetup{linkcolor=black}
%\tableofcontents
}
\newpage

\else

\begin{abstract}
Large transformer models have achieved state-of-the-art results in numerous natural language processing tasks. Among the pivotal components of the transformer architecture, the attention mechanism plays a crucial role in capturing token interactions within sequences through the utilization of softmax function.

Conversely, linear attention presents a more computationally efficient alternative by approximating the softmax operation with linear complexity. However, it exhibits substantial performance degradation when compared to the traditional softmax attention mechanism.

In this paper, we  bridge the gap in our theoretical understanding of the reasons behind the practical performance gap between softmax and linear attention. By conducting a comprehensive comparative analysis of these two attention mechanisms, we shed light on the underlying reasons for why softmax attention outperforms linear attention in most scenarios.
\end{abstract}

\fi

\section{Introduction}

Large language models (LLMs) like Transformer \cite{vsp+17}, BERT \cite{dclt18}, RoBERTa \cite{log+19}, XLNet \cite{ydy+19}, GPT-3 \cite{bmr+20}, OPT \cite{zrg+22}, PaLM \cite{cnd+22}, Llama \cite{tli+23}, Llama2 \cite{tms+23}, Adobe firefly \cite{adobe_firefly}, and BARD \cite{m23} 
have proven to be efficient methods for addressing complex natural language tasks such as  content creation, summarization, and dialogue systems \cite{tdh+22,ycri22,wtb+22}.
The utilization of attention mechanisms has revolutionized the landscape of computer vision and natural language processing, significantly enhancing network performance. However, a critical challenge lies in the escalating memory and computational demands associated with the prevalent dot-product attention mechanism, particularly as the increase of the input length. 
The quadratic computational complexity of this attention mechanism with respect to the number of tokens has historically hindered its applicability to processing lengthy sequences.

Recent efforts in the research community have been dedicated to developing efficient attention architectures, aiming to mitigate computation complexity and memory usage. Linear attention is one of the proposed method that has been widely studied \cite{lsdz20,kvpf20, szz+21, zwk22, lcz+23}.
In \cite{lsdz20}, they propose a Linear Attention Mechanism which is approximate to dot-product attention with much less memory and computational costs based on first-order approximation of Taylor
expansion.

In practical applications, softmax attention consistently demonstrates superior performance when compared to linear attention. We hypothesize that for Transformer,
\begin{center}
\textit{there exist some datasets that can only be effectively classified using softmax attention, while linear attention proves inadequate.}
\end{center}

In this paper, we delve into a comprehensive comparison between softmax attention and linear attention, examining their attributes from both experimental and theoretical perspectives.

Our construction is inspired by the hardness proofs in fine-grained complexity \cite{bis17,c18,cjw19,acss20,as23,as23_tensor,als+23,jx23}. Let $K \in \R^{d \times d}, Q \in \R^{d \times d}, V \in \R^{d \times d}$ denote the learned key matrix, query matrix and the value matrix respectively in the attention layer. Let $A_1,A_2$ denote the input sequences. The formulation of the softmax cross attention is defined as:
\begin{align}\label{eq:attention}
    \mathsf{Att}(X,Y) = D(X)^{-1} \exp(A_1 X A_2^\top) A_2 Y 
\end{align}
where $X \in \R^{d \times d}$ denotes the combined parameter $X:=QK^\top $,  $Y \in \R^{d \times d} := V$ and $D(X):= \diag(\exp(A_1 X A_2^\top ){\bf 1}_n) \in \R^{d \times d}$ denotes the softmax normalization.
For linear cross attention, we replace the $\exp$ with $\lin$, which stands for linear. Note that, self attention is a special case for the cross attention, i.e., $A_2 = A_1$.
Let $F_{\exp}$ represent the neural network employing softmax attention and ReLU activation, while $F_{\lin}$ denotes the neural network utilizing linear attention. Next, we state our main results.

% Consider two sets of datasets, $\D_0$ and $\D_1$, where $\D_1$ contains one outlier. In this context, we provide the following assertions, which implies the superiority of softmax attention. 
% \begin{itemize}
%     \item For each datum $A$ drawn from $\D_0$ and $\D_1$, it holds that $F_{\lin}=0$ with a probability of at least $1 - \delta/\poly(n)$, where $\delta \in (0,0.1)$.
%     \item For data drawn from $\D_0$, the probability that $F_{\exp}$ yields an output of $0$ is at least $1 - \delta/\poly(n)$. Conversely, for data from $\D_1$, the probability that $F_{\exp}$ produces an output greater than $0$ is at least $1 - \delta/\poly(n)$.
% \end{itemize}
% for simplicity, we let QK = I or {\bf 1}
% We first present a simplified example of the attention formulation to show our results. The comprehensive proof of our results, encompassing both self-attention and cross-attention scenarios, is ???.
% We also provide the experiments to validate the robustness of our theoretical findings across various datasets.

\subsection{Our Result}

For the self-attention, we need $A_1 = A_2$ in Eq.\eqref{eq:attention}.
\begin{theorem}[Self-attention, informal of Section \ref{sec:app_self_attention}]
There exists two (self-attention) datasets ${\cal D}_0 \subset \R^{n \times d}$ and ${\cal D}_1 \subset \R^{n \times d}$. There exists two four-layer neural networks: $F_{\exp} : \R^{n \times d} \rightarrow \R$ which uses softmax units and ReLU units, and $F_{\lin} :\R^{n \times d} \rightarrow \R$ which uses linear attention units and ReLU units such that with high probability (the randomness is over the weights of network)
\begin{itemize}
    \item $F_{\exp}$ can distinguish ${\cal D}_0$ and ${\cal D}_1$
    \item $F_{\lin}$ cannot distinguish ${\cal D}_0$ and ${\cal D}_1$
\end{itemize}
\end{theorem}

For cross attention, we need $A_1 \neq A_2$ in Eq.~\eqref{eq:attention}, and thus the function of neural network in fact takes two $n \times d$ matrices as inputs.
\begin{theorem}[Cross-attention, informal of Section \ref{sec:app_cross_attention}]
There exists two (cross-attention) datasets ${\cal D}_0 \subset \R^{2 n \times d}$ and ${\cal D}_1 \subset \R^{2 n \times d}$. There exists two four-layer neural networks: $F_{\exp} : \R^{n \times d} \times \R^{n \times d} \rightarrow \R$ which uses softmax units and ReLU units, and $F_{\lin} :\R^{n \times d} \times \R^{n \times d} \rightarrow \R$ which uses linear attention units and ReLU units such that with high probability (the randomness is over the weights of network)
\begin{itemize}
    \item $F_{\exp}$ can distinguish ${\cal D}_0$ and ${\cal D}_1$
    \item $F_{\lin}$ cannot distinguish ${\cal D}_0$ and ${\cal D}_1$
\end{itemize}
\end{theorem} %%% Section 1. Introduction

\section{Related Work}

The efficacy of Transformer architectures in particular relies entirely on a self-attention mechanism \cite{ptdu16,lfs+17} to compute a series of context-informed vector-space representations of the symbols in its input and output, which are then used to predict distributions over subsequent symbols as the model predicts the output sequence symbol-by-symbol.
Concrete evidence from studies \citep{tdp19, vb19, hl19, b22} underscores the pivotal role of attention with multilayer perceptron in transmitting crucial information, facilitating diverse probing assignments.

Contemporary investigations have delved deep into Transformers' capabilities. Topics explored range from Turing completeness \citep{pmb19,bpg20}, functional representation \citep{ybr+20, cdw+21}, to the representation of formal languages \citep{bag20, egz20, yppn21} and the mastering of abstract algebraic operations \citep{zbb+22}.
Some research alludes to the potential of Transformers to serve as universal adaptors for sequence-based tasks and even mimic the capabilities of Turing machines \citep{pmb19, bpg20}. 

However, the quadratic computational complexity of this attention mechanism with
respect to the number of tokens has historically hindered its applicability to processing lengthy sequences. 
\cite{dcl+21} introduced the Pixelated Butterfly model, a strategy employing a consistent sparsity pattern to speed up Transformer training. 
 Performer \cite{cld+20} is an example of the low-rank variant, which uses kernelization to speed up the computation.
There are also work that approximate attention computations during the inference phase, provided the precision is adequately maintained. 
\cite{lwd+23} highlights the phenomenon of contextual sparsity in LLM and its predictability. They leveraged this insight to accelerate LLM inference without compromising on output quality.  

Numerous investigations, encompassed by references such as \cite{cgrs19,kkl20,wlk+20,dkod20,kvpf20,cdw+21,cdl+22}, have illuminated various facets of this domain. Subsequent research, represented by works like \cite{zhdk23,as23,bsz23,dms23,kmz23,as23_tensor,hjk+23,ag23,mda22}, have delved deeply into the computation of attention matrices, highlighting its intricacies and advocating for optimized algorithms.

Furthermore, significant strides have been made in understanding the power of attention mechanisms within Transformers, as illustrated by studies \citep{dgv+18, vbc20, zkv+20, egkz21, szks21, wcm21, dsx23, dlms23}. The work of \cite{zpga23} underscored the potential of mid-scale masked language models to recognize syntactic components, presenting possibilities for partial parse tree reconstructions. This innovative concept, postulated by \cite{zpga23}, facilitated \cite{dgs23} in their exploration of tensor cycle rank approximation challenges. \cite{gms23} subsequently turned their lens towards the exponential regression in the context of the over-parameterized neural tangent kernel. While \cite{lsz23} engaged in evaluating a regularized version of the exponential regression, a notable omission was the normalization factor. In a distinct approach, \cite{dls23} emphasized softmax regression, encompassing this normalization factor, thereby differentiating their work from prior exponential regression research \cite{gms23,lsz23}.

{\bf Roadmap} We first provide a toy example, which simplify the attention formulation, before we propose our main results. In Section \ref{sec:pre} we provide some tools that used in our proof and the formulation of our toy model.
In Section \ref{sec:property_toy}, we provide some properties of the dataset that use in the proof of the toy example.
In Section \ref{sec:toy_result}, we provide the theoritical analysis of the performance of different models in binary classification task.
In Section \ref{sec:main_results}, we proposed our main results for both self-attention and cross-attention.
In Section \ref{sec:numerical_exp}, we show some experiments that shows the robustness of our theoretical results.

\section{Preliminary}\label{sec:pre}
Here in this section, we provide some preliminaries to be used. 

{\bf Notations.}

For a positive integer $n$, the set $\{1,2,\cdots,n\}$ is denoted by $[n]$.
Given vectors $u$ and $v$, their inner product is represented as $\langle u, v\rangle$.
For any $u \in \R^n$, the vector with entries $\exp (x)_i = \exp (x_i)$ is given by $\exp (u) \in \R^n$.
A vector of length $n$ with all entries being one is represented as ${\bf 1}n$.
Considering a matrix $A \in \R^{n \times d}$, its $i$-th column is referred to as $A{*,i}$ for every $i \in [d]$.
The element-wise product of two vectors $u$ and $v$ is denoted by $u \circ v$, where the $i$-th entry is $u_i v_i$.
We use $\E[]$ to denote expectation. We use $\Pr[]$ to denote the probability.

\subsection{Probability Tools}

\begin{lemma}[Hoeffding bound \cite{h63}]\label{lem:hoeffding_bound}
Let $X_1, \cdots, X_n$ denote $n$ independent bounded variables in $[a_i,b_i]$. Let $X = \sum_{i=1}^n X_i$, then we have
\begin{align*}
    \Pr[ | X - \E[X] | \geq t ] \leq 2 \exp \Big( - \frac{2t^2}{ \sum_{i=1}^n (b_i - a_i)^2 } \Big). 
\end{align*}
\end{lemma}

\subsection{Definitions of Functions}
\begin{definition}[Linear functions]
We define $u_{\lin} : \R^{n \times d} \rightarrow \R^n$
%\begin{align*}
$
    u_{\lin}(A;x): = A x
$. 
%\end{align*}
We define $\alpha_{\lin}: \R^{n \times d} \rightarrow \R$
%\begin{align*}
$
    \alpha_{\lin}(A;x) := \langle u_{\lin} (A,x) , {\bf 1}_n \rangle
$. 
%\end{align*}
We define $f_{\lin} : \R^{n \times d} \rightarrow \R^n$
%\begin{align*}
$
    f_{\lin}(A;x):= \alpha_{\lin}(A,x)^{-1} u_{\lin}(A,x)
$.
%\end{align*}
\end{definition}

\begin{definition}[Softmax functions]
We define $u_{\exp}: \R^{n \times d} \rightarrow \R^n$
%\begin{align*}
$
    u_{\exp}(A;x):= \exp(Ax).
$. 
%\end{align*}
We define 
%\begin{align*}
$
    \alpha_{\exp}(A,x):= \langle u_{\exp}(A,x) , {\bf 1}_n \rangle
%\end{align*}
$. 
We define $f_{\exp}(x)$ as follows 
%\begin{align*}
$
    f_{\exp}(A,x):= \alpha_{\exp}(A,x)^{-1} u_{\exp}(A,x)
%\end{align*} 
$.
\end{definition}

\begin{definition}[ReLU]
We define $\phi(z) := \max\{z,0\}$.

More generally, for parameter $\tau$, we define $\phi_{\tau}(z): = \max\{ z - \tau, 0\}$.
\end{definition}

\begin{definition}
Let $\tau > 0$ denote a parameter.
We define a three-layer neural network (with softmax attention and ReLU activation) $F_{\exp}: \R^{n \times d} \rightarrow \R$
%\begin{align*}
$
    F_{\exp}(A;x,y):= \phi( \sum_{j=1}^m \phi_{\tau}( \langle f_{\exp}(A,x) , y_j \rangle  )   )
$.
%\end{align*}
\end{definition}

\begin{figure}[!ht]
    \centering
    \includegraphics[width = 0.4\textwidth]{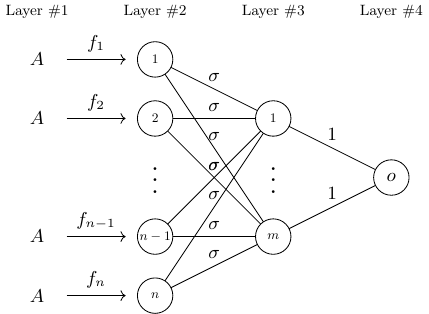}
    \caption{Visualization of our neural network in Section \ref{sec:property_toy}. Here $m = O(\log n)$.
    }
    \label{fig:network}
\end{figure}
\begin{figure*}[!ht]
    \centering
    \includegraphics[width = 0.9\textwidth]{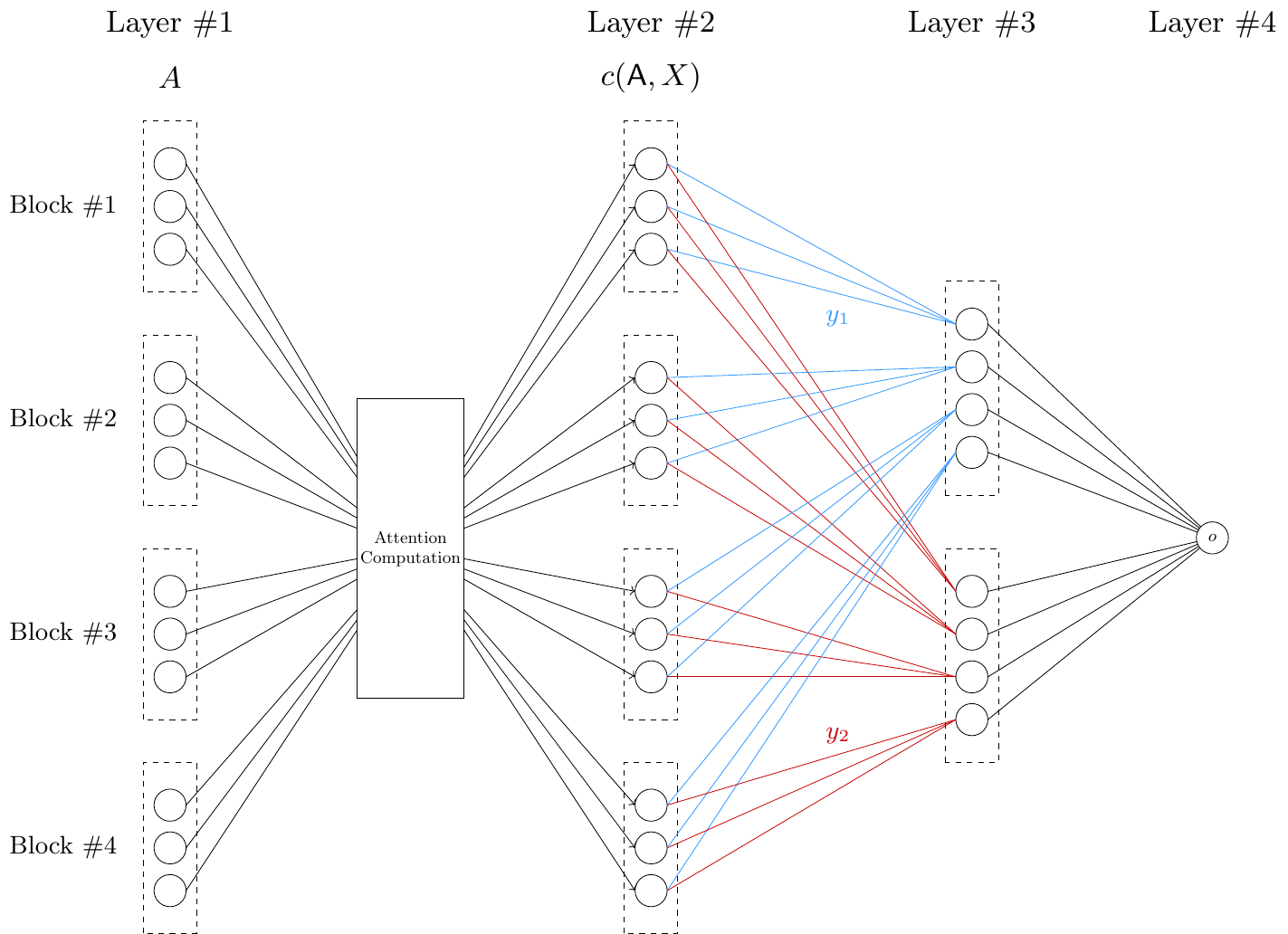}
    \caption{This is structure for self-attention. Let $A = A_1 = A_2 = A_3 \in \R^{n \times d}$. Let $\mathsf{A} = A_1 \otimes A_2$. Let $n=4$, $d=3$, $m=2$. Let $y_1 \in \R^d$. Let $c(\A,x) \in \R^{n \times d}$ which is $\R^{4 \times 3}$.  Let $c(\A,x)_{1} \in \R^3$ denote the first block in layer 2. For the first node in first block in layer 3, the computation is based on ReLU of $\langle c(\A,X)_1, y_1 \rangle$.
    }
    \label{fig:attention_network}
\end{figure*}

\begin{definition}
We define a four-layer neural network(with linear attention and ReLU activation) $F_{\lin}: \R^{n \times d} \rightarrow \R$
%\begin{align*}
$
    F_{\exp}(A;x,y):= \phi( \sum_{j=1}^m \phi_{\tau}( \langle f_{\lin}(A,x) , y_j \rangle  )   )
$
%\end{align*}
\end{definition}

\begin{definition}\label{def:y}
Let $C>1$ denote some constant. Let $m = C\log (n/\delta)$. 
Let $y \in \R^{n \times m}$, for each $j \in [m]$, we use $y_j$ to denote the $j$-th column of $y$. For each entry in $y_j$ we sampled it from Rademacher random variable distribution.
\end{definition}
The settings of the above neural network are natural in deep learning theory \cite{zsj+17,ll18,dzps19,azls19a,azls19b,sy19,szz21,bpsw21,als+23,mosw22}.  

\subsection{Definition of Datasets for Binary Classification}
\begin{definition}[Binary classification]\label{def:binary_classification}
Given two sets of datasets ${\cal D}_0$ and ${\cal D}_1$.
\begin{itemize}
    \item For each $A \in \R^{n \times d}$ from ${\cal D}_0$, we assume that $(Ax)_i \in [\log n, 1.4 \log n]$ for all $i \in [n]$
    \item For each $A \in \R^{n \times d}$ from ${\cal D}_1$, we assume that there is one index $j \in [n]$ such that $(Ax)_j = 4 \log n$ and for all $i \in [n] \backslash \{j\}$, we have $(Ax)_i \in [\log n, 1.4 \log n]$
\end{itemize}
\end{definition}

\section{Property of Dataset}
\label{sec:property_toy}

\begin{figure*}[!ht]
    \centering
    \subfloat[Ratio of Success with respect to $n$]
    {\includegraphics[width=0.48\textwidth]{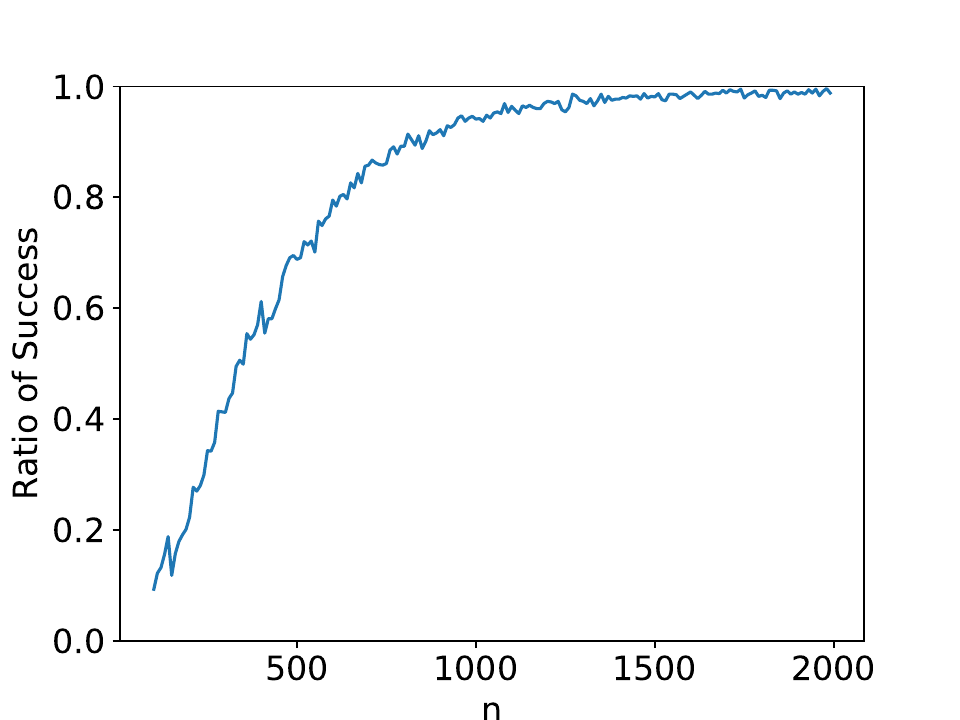}}\label{fig:toy_n}
    % \hspace{5mm}
    \subfloat[Ratio of Success with respect to $m$]  {\includegraphics[width=0.48\textwidth]{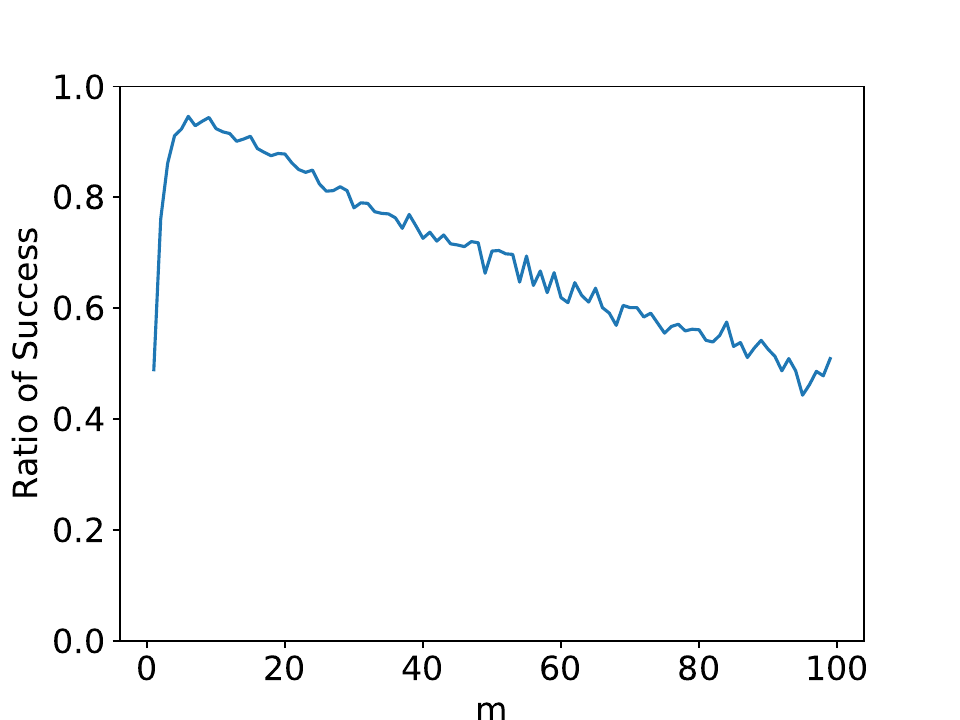}} \label{fig:toy_m}
    \caption{Figure for the softmax regression model } 
    \label{fig:toy}
\end{figure*}

\begin{lemma}\label{lem:property_of_dataset_0}
Let $\sigma \sim \{-1,+1\}^n$ denote a random sign vector. Let $C > 1$ be a sufficiently large constant. Let $\delta \in (0,0.1)$. 
Then, for each $A \in {\cal D}_0$, we have
%\begin{itemize}
%    \item 
    Part 1. 
   %$ \begin{align*}
   $
        ~ \Pr_{\sigma}[ | \sum_{i=1}^n f_{\lin}(A,x)_i \sigma_i | \leq C \frac{ \sqrt{ \log(n/\delta) }} { \sqrt{n} } ] 
        \geq ~ 1- \delta/ \poly(n)
    $
   % \end{align*}
Part 2. 
   $
     ~ \Pr_{\sigma}[ | \sum_{i=1}^n f_{\exp}(A,x)_i \sigma_i | \leq C \frac{ \sqrt{ \log(n/\delta)} }{  n^{0.1} } ] 
    \geq  ~ 1- \delta/ \poly(n)
    $
%\end{itemize}
\end{lemma}
\begin{proof}

{\bf Proof of Part 1.}
Note that we can show that
\begin{align*}
    | f_{\lin} (A,x)_i | = & ~ \alpha_{\lin}(A,x)^{-1} | u_{\lin}(A,x)_i | \\
    \leq & ~ (n \log n)^{-1} \cdot ( 1.4 \log n ) \\
    \leq & ~ \frac{2}{n}
\end{align*}
where the first step follows from the definition of $\alpha$, the second step follows from Definition \ref{def:binary_classification} and the last step follows from simple algebra.

Thus, applying Hoeffding inequality, we can get with probability $1-\delta /\poly(n)$, 
\begin{align*}
 | \sum_{i=1}^n f_{\lin}(A,x)_i \sigma_i | 
 \leq ~ C \frac{ \sqrt{ n \log(n/\delta) }} {n} 
 =  ~ C \frac{ \sqrt{ \log(n/\delta) } }{ \sqrt{n} }
\end{align*}

{\bf Proof of Part 2.}
For all $i \in [n]$, we know that
\begin{align*}
    | f_{\exp} (A,x)_i | = & ~ \alpha_{\exp}(A,x)^{-1} \cdot | u_{\exp}(A,x)_i | \\
    \leq & ~ (n \cdot n )^{-1} \cdot ( n^{1.4} ) \\
    = & ~ \frac{1}{n^{0.6}}
\end{align*}

 Thus, applying Hoeffding inequality, we can get with probability $1-\delta /\poly(n)$, 
\begin{align*}
 | \sum_{i=1}^n f_{\exp}(A,x)_i \sigma_i | 
 \leq & ~ C \frac{ \sqrt{ n \log(n/\delta) }} {n^{0.6}} \\
 \leq & ~ C \frac{ \sqrt{\log(n/\delta)
} }{ n^{0.1} }
\end{align*}

\end{proof}

\begin{lemma}\label{lem:property_of_dataset_1}
Let $\sigma \sim \{-1,+1\}^n$ denote a random sign vector. Let $C > 1$ be a sufficiently large constant. Let $\delta \in (0,0.1)$. 
Then, for each $A \in {\cal D}_1$, we have
\begin{itemize}
    \item Part 1.  $\Pr_{\sigma}[ | \sum_{i=1}^n f_{\lin}(A,x)_i \sigma_i | \leq C \sqrt{ \log(n/\delta) / n} ] \geq 1- \delta/ \poly(n)$
    \item Part 2. $\Pr_{\sigma}[ | \sum_{i=1}^n f_{\exp}(A,x)_i \sigma_i | \geq 1/4 ] \geq 1$
    \item Part 3. $\Pr[ \sum_{i=1}^n f_{\exp}(A,x)_i \sigma_i > 0 ] = 1/2$ and $\Pr[ \sum_{i=1}^n f_{\exp}(A,x)_i \sigma_i < 0 ] = 1/2$
\end{itemize}
\end{lemma}
\begin{proof}

{\bf Proof of Part 1.}
we can show that
\begin{align*}
    | f_{\lin} (A,x)_i | = & ~ \alpha_{\lin}(A,x)^{-1} | u(A,x)_i | \\
    \leq & ~ ( (n-1) \log n + 4 \log n)^{-1} \cdot ( 4 \log n ) \\
    = & ~ \frac{4}{n+3} \\
    \leq & ~ \frac{4}{n}
\end{align*}
where the first step follows from the definition of $f_{\lin}$, the second step follows from Definition \ref{def:binary_classification}, the third step follows from simple algebra and the last step follows from simple algebra.

Applying Hoeffding inequality again, we finish the proof.

{\bf Proof of Part 2.}

There is one index $j \in [n]$, we know that
\begin{align*}
    | f_{\exp} (A,x)_j | = & ~ \alpha_{\exp}(A,x)^{-1} \cdot | u_{\exp}(A,x)_j | \\
    \geq & ~ (n^{1.4} \cdot (n-1) + n^4 )^{-1} \cdot ( n^4 ) \\
    \geq & ~ \frac{1}{2}
\end{align*}

For all the $i \in [n] \backslash \{j\}$, we know that
\begin{align*}
    | f_{\exp} (A,x)_i | = & ~ \alpha_{\exp}(A,x)^{-1} \cdot | u_{\exp}(A,x)_i | \\
    \leq & ~ (n \cdot (n-1) + n^4 )^{-1} \cdot ( n^{1.4} ) \\
    \leq & ~ \frac{1}{n^{2.6}}
\end{align*}
Thus, it is obvious that
\begin{align*}
 | \sum_{i=1}^n f_{\exp}(A,x)_i \sigma_i | \geq \frac{1}{2} - (n-1) \cdot \frac{1}{n^{2.6}} \geq 1/4
\end{align*}
where the last step follows from $n \geq 4$.

{\bf Proof of Part 3.}
The sign is completely decided by $\sigma_j$, thus it has $1/2$ chance to be positive and $1/2$ chance to be negative.
\end{proof}

\section{Binary Classification}\label{sec:toy_result}
In this section, we provide an overview of theoretical analysis of the performance of different models in binary classification task. 
\subsection{Softmax Attention}
\begin{lemma}
For each data $A$ from ${\cal D}_0$, $F_{\exp}(A) = 0$ with probability at least $1-\delta/\poly(n)$.
\end{lemma}
\begin{proof}
Note that all the $y_l$ are independent, for each $l\in [m]$, we call Part 2 of Lemma~\ref{lem:property_of_dataset_0}, we can show that
$
  \phi_{\tau}( \langle f_{\exp}(A,x) , y_l \rangle   ) = 0
$. 

Since $m \leq \delta/\poly(n)$, we are allowed to take union bound over all $l \in [m]$. Thus, we have 
$
F_{\lin}(A,x) = \phi( \sum_{l=0}^m   \phi_{\tau}( \langle f_{\exp}(A,x) , y_l \rangle   ) ) = 0.
$
\end{proof}

\begin{lemma}
For each data $A$ from ${\cal D}_1$, $F_{\exp}(A) > 0 $ with probability at least $1-\delta/\poly(n)$.
\end{lemma}
\begin{proof}
By Part 2 and 3 of Lemma~\ref{lem:property_of_dataset_1}, we have 
$
  \phi_{\tau}( \langle f_{\exp}(A,x) , y_l \rangle   ) > 1/4
$
with probability $1/2$.

Since all $l \in [m]$ are independent, thus there exists one $l \in [m]$ such that \begin{align*}
  \phi_{\tau}( \langle f_{\exp}(A,x) , y_l \rangle   ) > 1/4
\end{align*}
the probability is $1-(1/2)^m \geq 1-\delta/\poly(n)$.

Thus, with probability $1-\delta/\poly(n)$, we have
\begin{align*}
F_{\exp}(A,x) = \phi( \sum_{l=0}^m   \phi_{\tau}( \langle f_{\exp}(A,x) , y_l \rangle   ) ) > 0
\end{align*}
holds. 
\end{proof}

\subsection{Linear Attention}

\begin{lemma}
For each data $A$ from ${\cal D}_0$, $F_{\lin}(A) = 0 $ with probability at least $1-\delta/\poly(n)$.
\end{lemma}
\begin{proof}
Note that all the $y_l$ are independent, for each $l\in [m]$, we call Part 1 of Lemma~\ref{lem:property_of_dataset_0}, we can show that
%\begin{align*}
$
  \phi_{\tau}( \langle f_{\lin}(A,x) , y_l \rangle   ) = 0
  $.
%\end{align*}

Since $m \leq \delta/\poly(n)$, we are allowed to take union bound over all $l \in [m]$. 
Thus, we have 
%\begin{align*}
$
F_{\lin}(A,x) = \phi( \sum_{l=0}^m   \phi_{\tau}( \langle f_{\lin}(A,x) , y_l \rangle   ) ) = 0
$. %\end{align*}
\end{proof}

\begin{lemma}
For each data $A$ from ${\cal D}_1$, $F_{\lin}(A) = 0 $ with probability at least $1-\delta/\poly(n)$.
\end{lemma}
\begin{proof}
Note that all the $y_l$ are independent, for each $l\in [m]$, we call Part 1 of Lemma~\ref{lem:property_of_dataset_1}, we can show that
%\begin{align*}
$
  \phi_{\tau}( \langle f_{\lin}(A,x) , y_l \rangle   ) = 0
$.
%\end{align*}

Since $m \leq \delta/\poly(n)$, we are allowed to take union bound over all $l \in [m]$.

Thus, we have 
%\begin{align*}
$
F_{\lin}(A,x) = \phi( \sum_{l=0}^m   \phi_{\tau}( \langle f_{\lin}(A,x) , y_l \rangle   ) ) = 0
%\end{align*}
$
\end{proof}

\section{Main Results}\label{sec:main_results}

%\subsection{Self-attention}
We provide more details for self-attention result, for details of cross-attention, we refer the readers to appendix.

\begin{definition}[Self-Attention dataset distribution, informal version of Definition \ref{def:dataset_self_attention}]\label{def:dataset_self_attention_informal}
We define $a_0 \in (0, 0.1)$.  We denote $a_1 \geq 0.7$. Let
$b + c = 1$ for $b \geq 0.1 , c \geq 0.1$. Assume $n = (d-2) t$ where $t$ is a positive integer. 

Given two sets of datasets ${\cal D}_0$ and ${\cal D}_1$. 
For each $A_1 \in {\cal D}_0$, we have the first column of $A_1$ is $e_{j_3} \cdot a_0 \sqrt{\log n}$ for some $j_3 \in [n]$.
From the second column to the $(d-1)$-th columns of $A_1$ are $\begin{bmatrix} I_{d-2} \\ I_{d-2} \\ \vdots \\ I_{d-2} \end{bmatrix} \cdot b \sqrt{\log n}$.
The last column of $A_1$ is ${\bf 1}_n \cdot c \sqrt{\log n}$.

 Assume $n = (d-2) t$ where $t$ is a positive integer.
 For each $A_1 \in {\cal D}_1$, we have
Type I column: the first column of $A_1$ is $e_{j_3} \cdot a_1 \sqrt{ \log n}$ for $j_3 \in [n]$.
Type II column: from the second column to the $(d-1)$-th columns of $A_1$ are  
$\begin{bmatrix} I_{d-2} \\ I_{d-2} \\ \vdots \\ I_{d-2} \end{bmatrix} \cdot b \sqrt{ \log n}$.
Type III column: the last column of $A_1$ is ${\bf 1}_n \cdot c \sqrt{\log n}$.

\end{definition}

\begin{figure*}[!ht]
    \centering
    \subfloat[Vary with $n$]
    {\includegraphics[width=0.32\textwidth]{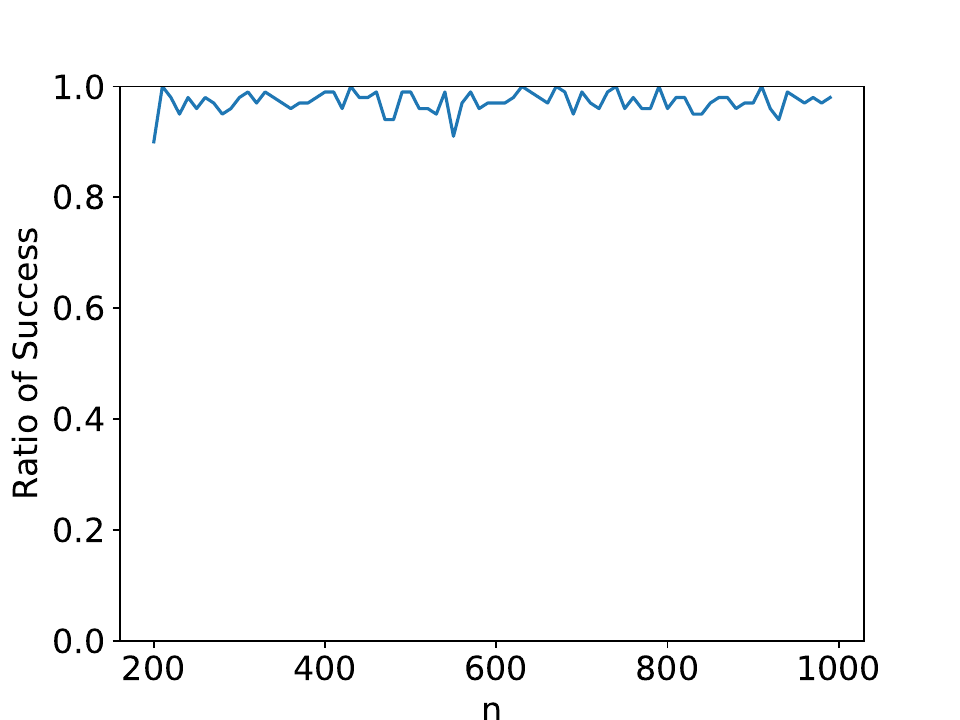}}\label{fig:self_n}
    \subfloat[Var with $d$]
    {\includegraphics[width=0.32\textwidth]{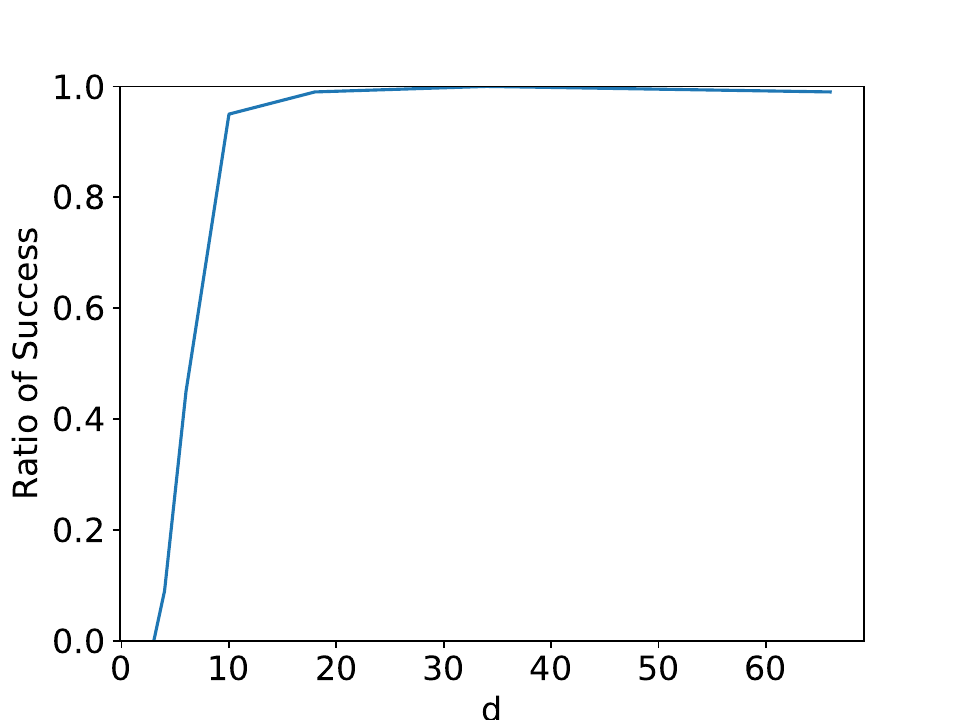}}\label{fig:self_d}
    \subfloat[Vary with $m$]
    {\includegraphics[width=0.32\textwidth]{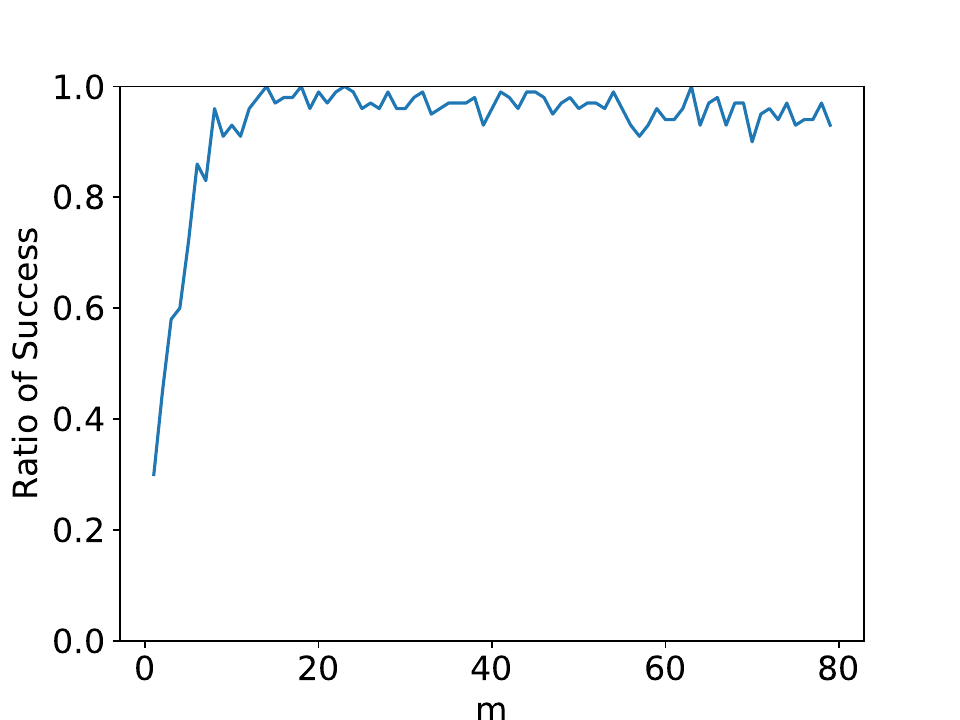}}\label{fig:self_m}
    \subfloat[Vary with $c$]  {\includegraphics[width=0.32\textwidth]{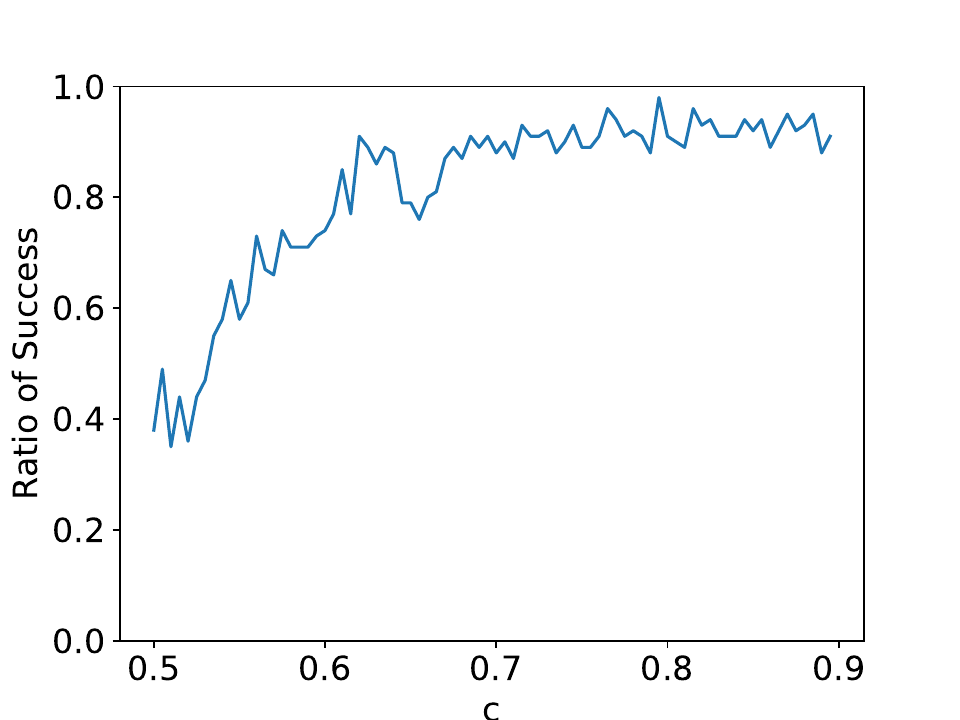}} \label{fig:self_c}
    \subfloat[Vary with $a_0$]  {\includegraphics[width=0.32\textwidth]{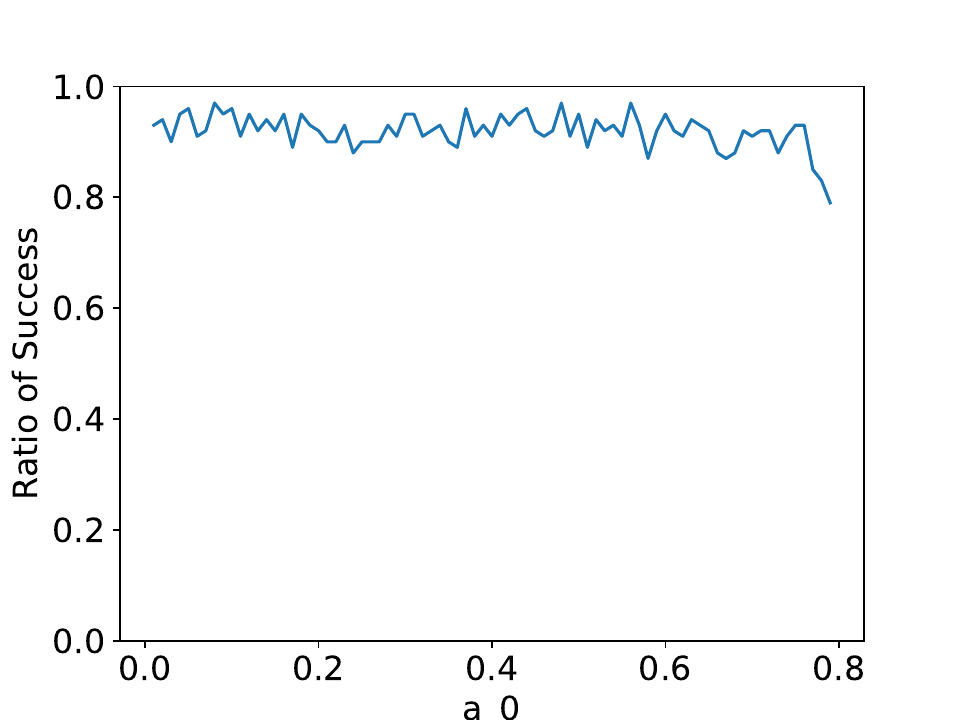}} \label{fig:self_a0}
    \subfloat[Vary with $a_1$]  {\includegraphics[width=0.32\textwidth]{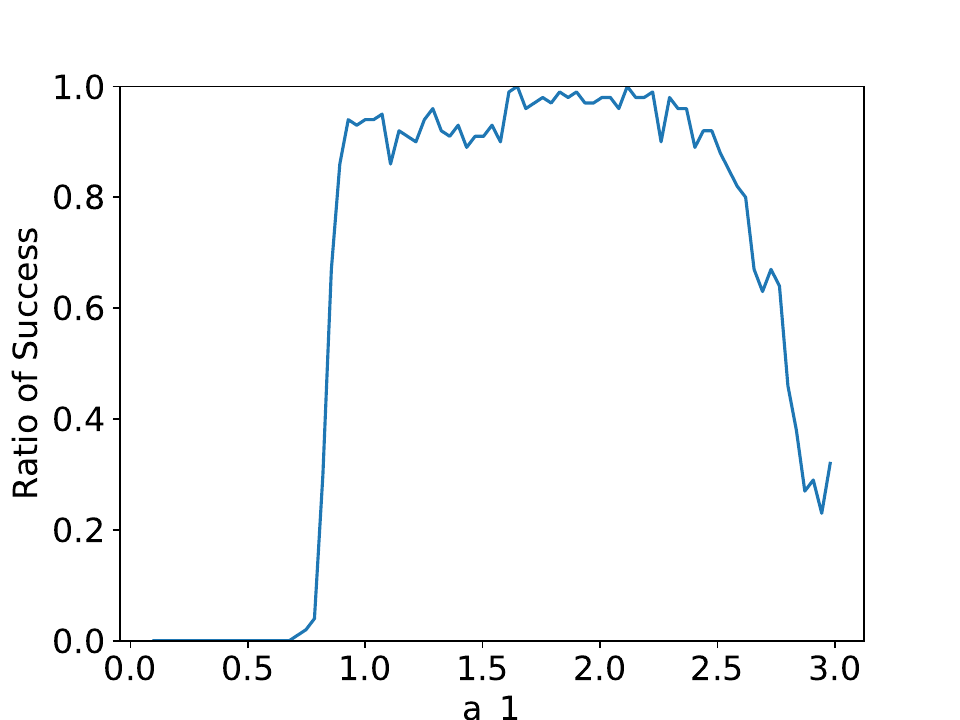}} \label{fig:self_a1}
    \caption{Self-Attention when $X = \mathrm{vec}(I_d)$. Let $n$ denote the length of input sentence. Let $d$ denote the embedding size. Let $m$ denotes the width of the second layer. Let $a_1,a_0,c$ denote the parameters of dataset ${\cal D}_0$ and ${\cal D}_1$.} 
    \label{fig:self}
\end{figure*}

\begin{figure*}[!ht]
    \centering
    \subfloat[Vary with $n$]
    {\includegraphics[width=0.32\textwidth]{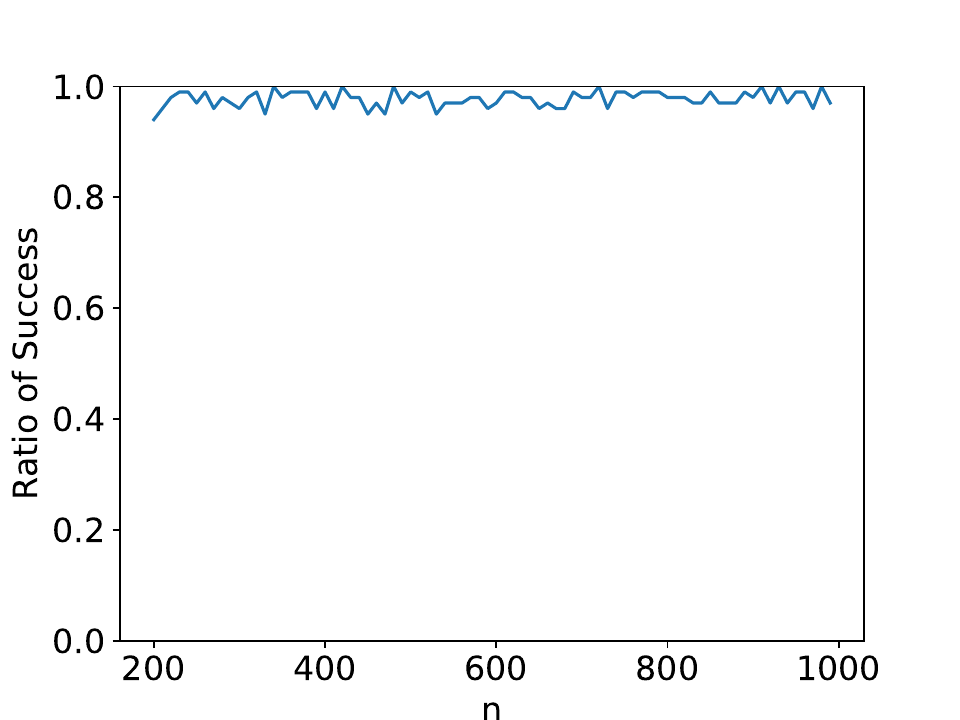}}\label{fig:self_n_all1}
    \subfloat[Vary with $d$]
    {\includegraphics[width=0.32\textwidth]{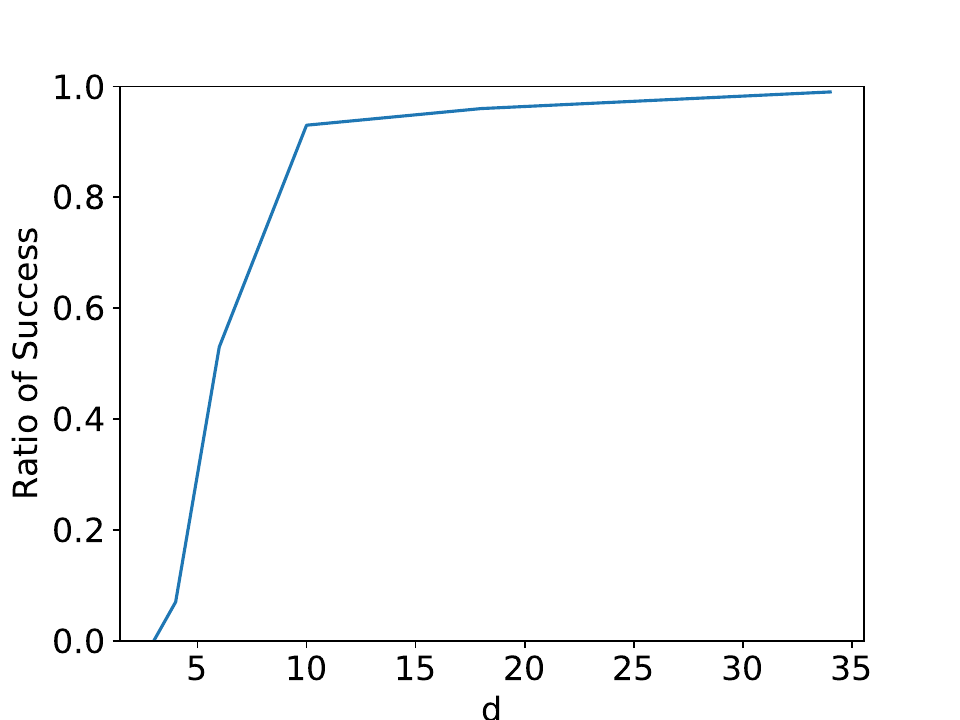}}\label{fig:self_d_all1}
    \subfloat[Vary with $m$]
    {\includegraphics[width=0.32\textwidth]{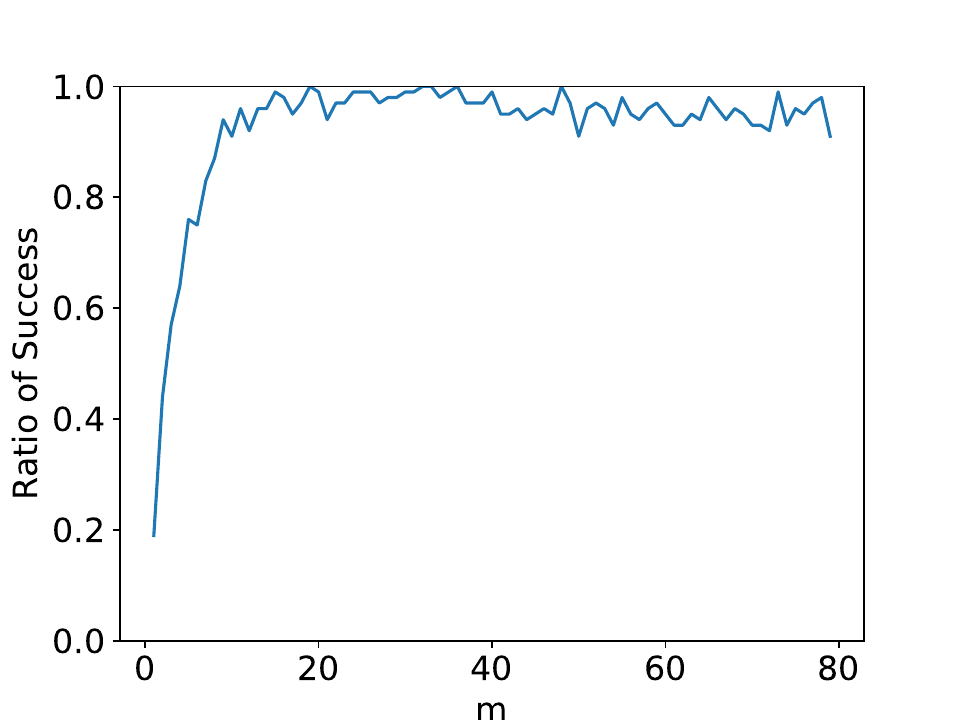}}\label{fig:self_m_all1}
    \subfloat[Vary with $c$]  {\includegraphics[width=0.32\textwidth]{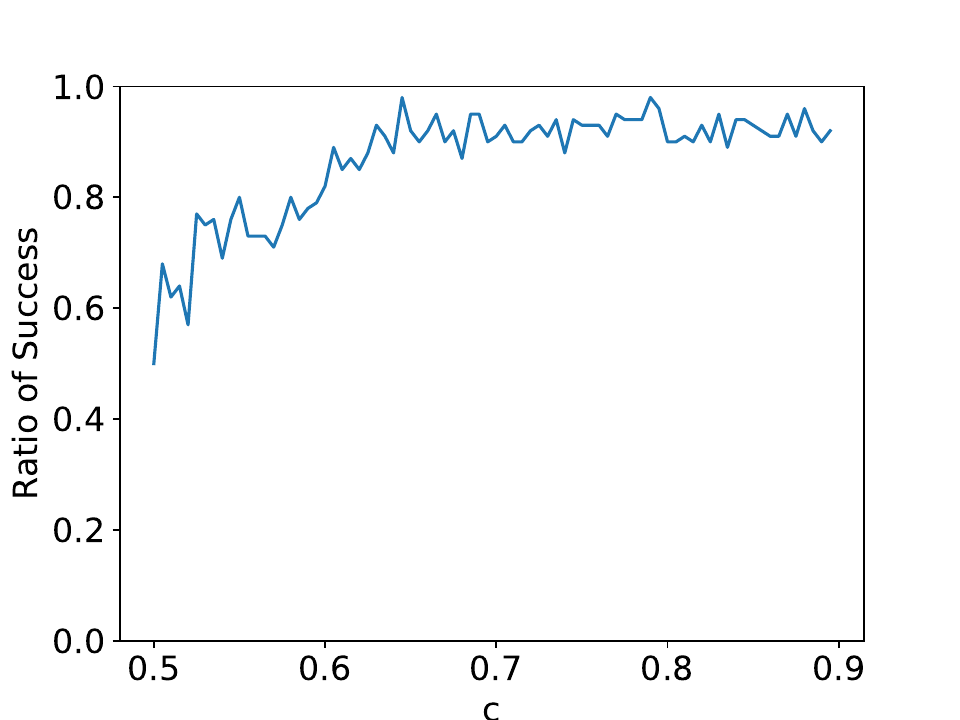}} \label{fig:self_c_all1}
    \subfloat[Vary with $a_0$]  {\includegraphics[width=0.32\textwidth]{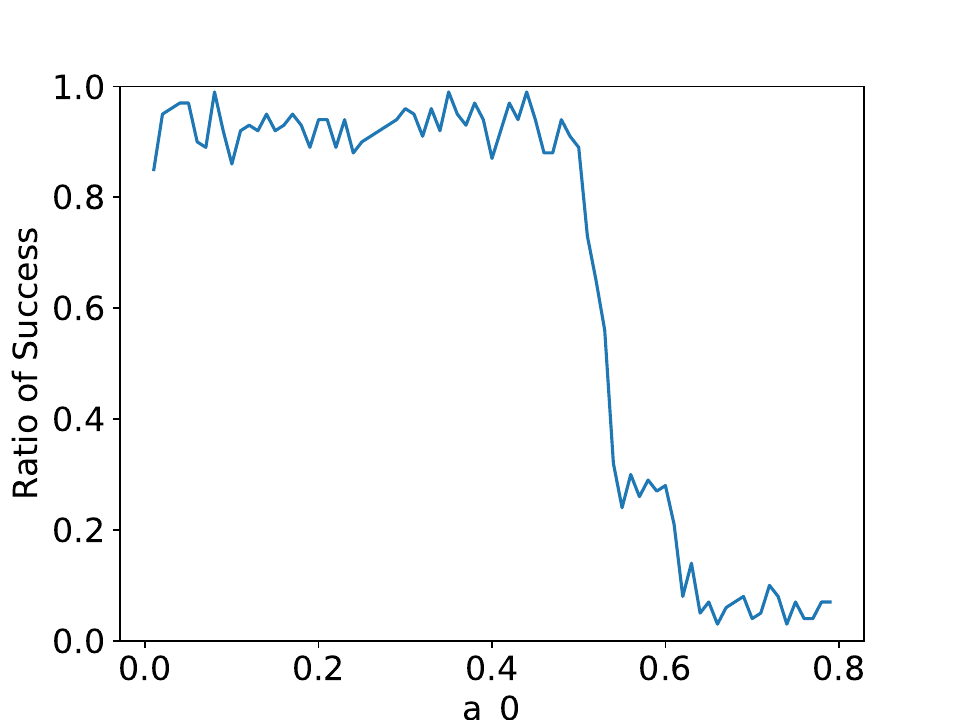}} \label{fig:self_a0_all1}
    \subfloat[Vary with $a_1$]  {\includegraphics[width=0.32\textwidth]{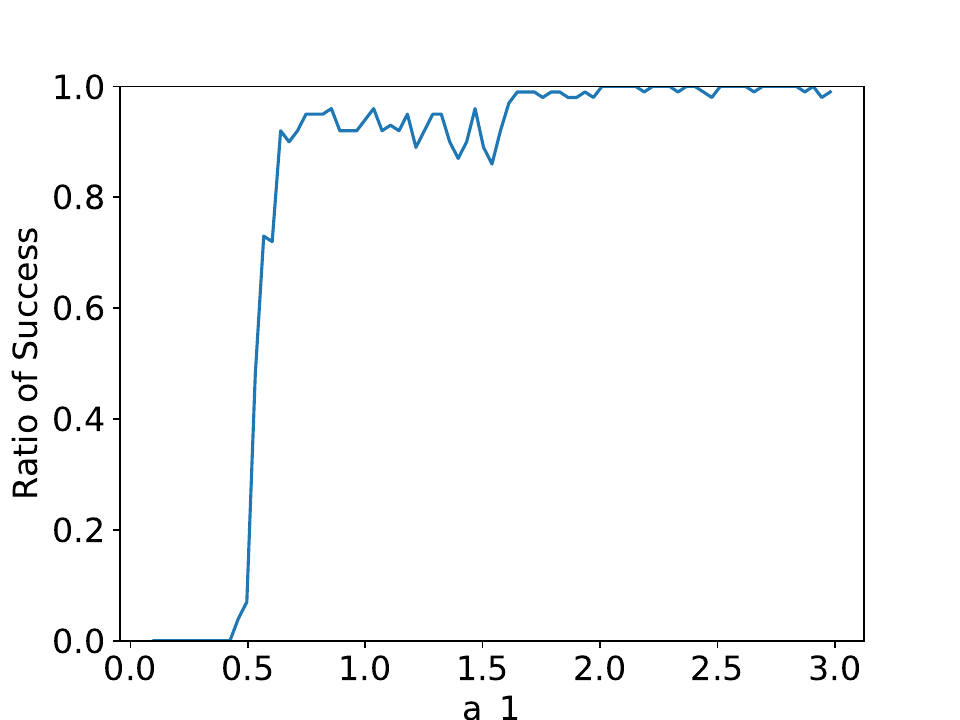}} \label{fig:self_a1_all1}
    \caption{Self-Attention when $X = \mathbf{1}_{n^2}$. Let $n$ denote the length of input sentence. Let $d$ denote the embedding size. Let $m$ denotes the width of the second layer. Let $a_1,a_0,c$ denote the parameters of dataset ${\cal D}_0$ and ${\cal D}_1$.} 
    \label{fig:self_all1}
\end{figure*}

\begin{theorem}[Main results, informal version of Section \ref{sec:app_self_attention}]
Let $\tau = (c+0.1)\sqrt{\log n}$
Let $d \in [ \omega(n^{0.02}) , n ]$
Let $m = O(\log (n/\delta))$
Let $x = {\bf 1}_{d^2}$
Let 
    \begin{align*}
        F_{\exp}(A_1):=\phi( \sum_{j_0=1}^n \sum_{l=1}^m \phi_{\tau}( \langle \mathsf{Att}_{j_0,*}, y_{l} \rangle) )
    \end{align*}
where $\mathsf{Att}_{j_0,*}$ denote the $j_0$-th line of Eq.~\eqref{eq:attention}.
 For any $A_1$ from ${\cal D}_1$ (Definition~\ref{def:dataset_self_attention})
With high probability $1-\delta/\poly(n)$, we have
\begin{align*}
    F_{\exp}(A_1) > 0,~F_{\exp}(A_1) = 0
\end{align*}

For any $A_1$ from ${\cal D}_0$ (Definition~\ref{def:dataset_self_attention})
With high probability $1-\delta/\poly(n)$, we have\begin{align*}
    F_{\lin}(A_1) = 0, ~F_{\lin}(A_1) = 0
\end{align*}
\end{theorem}

\section{Numerical Experiments}\label{sec:numerical_exp}
Here in this section, we present our numerical experiments of our proof. We ran simulation experiments on the softmax regression model, the self attention model and the cross attention model. We deploy all our experiments on an Apple MacBook Pro with M2 chip and 16GB of memory. The Python we use is version 3.9.12. 
\subsection{Softmax Regression Model}

We set the dataset as described in Section~\ref{sec:property_toy}. For a pair of inputs $(A_0,x_0) \in \mathcal{D}_0$ and $(A_1, x_1) \in \mathcal{D}_1$, we define the event $E_{\mathrm{success}}$ as
\begin{align*}
            E_{\mathrm{success}}
    :=   &~ F_{\exp}(A_1, x_1) > 0 
    \land ~ F_{\lin}(A_1, x_1) = 0 \\
    \land&~ F_{\exp}(A_0, x_0) = 0 
    \land~ F_{\lin}(A_0, x_0) = 0. 
\end{align*}
We divide our numerical experiments for parameters $n$ and $m$. To be specific, 
\begin{itemize}
    \item We deploy experiments for $n \in [100, 2000]$ with a step size of $10$. For each $n$, we set $m = \log n$. For each group of the parameters, we generated the models for $1000$ times and count the times $E_{\mathrm{success}}$ happens. The result can be found in Figure (a) in Figure~\ref{fig:toy}. 
    \item We fix $n = 1000$, and deploy experiments for $m \in [1, 100]$ with a step size of $1$. For each group of the parameters, we generated the models for $1000$ times and count the counts $E_{\mathrm{success}}$ happens. The result can be found in Figure (b) in Figure~\ref{fig:toy}. 
\end{itemize}

\subsection{Experiments for Self-Attention}

We set the dataset as described in Section~\ref{sec:self_dataset}. For a pair of inputs $(A_{01}, A_{02}, A_{03}) \in \mathcal{D}_0$ and $(A_{11}, A_{12}, A_{13}) \in \mathcal{D}_1$, we define the event $E_{\mathrm{success}}$ as
\begin{align*}
            E_{\mathrm{success}}
    :=   &~ F_{\exp}(A_{11}, A_{12}, A_{13}) > 0 \\
    \land&~ F_{\lin}(A_{11}, A_{12}, A_{13}) = 0 \\
    \land&~ F_{\exp}(A_{01}, A_{02}, A_{03}) = 0 \\
    \land&~ F_{\lin}(A_{01}, A_{02}, A_{03}) = 0. 
\end{align*}

\subsubsection{For \texorpdfstring{$X = \mathrm{vec}(I_d)$}{}}

We first fix $X = \mathrm{vec}(I_d)$, and deploy the numerical experiments for parameters $n$, $d$, $m$, $c$, $a_0$ and $a_1$. To be specific, 
\begin{itemize}
    \item We deploy experiments for $n \in [200, 1000]$ with a step size of $10$. For each $n$, we set $d = 12$, $\delta = 0.01$, $m = \max\{\log (n/\delta), 15\}$, $a_0 = 0.01$, $a_1 = 1.2$, $b = 0.2$, $c = 0.8$. For each set of parameters, we iteratively generated models 100 times and recorded the occurrences of successful events denoted as $E_{\mathrm{success}}$. The result can be found in Figure (a) in Figure~\ref{fig:self}. 
    
    \item We deploy experiments for $d \in \{4, 6, 10, 18, 34, 66\}$. For each $d$, we set $n = 256$, $\delta = 0.01$, $m = \max\{\log (n/\delta), 15\}$, $a_0 = 0.01$, $a_1 = 1.2$, $b = 0.2$, $c = 0.8$. For each set of parameters, we iteratively generated models 100 times and recorded the occurrences of successful events denoted as $E_{\mathrm{success}}$. The result can be found in Figure (b) in Figure~\ref{fig:self}. 
    
    \item We deploy experiments for $m \in [1, 80]$ with a step size of $1$. For each $m$, we set $n =200$, $d = 22$, $\delta = 0.01$, $a_0 = 0.01$, $a_1 = 1.2$, $b = 0.2$, $c = 0.8$. For each set of parameters, we iteratively generated models 100 times and recorded the occurrences of successful events denoted as $E_{\mathrm{success}}$. The result can be found in Figure (c) in Figure~\ref{fig:self}. 
    
    \item We deploy experiments for $a_1 \in [0.1, 3]$ with a step size of $0.036$. For each $a_1$, we set $n =200$, $d = 22$, $\delta = 0.01$, $m = \max\{\log (n/\delta), 15\}$, $a_0 = 0.01$, $b = 0.2$, $c = 0.8$. For each set of parameters, we iteratively generated models 100 times and recorded the occurrences of successful events denoted as $E_{\mathrm{success}}$. The result can be found in Figure (d) in Figure~\ref{fig:self}. 
    
    \item We deploy experiments for $a_0 \in [0.01, 0.8]$ with a step size of $0.01$. For each $a_0$, we set $n =200$, $d = 22$, $\delta = 0.01$, $m = \max\{\log (n/\delta), 15\}$, $a_0 = 0.01$, $a_1 = 1.2$, $b = 0.2$, $c = 0.8$. For each set of parameters, we iteratively generated models 100 times and recorded the occurrences of successful events denoted as $E_{\mathrm{success}}$. The result can be found in Figure (e) in Figure~\ref{fig:self}. 
    
    \item We deploy experiments for $c \in [0.5, 0.9]$ with a step size of $0.005$. For each $c$, we set $n =200$, $d = 22$, $\delta = 0.01$, $m = \max\{\log (n/\delta), 15\}$, $a_1 = 1.2$, $b = 0.2$. For each set of parameters, we iteratively generated models 100 times and recorded the occurrences of successful events denoted as $E_{\mathrm{success}}$. The result can be found in Figure (f) in Figure~\ref{fig:self}. 
\end{itemize}

\subsubsection{For \texorpdfstring{$X = \mathbf{1}_{n^2}$}{}}

For the setting that $X = \mathbf{1}_{n^2}$, we similarly deploy the experiments on parameters $n$, $d$, $m$, $c$, $a_0$ and $a_1$. The parameter setting and choice are the same as above. The experiment result is shown in Figure~\ref{fig:self_all1}.

\section{Conclusion}\label{sec:conclusion}

The transformer architecture, propelled by its attention mechanism, has revolutionized natural language processing tasks. Particularly, the softmax function utilized in the attention mechanism plays a crucial role in grasping the token interactions within sequences. On the flip side, linear attention, while computationally more efficient, falls short in its performance compared to softmax attention. This paper delves into the core reasons for this observed discrepancy. Through meticulous comparative analysis, it has been elucidated that softmax-based neural networks are more adept at distinguishing between certain datasets compared to their linear attention counterparts, both in self-attention and cross-attention scenarios. This pivotal revelation offers profound insights into the intrinsic workings of attention mechanisms, thereby guiding the path for more optimized future model developments.

\ifdefined\isarxiv
\section*{Acknowledgments}
The authors would like to thank Majid Daliri and Chenyang Li for helpful discussions. 
%\bibliographystyle{alpha}
%\bibliography{ref}
\else
\bibliography{ref}
\bibliographystyle{plainnat}%{apalike}%{alpha}
\section*{Checklist}
\begin{enumerate}

 \item For all models and algorithms presented, check if you include:
 \begin{enumerate}
   \item A clear description of the mathematical setting, assumptions, algorithm, and/or model. [Yes]
   \item An analysis of the properties and complexity (time, space, sample size) of any algorithm. [Yes]
   \item (Optional) Anonymized source code, with specification of all dependencies, including external libraries. [Not Applicable]%\Yichuan{Since this question is optional, shall we just remove it from here? Or we choose N/A?}
 \end{enumerate}

 \item For any theoretical claim, check if you include:
 \begin{enumerate}
   \item Statements of the full set of assumptions of all theoretical results. [Yes]
   \item Complete proofs of all theoretical results. [Yes]
   \item Clear explanations of any assumptions. [Yes]     
 \end{enumerate}

\item For all figures and tables that present empirical results, check if you include:
    \begin{enumerate}
    \item The code, data, and instructions needed to reproduce the main experimental results (either in the supplemental material or as a URL). [Yes]
    \item All the training details (e.g., data splits, hyperparameters, how they were chosen). [Yes]
    \item A clear definition of the specific measure or statistics and error bars (e.g., with respect to the random seed after running experiments multiple times). [Yes]
    \item A description of the computing infrastructure used. (e.g., type of GPUs, internal cluster, or cloud provider). [Yes]
    \end{enumerate}

 \item If you are using existing assets (e.g., code, data, models) or curating/releasing new assets, check if you include:
 \begin{enumerate}
   \item Citations of the creator If your work uses existing assets. [Not Applicable]
   \item The license information of the assets, if applicable. [Yes/No/Not Applicable]
   \item New assets either in the supplemental material or as a URL, if applicable. [Not Applicable]
   \item Information about consent from data providers/curators. [Not Applicable]
   \item Discussion of sensible content if applicable, e.g., personally identifiable information or offensive content. [Not Applicable]
 \end{enumerate}

 \item If you used crowdsourcing or conducted research with human subjects, check if you include:
 \begin{enumerate}
   \item The full text of instructions given to participants and screenshots. [Not Applicable]
   \item Descriptions of potential participant risks, with links to Institutional Review Board (IRB) approvals if applicable. [Not Applicable]
   \item The estimated hourly wage paid to participants and the total amount spent on participant compensation. [Not Applicable]
 \end{enumerate}

 \end{enumerate}
 %%%Zhao: I think this is the correct place to put checklist
\fi

\newpage
\onecolumn
\appendix
\section*{Appendix}

{\bf Roadmap.}
We organize the appendix as follows. Section~\ref{sec:app_prel} provides some preliminaries. We discuss more related work in Section~\ref{sec:app_related_work}. Section~\ref{sec:app_analysis_att} describes the attention model setting. Section~\ref{sec:app_self_attention} gives analysis for self attention when $QK^\top = \mathbf{1}^{d \times d}$ while Section~\ref{sec:app_self_identity} gives a further discussion when $QK^\top = I_d$. Section~\ref{sec:app_cross_attention} provides analysis for cross attention model, with the result figure of our experiment.

\section{Preliminary}
\label{sec:app_prel}

{\bf Notations.}
We used $\R$ to denote real numbers. We use $A \in \R^{n \times d}$ to denote an $n \times d$ size matrix where each entry is a real number. We use $e_j$ to denote the unit vector where $j$-th entry is $1$ and others are $0$. For any positive integer $n$, we use $[n]$ to denote $\{1,2,\cdots, n\}$. For a matrix $A \in \R^{n \times d}$, we use $a_{i,j}$ to denote the an entry of $A$ which is in $i$-th row and $j$-th column of $A$, for each $i \in [n]$, $j \in [d]$. For two vectors $a, b$, we use $\langle a, b\rangle$ to denote their inner product. We use ${\bf 1}_n$ to denote a length-$n$ vector where all the entries are ones, and use ${\bf 1}_{n \times n}$ to denote a $n \times n$ matrix where all the entries are ones. For a vector $x$ or a matrix $A$, we use $\exp(x)$ and $\exp(A)$ to denote the entry-wise exponential operation on them. For matrices $A \in \R^{n_1 \times d_1}$ and $B \in \R^{n_2 \times d_2}$, we use $A \otimes B \in \R^{n_1 n_2 \times d_1d_2}$ to denote a matrix such that its $((i_1 - 1)n_2 + i_2, (j_1 - i)d_2 + j_2)$-th entry is $A_{i_1,j_1} \cdot B_{i_2, j_2}$ for all $i_1 \in [n_1], j_1 \in [d_1], i_2 \in [n_2], j_2 \in [d_2]$. For a matrix $A \in \R^{n \times d}$, we use $\mathrm{vec}(A) \in \R^{nd}$ to denote the vectorization of $A$. We use $I_d$ to denote the $d \times d$ identity matrix. We use $A^\top$ to denote the transpose of a matrix $A$.

Using standard tensor-trick \cite{gsx23_incontext,gsy23_coin,gswy23,as23,as23_tensor}, we know that
\begin{fact}[Tensor-trick]\label{fac:tensor_trick}
Let $A_1, A_2 \in \R^{n \times d}$, let $Q \in \R^{d \times d}$, let $K \in \R^{d \times d}$, we have
\begin{align*}
\vect( \exp(A_1 Q K^\top A_2^\top) ) = \exp(\A \vect(QK^\top))
\end{align*}
where $\A:= A_1 \otimes A_2 \in \R^{n^2 \times d^2}$.
\end{fact}

\section{More Related Works}
\label{sec:app_related_work}
\paragraph{Theory of Transformer.}
With the ascent of LLMs, there's a heightened interest in comprehending their learning prowess and deepening the theoretical underpinnings of transformer models. A key focus is the in-context learning ability of transformers. \cite{gtlv22} empirically demonstrated that transformers can adeptly learn linear function classes in-context. \cite{asa+22} linked in-context learning in transformers to conventional learning algorithms, a connection further validated through linear regression. \cite{zfb23} explored the in-context learning of a single-head self-attention layer, noting its strengths and weaknesses. \cite{wzw23} viewed LLMs through a Bayesian lens, interpreting in-context learning as a Bayesian selection process. 
Delving into the transformer's architecture, \cite{psza23} introduced the "skill location" concept, emphasizing how minor parameter tweaks during fine-tuning can drastically impact performance and continual learning. 
On the capabilities front, \cite{sht23} analyzed the strengths and limitations of transformers, highlighting their varying growth rates in different tasks. \cite{bce+23} provided an in-depth analysis of GPT-4 \cite{o23}, lauding its versatility across domains and advocating for more advanced models. Our research now pivots to a unique 2-layer regression problem, inspired by the transformer paradigm.

\paragraph{Boosting Computation of Attention.}
The task of fine-tuning pre-trained LLMs presents challenges, primarily due to their extensive parameter sets. Efforts have been directed towards devising efficient methods for computing the attention module. The use of locality sensitive hashing (LSH) for attention approximation has been a topic of discussion in several studies, as seen in \cite{kkl20} and \cite{clp+21}. Specifically, \cite{kkl20} introduced two methods to enhance computational efficiency. They employed LSH as an alternative to dot product attention, resulting in a notable reduction in time complexity. Additionally, they adopted a reversible residual layer in place of the conventional residual layer. On the other hand, \cite{clp+21} refined the approximation technique, noting that LSH doesn't consistently demand updates to model parameters. In a different approach, \cite{pmxa23} proposed approximation methods that leverage a transformer-in-transformer (TinT) model to emulate both the forward pass and back-propagation of a transformer, leading to enhanced parameter efficiency. \cite{mgn+23} delved into the efficient fine-tuning of LLMs, especially those with substantial memory requirements. Building upon the traditional ZO-SCD optimizer, they crafted the memory-efficient MeZO gradient estimator that operates using only the forward pass. 
Furthermore, \cite{as23} established a stringent bound for static attention, while \cite{bsz23} validated results concerning the dynamic attention issue. Lastly, \cite{gsyz23} unveiled a quantum algorithm tailored for attention computation.

\paragraph{Optimizer for LLMs.}
Gradient-based algorithms remain a cornerstone in machine learning. In recent times, there has been a surge in research efforts aimed at devising efficient optimizers tailored for LLM-centric optimization challenges. \cite{clmy21} explored large-scale optimization scenarios where basic vector operations on decision variables become unfeasible. To address this, they employed a block gradient estimator, formulating an algorithm that markedly cuts down both query and computational complexities per iteration. Taking a different approach, \cite{rsm+23} introduced the Direct Preference Optimization algorithm. This method fine-tunes LLMs directly using a specified human preference dataset, bypassing the need for explicit reward models or reinforcement learning techniques. In another significant contribution, \cite{llh+23} unveiled an adept second-order optimizer for LLMs. This optimizer is grounded in diagonal Hessian approximation coupled with a clipping mechanism. Interestingly, they also eased the stipulation that the Hessian must be Lipschitz continuous in their convergence proof. Drawing inspiration from this innovative perspective, our research employs a congruent proof methodology, especially when addressing the ReLU function within our regression analysis.

\section{Analysis for \texorpdfstring{$Q,K,V$}{} Attention}
\label{sec:app_analysis_att}
In Section \ref{sec:app:linear_attention}, we give the definition of linear attention.
In Section \ref{sec:app:softmax_attention}, we give the definition of softmax attention.

\subsection{Linear Attention}\label{sec:app:linear_attention}
\begin{definition}[Linear Attention]
Given $A_1, A_2, A_3 \in \R^{n \times d}$.

Let $x = \vect(QK^\top) \in \R^{d^2}$. Let $\A = A_1 \otimes A_2 \in \R^{n^2 \times d^2}$. 

Let $V \in \R^{d \times d}$.

Let $\A_{j_0} \in \R^{n \times d^2}$ denote the $j_0$-th block of $\A$.

For each $j_0 \in [n]$, we define $u_{\lin}(\A,x)_{j_0} \in \R^{n}$ as follows
\begin{align*}
    u_{\lin}(\A,x)_{j_0} := \A_{j_0} x
\end{align*}
We define $\alpha_{\lin}(\A,x)_{j_0}$ as follows
\begin{align*}
 \alpha_{\lin}(\A,x)_{j_0} := \langle  u_{\lin}(\A,x)_{j_0} , {\bf 1}_n \rangle
\end{align*}
We define $f_{\lin}(\A,x)_{j_0}$ as follows
\begin{align*}
    f_{\lin}(\A,x)_{j_0} = \alpha_{\lin}(\A,x)_{j_0}^{-1} \cdot u_{\lin}(\A,x)_{j_0}
\end{align*}
We define $c_{\lin}(\A,x)_{j_0} \in \R^d$ as follows
\begin{align*}
    c_{\lin}(\A,x)_{j_0,i_0}:= \langle \underbrace{ f_{\lin}(\A,x)_{j_0} }_{n \times 1}, \underbrace{ (A_3 V)_{i_0}  }_{n \times 1}\rangle, ~~~\forall i_0 \in [d]
\end{align*}
\end{definition}

\begin{definition}
Let $y\in \R^{d \times m}$. Let $y_{l} \in \R^d$ denote the $l$-th column of $y$. 
We define
\begin{align*}
F_{\lin}(A_1,A_2,A_3) := \phi( \sum_{j_0=1}^n \sum_{l=1}^m \phi_{\tau} ( \langle \underbrace{ c_{\lin} (\A,x)_{j_0} }_{d \times 1}, \underbrace{ y_{l} }_{d \times 1} \rangle ) )
\end{align*}
\end{definition}
\begin{figure}[!ht]
    \centering
    \includegraphics[width = 0.8\textwidth]{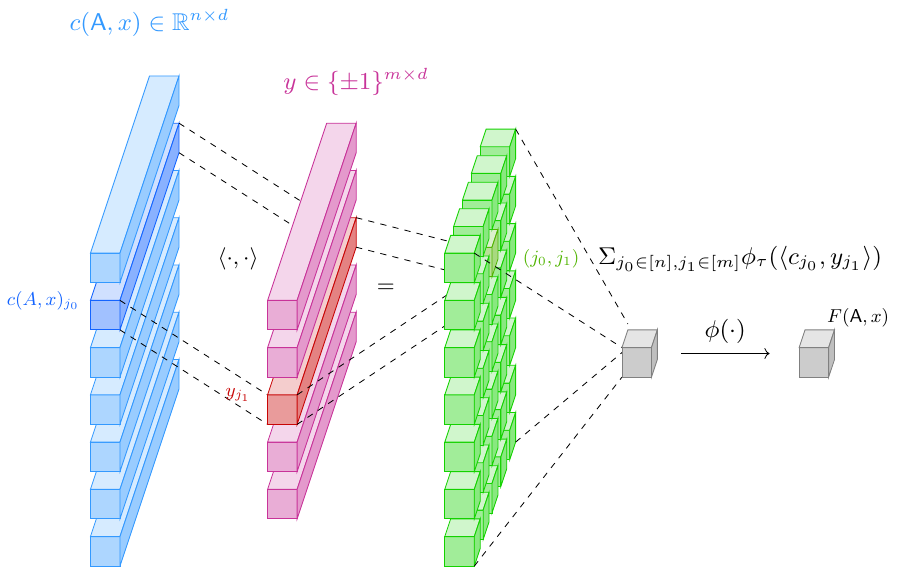}
    \caption{Visualization of our attention computation. 
    }
    \label{fig:att_network}
\end{figure}
\begin{claim}[Equivalence Formula]
If the following conditions hold,
\begin{itemize}
    \item Let $x = \vect(QK^\top)$
    \item Let $(A_1)_{j_0,*}$ denote the $j_0$-th row of $A_1 \in \R^{n \times d}$
    \item Let $\A = A_1 \otimes A_2$
\end{itemize}
Then, we have
\begin{itemize}
    \item $u_{\lin}(\A,x)_{j_0} = ( (A_1)_{j_0,*} Q K^\top A_2^\top )^\top$
    \item $\alpha_{\lin}(A,x)_{j_0} = \langle ( (A_1)_{j_0,*} Q K^\top A_2^\top )^\top, {\bf 1}_n \rangle$
\end{itemize}
\end{claim}
\begin{proof}
The proofs are directly following from tensor-trick (Fact~\ref{fac:tensor_trick}).
\end{proof}

\subsection{Softmax Attention}\label{sec:app:softmax_attention}

\begin{definition}[Softmax Attention]
Given $A_1, A_2, A_3 \in \R^{n \times d}$.

Let $x = \vect(QK^\top) \in \R^{d^2}$. Let $\A = A_1 \otimes A_2 \in \R^{n^2 \times d^2}$. 

Let $\A_{j_0} \in \R^{n \times d^2}$ denote the $j_0$-th block of $\A$.

For each $j_0 \in [n]$, we define $u_{\exp}(\A,x)_{j_0} \in \R^{n}$ as follows
\begin{align*}
    u_{\exp}(\A,x)_{j_0} := \exp( \A_{j_0} x )
\end{align*}
We define $\alpha_{\exp}(\A,x)_{j_0}$ as follows
\begin{align*}
 \alpha_{\exp}(\A,x)_{j_0} := \langle  u_{\exp}(\A,x)_{j_0}, {\bf 1}_n \rangle
\end{align*}
We define $f_{\exp}(\A,x)_{j_0}$ as follows
\begin{align*}
    f_{\exp}(\A,x)_{j_0} = \alpha_{\exp}(\A,x)_{j_0}^{-1} \cdot u_{\exp}(\A,x)_{j_0}
\end{align*}
We define $c_{\exp}(\A,x)_{j_0} \in \R^d$ as follows
\begin{align*}
    c_{\exp}(\A,x)_{j_0,i_0}:= \langle f_{\exp}(\A,x)_{j_0} , (A_3 V)_{i_0} \rangle, ~~~\forall i_0 \in [d]
\end{align*}
\end{definition}

\begin{claim}[Equivalence Formula]
If the following conditions hold,
\begin{itemize}
    \item Let $x = \vect(QK^\top)$
    \item Let $(A_1)_{j_0,*}$ denote the $j_0$-th row of $A_1 \in \R^{n \times d}$
    \item Let $\A = A_1 \otimes A_2$
\end{itemize}
Then, we have
\begin{itemize}
    \item $u_{\exp}(\A,x)_{j_0} =  ( \exp( (A_1)_{j_0,*} Q K^\top A_2^\top ) )^\top$
    \item $\alpha_{\exp}(A,x)_{j_0} = \langle ( \exp( (A_1)_{j_0,*} Q K^\top A_2 )^\top )^\top, {\bf 1}_n \rangle$
\end{itemize}
\end{claim}
\begin{proof}
The proofs are directly following from tensor-trick (Fact~\ref{fac:tensor_trick}).
\end{proof}

\begin{definition}
Let $y\in \R^{n \times m}$. Let $y_{l} \in \R^n$ denote the $l$-th column of $y$. 
We define
\begin{align*}
F_{\exp}(A_1,A_2,A_3) := \phi( \sum_{j_0=1}^n \sum_{l=1}^m \phi_{\tau} ( \langle c_{\exp} (\A,x)_{j_0}, y_{l} \rangle  ) )
\end{align*}
\end{definition}
%\newpage
%\newpage
\section{Self-Attention Dataset}\label{sec:app_self_attention}

 In Section \ref{sec:self_dataset}, we give the definition of the dataset used in the following sections for self-attention.
In Section \ref{sec:app:data_1_with_exp}, we show that the output of the $F_{\exp}$ is greater than $0$ with high probability.
In Section \ref{sec:app:data_0_with_exp}, we show that the output of $F_{\exp}$ is equal to $0$ with high probability. 
In Section \ref{sec:app:data_1_with_lin}, we show that the output of the $F_{\lin}$ is equal to $0$ with high probability.
In Section \ref{sec:app:data_0_with_lin}, we show that the output of $F_{\lin}$ is equal to $0$ with high probability.

\subsection{Definition of Dataset}
\label{sec:self_dataset} 

\begin{definition}[Self-Attention dataset distribution]\label{def:dataset_self_attention}
We define
\begin{itemize}
    \item $a_0 \in (0, 0.1)$
    \item Let $a_1 \geq 0.7$
    \item $b + c = 1$ for $b \geq 0.1 , c \geq 0.1$
\end{itemize}

Given two sets of datasets ${\cal D}_0$ and ${\cal D}_1$ 
\begin{itemize}
    \item For each $\{ A_1, A_2, A_3 \} \in {\cal D}_0$, we have
    \begin{itemize}
        \item $A_1 = A_2 = A_3 \in \R^{n \times d}$
        \item Assume $n = (d-2) t$ where $t$ is a positive integer
        \item the first column of $A_1$ is $e_{j_3} \cdot a_0 \sqrt{\log n}$ for some $j_3 \in [n]$
        \item from the second column to the $(d-1)$-th columns of $A_1$ are $\begin{bmatrix} I_{d-2} \\ I_{d-2} \\ \vdots \\ I_{d-2} \end{bmatrix} \cdot b \sqrt{\log n}$
        \item the last column of $A_1$ is ${\bf 1}_n \cdot c \sqrt{\log n}$
    \end{itemize}
    \item For each $\{ A_1, A_2, A_3 \} \in {\cal D}_1$, we have
    \begin{itemize}
        \item $A_1 = A_2 = A_3 \in \R^{n \times d}$
        \item Assume $n = (d-2) t$ where $t$ is a positive integer
        \item Type I column: the first column of $A_1$ is $e_{j_3} \cdot a_1 \sqrt{ \log n}$ for $j_3 \in [n]$
        \item Type II column: from the second column to the $(d-1)$-th columns of $A_1$ are  
        $\begin{bmatrix} I_{d-2} \\ I_{d-2} \\ \vdots \\ I_{d-2} \end{bmatrix} \cdot b \sqrt{ \log n}$
        \item Type III column: the last column of $A_1$ is ${\bf 1}_n \cdot c \sqrt{\log n}$
    \end{itemize}
\end{itemize}
\end{definition}

%\newpage
\subsection{Dataset 1 with \texorpdfstring{$F_{\exp}$}{}}\label{sec:app:data_1_with_exp}
In Section~\ref{sec:ds_1_u_f_exp} we analyse the property of dataset 1 with respect to function $u_{\exp}$ and $f_{\exp}$. In Section~\ref{sec:ds_1_c_exp} we analyse the property of dataset 1 with respect to function $c_{\exp}$. In Section~\ref{sec:ds_1_y_exp} we analyse the property of dataset 1 with respect to function $c_{\exp}$ with random signs. In Section~\ref{sec:ds_1_F_exp} we show the property of dataset 1 with respect to the output of $F_{\exp}$. 

\subsubsection{Dataset 1 Property when applying function \texorpdfstring{$u_{\exp}$}{} and \texorpdfstring{$f_{\exp}$}{}}
\label{sec:ds_1_u_f_exp}

\begin{lemma}\label{lem:self_dataset_1_f_exp}
If the following conditions hold
\begin{itemize} 
    \item Let $b+c = 1$
    \item Let $\{ A_1, A_2, A_3 \}$ from dataset ${\cal D}_1$ (see Definition~\ref{def:dataset_self_attention})
    \item Let $QK^\top = {\bf 1}_{d \times d} $ (This is the major difference compared to Lemma~\ref{lem:self_dataset_1_f_exp:part2})
\end{itemize} 
Then, for $u_{\exp}(\A,x)_{j_0,j_1}$ and $f_{\exp}(\A,x)_{j_0,j_1}$ entry we have
\begin{itemize}
    \item {\bf Part 1.} For $j_0 = j_3$, 
    \begin{itemize}
        \item {\bf Part 1a.} For $j_1 = j_3$, then  $u_{\exp}(\A,x)_{j_0,j_1} = n^{(1+a_1)^2 } $.
        \item {\bf Part 1b.} For $j_1 \neq j_3$, then  $u_{\exp}(\A,x)_{j_0,j_1} = n^{1+a_1} $.
    \end{itemize}
    \item {\bf Part 2.} For $j_0 \neq j_3$
    \begin{itemize}
        \item {\bf Part 2a.} For $j_1 = j_3$, then $u_{\exp}(\A,x)_{j_0,j_1} = n^{1+a_1}  $
        \item {\bf Part 2b.} For $j_1 \neq j_3$, then $u_{\exp}(A,x)_{j_0,j_1} = n$.
    \end{itemize}
    \item {\bf Part 3.} For $j_0 = j_3$,
    \begin{itemize}
        \item {\bf Part 3a.} For $j_1 = j_3$, then  $f_{\exp}(\A,x)_{j_0,j_1} \geq 1/2$ (if $(1+a_1)^2 > 2 + a_1$)
        \item {\bf Part 3b.} For $j_1 \neq j_3$, then  $f_{\exp}(\A,x)_{j_0,j_1} \leq 1/n$
    \end{itemize}
    \item {\bf Part 4.} For $j_0 \neq j_3$,
    \begin{itemize}
        \item {\bf Part 4a.} For $j_1 = j_3$, then  $f_{\exp}(\A,x)_{j_0,j_1} \leq 1/n^{1-a_1} $  
        \item {\bf Part 4b.} For $j_0 \neq j_3$, then  $f_{\exp}(\A,x)_{j_0,j_1} \leq 1/n$
    \end{itemize}
\end{itemize}
\end{lemma}
\begin{proof}
 
{\bf Proof of Part 1.}

{\bf Proof of Part 1a.}
For $j_0 = j_3$ and $j_1 = j_3$, by computing the tensor product of two $3$-sparse vector
\begin{align*}
& ~ u_{\exp}(\A,x)_{j_0,j_1} \notag \\
= & ~ \exp(  \langle  (a_1\sqrt{\log n}, 0 , \cdots, 0, b \sqrt{\log n}, 0, \cdots, 0, c \sqrt{\log n}) \otimes (a_1\sqrt{\log n}, 0 , \cdots, 0, b \sqrt{\log n}, 0, \cdots, 0, c \sqrt{\log n}) , {\bf 1}_{d \times d} \rangle ) \\ 
= & ~ \exp( ( a_1 \sqrt{\log n} + b \sqrt{\log n} +  c \sqrt{\log n} )^2 ) \\
= & ~ \exp( (a_1 + b + c)^2 \log n  )   \\
= & ~ n^{ (1+a_1)^2 }
\end{align*}
where the first step follows from Definition \ref{def:dataset_self_attention}, the second step follows from algebra, the second step follows from simple algebra and the last step follows from simple algebra.

{\bf Proof of Part 1b.}
For $j_0 = j_3$ and $j_1 \neq j_3$. 
\begin{align*}
u_{\exp}(\A,x)_{j_0,j_1} 
= & ~ \exp( (a_1 \sqrt{\log n} + b \sqrt{\log n}  + c \sqrt{\log n} ) ( b \sqrt{\log n}  + c \sqrt{\log n} )  ) \\
= & ~ \exp( ( (b + c)^2 + a_1(b+c) ) \log n ) \\
= & ~ n^{1+a_1}
\end{align*}
where the first step follows from Definition \ref{def:dataset_self_attention},  
the second step follows from simple algebra and the last step follows from simple algebra.

{\bf Proof of Part 2.}

{\bf Proof of Part 2a.}

For $j_0 \neq j_3$ and $j_1 = j_3$, this case is same as Part 1b.

{\bf Proof of Part 2b.}

For $j_0 \neq j_3$ and $j_1 \neq j_3$,

For the case that we tensor a $2$-sparse vector with another $2$-sparse vector, we have 
\begin{align*}
u_{\exp}(\A,x)_{j_0,j_1} 
= & ~ \exp( ( b \sqrt{\log n} + c \sqrt{\log n} )^2 ) \\
= & ~ \exp( ( b+c )^2 \log n ) \\
= & ~ n
\end{align*}

{\bf Proof of Part 3.}

{\bf Proof of Part 3a.}
We can show
\begin{align*}
f_{\exp}(\A,x)_{j_0,j_1} = & ~ \langle  u_{\exp}(\A,x)_{j_0}, {\bf 1}_n \rangle^{-1} \cdot u_{\exp}(\A,x)_{j_0,j_1}\\ 
= & ~ \frac{1}{ n^{(a_1+1)^2} + (n-1) n^{1+a_1} } \cdot n^{(1+a_1)^2}\\
\geq & ~ \frac{1}{2}
\end{align*}
where the first step follows from the definition of $f_{\exp}$, the second step follows from  {\bf Part 1} and {\bf Part 2} and the last step follows from $(a_1+1)^2 > 2 + a_1$.

{\bf Proof of Part 3b.}
\begin{align*}
f_{\exp}(\A,x)_{j_0,j_1} = & ~ \langle  u_{\exp}(\A,x)_{j_0}, {\bf 1}_n \rangle^{-1} \cdot u_{\exp}(\A,x)_{j_0,j_1}\\ 
= & ~ \frac{1}{ n^{(a_1+1)^2} + (n-1) n^{1+a_1} } \cdot n^{(1+a_1)}\\
\leq & ~ 1/n
\end{align*}

{\bf Proof of Part 4.}

{\bf Proof of Part 4a.}

We can show
\begin{align*}
f_{\exp}(\A,x)_{j_0,j_1} = & ~ \frac{n^{(1+a_1)}}{ n^{1+a_1} + (n-1) \cdot n } \\
\leq & ~ \frac{1}{n^{1-a_1}}
\end{align*}
where the first step follows from  {\bf Part 1} and {\bf Part 2}  
and the last step follows from simple algebra.

{\bf Proof of Part 4b.}

We can show
\begin{align*}
f_{\exp}(\A,x)_{j_0,j_1} = & ~ \frac{n}{ n^{1+a_1} + (n-1) \cdot n } \\
\leq & ~ \frac{1}{n}
\end{align*}
where the first step follows from  {\bf Part 1} and {\bf Part 2} and the last step follows from simple algebra.

\end{proof}

\subsubsection{Dataset 1 Property when applying function \texorpdfstring{$c_{\exp}$}{}}
\label{sec:ds_1_c_exp}

\begin{lemma}\label{lem:self_dataset_1_c_exp}
If the following conditions hold
\begin{itemize}
    \item Let $\{A_1, A_2, A_3 \}$ from dataset ${\cal D}_1$ (see Definition~\ref{def:dataset_self_attention})
    \item Let $QK^\top = {\bf 1}_{d \times d}$
    \item Let $V = I_d$
    \item Let $n = t(d-2)$
\end{itemize}
Then for $c_{\exp}(\A,x)_{j_0,i_0} $ entry we have
\begin{itemize}
    \item {\bf Part 1.} For $j_0=j_3$ and $i_0 = 1$, we have $ c_{\exp}(\A,x)_{j_0,i_0}  \geq \frac{1}{2} \cdot a_1 \sqrt{ \log n}$
    \item {\bf Part 2.} For $j_0=j_3$ and $i_0 \in \{2,\cdots,d-1\}$
    \begin{itemize}
        \item There is only one $i_0$, we have $c_{\exp}(\A,x)_{j_0,i_0} \geq \frac{1}{2} b \sqrt{\log n}$
        \item For the rest of $i_0$, we have $c_{\exp}(\A,x)_{j_0,i_0} \leq \frac{t}{n} b \sqrt{ \log n} $
    \end{itemize}
    \item {\bf Part 3.} For $j_0 = j_3$ and $i_0 = d$, we have $ c_{\exp}(\A,x)_{j_0,i_0} = c \sqrt{\log n}$ 
    \item {\bf Part 4.} For $j_0 \neq j_3$ and $i_0 = 1$, we have $ c_{\exp}(\A,x)_{j_0,i_0} \leq \frac{1}{n^{1-a_1}} a_1 \sqrt{ \log n}$
    \item {\bf Part 5.} For $j_0\neq j_3$ and $i_0 \in \{2,\cdots,d-1\}$, we have $ c_{\exp}(\A,x)_{j_0,i_0} \leq \frac{t}{n} b \sqrt{ \log n}$
    \item {\bf Part 6.} For $j_0\neq j_3$ and $i_0 = d$, we have $ c_{\exp}(\A,x)_{j_0,i_0} = c \sqrt{\log n}$
\end{itemize}
\end{lemma}
\begin{proof}

{\bf Proof of Part 1.}

It follows from Part 3 of Lemma~\ref{lem:self_dataset_1_f_exp} and Type I column in $A_3$ (Definition~\ref{def:dataset_self_attention}).

We can show for one-hot vector $e_{j_3} \in \R^n$, 
\begin{align*}
\langle f_{\exp}(\A,x)_{j_0}, (A_3 V)_{i_0} \rangle
= & ~ \langle f_{\exp}(\A,x)_{j_0} , e_{j_3} \cdot a_1 \sqrt{\log n} \rangle \\
\geq & ~ \frac{1}{2} a_1 \sqrt{ \log n}
\end{align*}
where the first step follows from Type I column in $A_3$  (Definition~\ref{def:dataset_self_attention}) and the last step follows from Part 3 of Lemma~\ref{lem:self_dataset_1_f_exp}.

{\bf Proof of Part 2.}

It follows from Part 3 of Lemma~\ref{lem:self_dataset_1_f_exp} and Type II column in $A_3$ (Definition~\ref{def:dataset_self_attention}).

There are two cases we need to consider, there is one $(A_3 V)_{i_0}$'s $j_3$ coordinate is $1$, in this situation, we know
\begin{align*}
\langle f_{\exp}(\A, x)_{j_0}, (A_3 V)_{i_0} \rangle \geq \frac{1}{2} \cdot b \sqrt{ \log n}
\end{align*}

For the other case, $(A_3 V)_{i_0}$'s $j_3$ coordinate is $0$, in this case we use the property that $(A_3 V)_{i_0}$ is $t$-sparse vector. Then, we have
\begin{align*}
\langle f_{\exp}(\A, x)_{j_0}, (A_3 V)_{i_0} \rangle
\leq & ~ t \cdot \frac{1}{n} \cdot b \sqrt{ \log n} \\
\leq & ~ \frac{t}{n} \cdot b \sqrt{\log n},
\end{align*}
where the first step follows from entries in $f_{\exp}(\mathsf{A}, x)_{j_0}$ are bounded by $\frac{1}{n}$ and at most $t$ entries in $(A_3V)_{i_0}$ are $b\sqrt{\log n}$ and rest are zeros, the second step follows from simple algebra. 

{\bf Proof of Part 3.}

It follows from the fact that $f_{\exp}(\A,x)_{j_0}$ is a normalized vector (so $\langle f_{\exp}(\A,x)_{j_0}, {\bf 1}_n\rangle = 1$) and Type III column in $A_3$ (Definition~\ref{def:dataset_self_attention}).

We can show
\begin{align*}
\langle f_{\exp}(\A,x)_{j_0}, (A_3 V)_{i_0} \rangle = & ~ \langle f_{\exp}(\A,x)_{j_0}, c\sqrt{\log n} \cdot {\bf 1}_n \rangle \\
= & ~ \langle f_{\exp}(\A,x)_{j_0}, {\bf 1}_n \rangle \cdot c \sqrt{\log n} \\
= & ~ c \sqrt{\log n}
\end{align*}

{\bf Proof of Part 4.}

It follows from Part 4 of Lemma~\ref{lem:self_dataset_1_f_exp} and Type I column in $A_3$ (Definition~\ref{def:dataset_self_attention}).
\begin{align*}
\langle f_{\exp}(\A,x)_{j_0}, (A_3 V)_{i_0} \rangle = & ~ \langle f_{\exp}(\A,x)_{j_0}, e_{j_3} \cdot a_1 \sqrt{\log n} \rangle \\
\leq & ~ \frac{1}{n^{1-a_1}} a_1 \sqrt{ \log n}
\end{align*}
where the first step follows from Type I column in $A_3$ and the last step follows from, for $j_0 \neq j_3$, then  $f_{\exp}(\A,x)_{j_0,j_1} \leq 1/n^{1-a_1}$ (Part 4 of Lemma~\ref{lem:self_dataset_1_f_exp}).

{\bf Proof of Part 5.}

It follows from Part 4 of Lemma~\ref{lem:self_dataset_1_f_exp} and Type II column in $A_3$ (Definition~\ref{def:dataset_self_attention}).
\begin{align*}
\langle f_{\exp}(\A,x)_{j_0}, (A_3 V)_{i_0} \rangle = & ~  t b \sqrt{ \log n} \cdot  \frac{1}{n} \\
\leq & ~ \frac{t}{n} b \sqrt{\log n}
\end{align*}
where the first step follows from Type II column in $A_3$ and the last step follows from, for $j_0 \neq j_3$, then  $f_{\exp}(\A,x)_{j_0,j_1} \leq 1/n$ (Part 4 of Lemma~\ref{lem:self_dataset_1_f_exp}).

{\bf Proof of Part 6.}

It follows from the fact that $f_{\exp}(\A,x)_{j_0}$ is a normalized vector and Type III column in $A_3$ (Definition~\ref{def:dataset_self_attention}). 

We can show
\begin{align*}
\langle f_{\exp}(\A,x)_{j_0}, (A_3 V)_{i_0} \rangle = & ~ \langle f_{\exp}(\A,x)_{j_0}, c \sqrt{\log n} \cdot {\bf 1}_n \rangle \\
= & ~ \langle f_{\exp}(\A,x)_{j_0}, {\bf 1}_n \rangle \cdot c \sqrt{\log n} \\
= & ~ c \sqrt{\log n}
\end{align*}

where the first step follows from Type III column in $A_3$ (Definition~\ref{def:dataset_self_attention}), the second step follows from simple algebra and the last step follows from the fact that $f_{\exp}(\A,x)_{j_0}$ is a normalized vector.
\end{proof}

\subsubsection{Dataset 1 Property when applying function \texorpdfstring{$c_{\exp}$}{} with random signs}
\label{sec:ds_1_y_exp}

\begin{lemma}\label{lem:self_dataset_1_random_exp}
If the following conditions hold
\begin{itemize}
    \item Let $\{A_1, A_2,A_3\}$ be from dataset ${\cal D}_1$ (see Definition~\ref{def:dataset_self_attention})
    \item Let $t \sqrt{d} = o(n^{0.99})$ (since $t(d-2) = n$, then we know $\sqrt{d} = \omega(n^{0.01} )$, this implies $d= \omega(n^{0.02})$) .  
\end{itemize}
Then, for each random string $\sigma \in \{-1,+1\}^d$, we have
\begin{itemize}
    \item {\bf Part 1.} If $j_0 = j_3$, then $\Pr[ \langle c_{\exp}(\A,x)_{j_0} , \sigma \rangle \geq (c+0.1) \sqrt{\log n} ] \geq 1/10$
    \item {\bf Part 2.} If $j_0 \neq j_3$, then $\Pr[ \langle c_{\exp}(\A,x)_{j_0} , \sigma \rangle < (c+0.1) \sqrt{\log n} ] \geq 1 - \delta/\poly(n)$
\end{itemize}
\end{lemma}
\begin{proof}
{\bf Proof of Part 1.}

For $j_0 = j_3$, following  Part 1,2,3 of Lemma~\ref{lem:self_dataset_1_c_exp}, there are three cases for $c_{\exp}(\A,x)_{j_0,i_0}$.

For $i_0 = 1$, we have $c_{\exp}(\A,x)_{j_0,i_0} \geq \frac{1}{2} a_1 \sqrt{ \log n}$

For $i_0 \in \{ 2, \cdots, d-1 \}$, there is one $i_0$ such that $c_{\exp}(\A,x)_{j_0,i_0} \geq \frac{1}{2} b \sqrt{\log n}$. For the rest of $i_0$, we have $c_{\exp}(\A,x)_{j_0,i_0} \leq \frac{t}{n } b \sqrt{\log n}$.

For $i_0 = d$, we have $c_{\exp}(\A,x)_{j_0,i_0} = c\sqrt{\log n}$.

Let $S =\{1, i_0' ,d\}$ denote a set of three indices, and $i_0'$ denote that special index.

By hoeffding inequality (Lemma~\ref{lem:hoeffding_bound}), we know that
\begin{align*}
| \langle c_{\exp}(\A,x)_{j_0,[d] \backslash S} , \sigma \rangle  | \leq & ~ O( \sqrt{\log (n/\delta)} ) \cdot \frac{ t \sqrt{d-3} }{ n } \cdot b\sqrt{ \log n} \\
\leq & ~ 0.1 \sqrt{\log n}
\end{align*}
with probability at least $1-\delta/\poly(n)$. Here the last step due to $t\sqrt{d} = o(n^{0.99})$ and $\poly(\log n) \leq n^{0.01}$.

It is obvious that, with probability at least $1/8$, we have 
\begin{align*}
    \langle c_{\exp}(\A,x)_{j_0,S} , \sigma \rangle \geq  \frac{1}{2}a_1 \sqrt{ \log n} + \frac{1}{2} b \sqrt{\log n} + c \sqrt{\log n} \geq (c+0.2) \cdot \sqrt{\log n}
\end{align*}  
Since the probability that $\sigma_{i_0} = 1$ for all three cases is $\frac{1}{8}$.

Hence, combining above two events, we have
\begin{align*}
    \Pr[ \langle c_{\exp}(\A,x)_{j_0} , \sigma \rangle \geq (c + 0.1)\sqrt{\log n} ] \geq 1/10
\end{align*}

{\bf Proof of Part 2.}
It follows from Part 4,5,6 of Lemma~\ref{lem:self_dataset_1_c_exp} and Hoeffding inequality (Lemma~\ref{lem:hoeffding_bound}).

Let $S =\{1,d\}$. 

By hoeffding inequality (Lemma~\ref{lem:hoeffding_bound}), we know that
\begin{align*}
| \langle c_{\exp}(\A,x)_{j_0, [d]\backslash S} , \sigma \rangle  | \leq & ~ O( \sqrt{\log (n/\delta)} ) \cdot \frac{ t \sqrt{d} }{ n } \cdot b\sqrt{ \log n} \\
\leq & ~ 0.05 \sqrt{\log n}
\end{align*}
with probability at least $1-\delta/\poly(n)$. Here the last step due to $t\sqrt{d} = o(n^{0.99})$ and $\poly(\log n) \leq n^{0.01}$.

It is obvious that 
\begin{align*}
|c_{\exp}(\A,x)_{j_0,S} , \sigma \rangle| \leq c\sqrt{\log n} + 0.05 \sqrt{\log n}
\end{align*}

Thus, 
\begin{align*}
    \Pr[ \langle c_{\exp}(\A,x)_{j_0} , \sigma \rangle \leq (c + 0.1)\sqrt{\log n} ] \geq 1- \delta/ \poly(n) %1/10
\end{align*}
Now, we complete the proof.
\end{proof}

\subsubsection{Dataset 1 Property when applying function \texorpdfstring{$F_{\exp}$}{} }
\label{sec:ds_1_F_exp}
\begin{theorem}
If the following conditions hold
\begin{itemize}
    \item Let $d \in [ \omega(n^{0.02}) , n ]$
    \item Let $\tau = (c+0.1)\sqrt{\log n}$
    \item Let $m = O(\log (n/\delta))$
    \item For any $\{A_1,A_2,A_3\}$ from ${\cal D}_1$ (Definition~\ref{def:dataset_self_attention})
    \item Let $x = {\bf 1}_{d^2}$
    \item Let $F_{\exp}(A_1,A_2,A_3):=\phi( \sum_{j_0=1}^n \sum_{j_1=1}^m \phi_{\tau}( \langle c_{\exp}(\A,x) , y_{j_1} \rangle) )$
\end{itemize}
Then we have
\begin{itemize}
    \item With high probability $1-\delta/\poly(n)$, $F_{\exp}(A_1,A_2,A_3) > 0$
\end{itemize}
\end{theorem}
\begin{proof}
It follows from using Lemma~\ref{lem:self_dataset_1_random_exp}.
\end{proof}

%\newpage
\subsection{Dataset 0 with \texorpdfstring{$F_{\exp}$}{}}\label{sec:app:data_0_with_exp}
In Section~\ref{sec:ds_0_u_f_exp} we analyse the property of dataset 0 with respect to function $u_{\exp}$ and $f_{\exp}$. In Section~\ref{sec:ds_0_c_exp} we analyse the property of dataset 0 with respect to function $c_{\exp}$. In Section~\ref{sec:ds_0_y_exp} we analyse the property of dataset 0 with respect to function $c_{\exp}$ with random signs. In Section~\ref{sec:ds_0_F_exp} we show the property of dataset 0 with respect to the output of $F_{\exp}$. 

\subsubsection{Dataset 0 Property when applying function \texorpdfstring{$u_{\exp}$}{} and \texorpdfstring{$f_{\exp}$}{}}
\label{sec:ds_0_u_f_exp}

\begin{lemma}\label{lem:self_dataset_0_f_exp}
If the following conditions hold
\begin{itemize}
    \item Let $\{ A_1, A_2, A_3 \}$ from dataset ${\cal D}_1$ (see Definition~\ref{def:dataset_self_attention})
    \item Let $QK^\top ={\bf 1}_{d \times d} $
\end{itemize} 
Then, for $u_{\exp}(\A,x)_{j_0,j_1}$ and $f_{\exp}(\A,x)_{j_0,j_1}$ entry we have
\begin{itemize}
    \item {\bf Part 1.} For $j_0 = j_3$, 
    \begin{itemize}
        \item {\bf Part 1a.} For $j_1 = j_3$, then  $u_{\exp}(\A,x)_{j_0,j_1} = n^{ (1+ a_0 )^2 } $.
        \item {\bf Part 1b.} For $j_1 \neq j_3$, then  $u_{\exp}(\A,x)_{j_0,j_1} = n^{ (1+a_0) }$.
    \end{itemize}
    \item {\bf Part 2.} For $j_0 \neq j_3$,
    \begin{itemize}
        \item {\bf Part 2a.} For $j_1 = j_3$, then $u_{\exp}(\A,x)_{j_0,j_1} = n^{1+a_0}$.
        \item {\bf Part 2b.} For $j_1 \neq j_3$, then $u_{\exp}(\A,x)_{j_0,j_1} = n $.
    \end{itemize}
    \item {\bf Part 3.} For $j_0 = j_3$, 
    \begin{itemize}
        \item {\bf Part 3a.} For $j_1 = j_3$, then  $f_{\exp}(\A,x)_{j_0,j_1} \leq 1/ n^{1-2a_0}$. (if $a_0 \in (0,0.1)$)
        \item {\bf Part 3b.} For $j_1 \neq j_3$, then  $f_{\exp}(\A,x)_{j_0,j_1} \leq 1/n$.
    \end{itemize}
    \item {\bf Part 4.} For $j_0 \neq j_3$,
    \begin{itemize}
        \item {\bf Part 4a.} For $j_1 = j_3$, then  $f_{\exp}(\A,x)_{j_0,j_1} \leq 1/n^{1-a_0}$.
        \item {\bf Part 4b.} For $j_0 \neq j_3$, then  $f_{\exp}(\A,x)_{j_0,j_1} \leq 1/n$.
    \end{itemize}
\end{itemize}
\end{lemma}
\begin{proof}
 
{\bf Proof of Part 1.}

{\bf Proof of Part 1a.}
For $j_0 = j_3$ and $j_1 = j_3$, by computing the tensor of two $3$-sparse vector
\begin{align*}
u_{\exp}(\A,x)_{j_0,j_1}
= & ~ \exp( ( a_0 \sqrt{\log n} + b \sqrt{\log } + c \sqrt{\log n} )^2 ) \\
= & ~ \exp( (1 + a_0 )^2 \log n  )  
\end{align*}
where the first step follows from Definition \ref{def:dataset_self_attention}, the second step follows from simple algebra and the last step follows from simple algebra.

{\bf Proof of Part 1b.}
For $j_0 = j_3$ and $j_1 \neq j_3$.

In this case we tensor a $2$-sparse vector with another $3$-sparse vector, we get
\begin{align*}
u_{\exp}(\A,x)_{j_0,j_1} 
= & ~ \exp( ( b \sqrt{\log n} + c \sqrt{\log n} ) \cdot ( a_0 \sqrt{ \log n} + b  \sqrt{\log n} + c \sqrt{\log n} ) ) \\
= & ~ \exp( ( (b+c)^2 + a_0(b+c) ) \log n) \\
= & ~ n^{1+a_0}
\end{align*}
where the first step follows from Definition \ref{def:dataset_self_attention}, the second step follows from simple algebra and the last step follows from simple algebra.

{\bf Proof of Part 2.}

{\bf Proof of Part 2a.}

For $j_0 \neq j_3$ and $j_1 = j_3$, this case is same as Part 1b.

{\bf Proof of Part 2b.}

For $j_0 \neq j_3$ and $j_1 \neq j_3$,

For the case that we tensor a $2$-sparse vector with another $2$-sparse vector, we have 
\begin{align*}
u_{\exp}(\A,x)_{j_0,j_1} 
= & ~ \exp( ( b \sqrt{\log n} + c \sqrt{\log n} )^2 ) \\
= & ~ \exp( (b+c)^2 \log n ) \\
= & ~ n
\end{align*}

{\bf Proof of Part 3.}

{\bf Proof of Part 3a.}
We can show
\begin{align*}
f_{\exp}(\A,x)_{j_0,j_1} = & ~ \langle  u_{\exp}(\A,x)_{j_0}, {\bf 1}_n \rangle^{-1} \cdot u_{\exp}(\A,x)_{j_0,j_1}\\ 
= & ~ \frac{1}{ n^{(a_0+1)^2} + (n-1) n^{1+a_0} } \cdot n^{(1+a_0)^2}\\
\leq & ~ \frac{n^{1+2a_0 +a_0^2}}{n^{2+a_0}} \\
= & ~ \frac{1}{n^{1-a_0-a_0^2}} \\
\leq & ~ \frac{1}{n^{1-2a_0}}
\end{align*}
where the first step follows from the definition of $f_{\exp}$, the second step follows from  {\bf Part 1} and {\bf Part 2},  the third step follows from $(a_0+1)^2 \geq (a_0+1)$, and the last step follows from $a_0^2 \leq a_0$.

{\bf Proof of Part 3b.}
\begin{align*}
f_{\exp}(\A,x)_{j_0,j_1} = & ~ \langle  u_{\exp}(\A,x)_{j_0}, {\bf 1}_n \rangle^{-1} \cdot u_{\exp}(\A,x)_{j_0,j_1}\\ 
= & ~ \frac{1}{ n^{(a_0+1)^2} + (n-1) n^{1+a_0} } \cdot n^{(1+a_0)}\\
\leq & ~ 1/n
\end{align*}

{\bf Proof of Part 4.}

{\bf Proof of Part 4a.}

We can show
\begin{align*}
f_{\exp}(\A,x)_{j_0,j_1} = & ~ \frac{n^{(1+a_0)}}{ n^{1+a_0} + (n-1) \cdot n } \\
\leq & ~ \frac{1}{n^{1-a_0}}
\end{align*}
where the first step follows from  {\bf Part 1} and {\bf Part 2}  
and the last step follows from simple algebra.

{\bf Proof of Part 4b.}

We can show
\begin{align*}
f_{\exp}(\A,x)_{j_0,j_1} = & ~ \frac{n}{ n^{1+a_0} + (n-1) \cdot n } \\
\leq & ~ \frac{1}{n}
\end{align*}
where the first step follows from  {\bf Part 1} and {\bf Part 2} and the last step follows from simple algebra.

\end{proof}

\subsubsection{Dataset 0 Property when applying function \texorpdfstring{$c_{\exp}$}{}}
\label{sec:ds_0_c_exp}
\begin{lemma}\label{lem:self_dataset_0_c_exp}
If the following conditions hold
\begin{itemize}
    \item Let $\{A_1, A_2, A_3 \}$ from dataset ${\cal D}_0$ (see Definition~\ref{def:dataset_self_attention})
    \item Let $QK^\top = {\bf 1}_{d \times d}$
    \item Let $V = I_d$
    \item Let $n = t(d-2)$
\end{itemize}
Then for $c_{\exp}(\A,x)_{j_0,i_0} $ entry we have
\begin{itemize}
    \item {\bf Part 1.} For $j_0=j_3$ and $i_0 = 1$, we have $ c_{\exp}(\A,x)_{j_0,i_0}  \leq \frac{1}{n^{1-2a_0}}a_0 \sqrt{  \log n}$
    \item {\bf Part 2.} For $j_0=j_3$ and $i_0 \in \{2,\cdots,d-1\}$,
        we have $c_{\exp}(\A,x)_{j_0,i_0} \leq \frac{t}{n^{1-2a_0}} b \sqrt{ \log n}  $ 
    \begin{itemize}
        \item there is one index $i_0$ such that $\leq ( \frac{1}{ n^{1-2a_0} } + \frac{t-1}{n}  ) b\sqrt{\log n}$
        \item the other indices $i_0$ are $\leq \frac{t}{n} b \sqrt{\log n}$
    \end{itemize}
    \item {\bf Part 3.} For $j_0 = j_3$ and $i_0 = d$, we have $ c_{\exp}(\A,x)_{j_0,i_0}  = c \sqrt{\log n}$ 
    \item {\bf Part 4.} For $j_0 \neq j_3$ and $i_0 = 1$, we have $ c_{\exp}(\A,x)_{j_0,i_0} \leq \frac{1}{n^{1-a_0}} a_0 \sqrt{ \log n}$
    \item {\bf Part 5.} For $j_0\neq j_3$ and $i_0 \in \{2,\cdots,d-1\}$, we have $ c_{\exp}(\A,x)_{j_0,i_0} \leq \frac{t}{n^{1-a_0}} b \sqrt{ \log n}$
    \begin{itemize}
        \item there is one index $i_0$ such that $\leq ( \frac{1}{ n^{1-a_0} } + \frac{t-1}{n}  ) b\sqrt{\log n}$
        \item the other indices $i_0$ are $\leq \frac{t}{n} b \sqrt{\log n}$
    \end{itemize}
    \item {\bf Part 6.} For $j_0\neq j_3$ and $i_0 = d$, we have $ c_{\exp}(\A,x)_{j_0,i_0} = c\sqrt{\log n}$
\end{itemize}
\end{lemma}
\begin{proof}
{\bf Proof of Part 1.}

It follows from Part 3 of Lemma~\ref{lem:self_dataset_0_f_exp}, and Type I column in $A_3$.

{\bf Proof of Part 2.}

It follows from Part 3 of Lemma~\ref{lem:self_dataset_0_f_exp}, and Type II column in $A_3$.

There are two types of entries in $f_{\exp}(\A,x)_{j_0}$:
\begin{itemize}
    \item  there is one entry at most $1/n^{1-2 a_0}$
    \item there are $n-1$ entries at most $1/n$
\end{itemize} 

For $(A_3 V)_{i_0} \in \R^n$, the sparsity is $t$.

Thus there is one index $i_0$
\begin{align*}
\langle f_{\exp}(\A,x)_{j_0}, (A_3 V)_{i_0} \rangle \leq ( \frac{1}{ n^{1-2a_0}} + \frac{t-1}{n} ) b \sqrt{\log n}.
\end{align*}
For the rest of indices $i_0$, we have
\begin{align*}
    \langle f_{\exp}(\A,x)_{j_0}, (A_3 V)_{i_0} \rangle \leq \frac{t}{n} b \sqrt{\log n}.
\end{align*}

{\bf Proof of Part 3.}

It follows from Part 3 of Lemma~\ref{lem:self_dataset_0_f_exp}, and Type III column in $A_3$.

{\bf Proof of Part 4.}

It follows from Part 4 of Lemma~\ref{lem:self_dataset_0_f_exp}, and Type I column in $A_3$.

{\bf Proof of Part 5.}

It follows from Part 4 of Lemma~\ref{lem:self_dataset_0_f_exp}, and Type II column in $A_3$. The proof is similar to Part 2 of this Lemma.

{\bf Proof of Part 6.}

It follows from Part 4 of Lemma~\ref{lem:self_dataset_0_f_exp}, and Type III column in $A_3$.
\end{proof}

\subsubsection{Dataset 0 Property when applying function \texorpdfstring{$c_{\exp}$}{} with random signs}
\label{sec:ds_0_y_exp}
\begin{lemma}\label{lem:self_dataset_0_random_exp}
If the following conditions hold
\begin{itemize}
    \item Let $\{A_1, A_2,A_3,\}$ be from dataset ${\cal D}_0$ (see Definition~\ref{def:dataset_self_attention})
    \item Let $t \sqrt{d} = o(n^{1-2a_0-0.01})$ (since $t(d-2) = n$, then this implies $d = \omega(n^{4 a_0+0.02})$)
\end{itemize}
Then, for each random string $\sigma \in \{-1,+1\}^d$, we have
\begin{itemize}
    \item {\bf Part 1.} If $j_0 = j_3$, then $\Pr[ |\langle c_{\exp}(\A,x)_{j_0} , \sigma \rangle| < (c+0.1) \sqrt{\log n} ] \ge 1-\delta/\poly(n)$
    \item {\bf Part 2.} If $j_0 \neq j_3$, then $\Pr[ | \langle c_{\exp}(\A,x)_{j_0} , \sigma \rangle | < (c+0.1) \sqrt{\log n} ] \ge 1 - \delta/\poly(n)$ 
\end{itemize}
\end{lemma}
\begin{proof}
{\bf Proof of Part 1.}
It follows from Part 1,2,3 of Lemma~\ref{lem:self_dataset_1_c_exp}, random sign distribution.

{\bf Proof of Part 2.}
It follows from Part 4,5,6 of Lemma~\ref{lem:self_dataset_1_c_exp} and Hoeffding inequality.

By hoeffding inequality, we know that
\begin{align*}
| \langle c_{\exp}(\A,x)_{j_0} , \sigma \rangle - c\sqrt{\log n} | \leq & ~ O( \sqrt{\log (n/\delta)} ) \cdot \frac{ t \sqrt{d} }{ n^{1-2a_0} } b \sqrt{ \log n} \\
\leq & ~ 0.1 \sqrt{\log n}
\end{align*}
with probability at least $1-\delta/\poly(n)$. Here the last step due to $t\sqrt{d} = o(n^{1-2a_0 - 0.01})$ and and $\poly(\log n) \leq n^{0.01}$.
\end{proof}

\subsubsection{Dataset 0 Property when applying function \texorpdfstring{$F_{\exp}$}{} }
\label{sec:ds_0_F_exp}
\begin{theorem}
If the following conditions hold
\begin{itemize}
    \item Let $d \in [ \omega(n^{4a_0 + 0.02}) , n ]$
    \item Let $\tau = (c+0.1)\sqrt{\log n}$
    \item Let $m = O(\log (n/\delta))$
    \item For any $\{A_1,A_2,A_3\}$ from ${\cal D}_1$ (Definition~\ref{def:dataset_self_attention})
    \item Let $F_{\exp}(A_1,A_2,A_3):=\phi( \sum_{j_0=1}^n \sum_{j_1=1}^m \phi_{\tau}( \langle c_{\exp}(\A,x)_{j_0} , y_{j_1} ) )$
\end{itemize}
Then we have
\begin{itemize}
    \item With high probability $1-\delta/\poly(n)$, $F_{\exp}(A_1,A_2,A_3) = 0$.  
\end{itemize}
\end{theorem}
\begin{proof}
It follows from using Lemma~\ref{lem:self_dataset_0_random_exp}.  
\end{proof}

%\newpage

\subsection{Dataset 1 with \texorpdfstring{$F_{\lin}$}{}}\label{sec:app:data_1_with_lin}
 
In Section~\ref{sec:ds_1_u_f_lin} we analyse the property of dataset 1 with respect to function $u_{\lin}$ and $f_{\lin}$. In Section~\ref{sec:ds_1_c_lin} we analyse the property of dataset 1 with respect to function $c_{\lin}$. In Section~\ref{sec:ds_1_y_lin} we analyse the property of dataset 1 with respect to function $c_{\lin}$ with random signs. In Section~\ref{sec:ds_1_F_lin} we show the property of dataset 1 with respect to the output of $F_{\lin}$.

\subsubsection{Dataset 1 Property when applying function \texorpdfstring{$u_{\lin}$}{} and \texorpdfstring{$f_{\lin}$}{}}
\label{sec:ds_1_u_f_lin}

\begin{lemma}\label{lem:self_dataset_1_f_lin}
If the following conditions hold
\begin{itemize} 
    \item Let $b+c = 1$
    \item Let $\{ A_1, A_2, A_3 \}$ from dataset ${\cal D}_1$ (see Definition~\ref{def:dataset_self_attention})
    \item Let $QK^\top = {\bf 1}_{d \times d} $
\end{itemize} 
Then, for $u_{\lin}(\A,x)_{j_0,j_1}$ and $f_{\lin }(\A,x)_{j_0,j_1}$ entry we have

\begin{itemize}
    \item {\bf Part 1.} For $j_0 = j_3$, 
    \begin{itemize}
        \item {\bf Part 1a.} For $j_1 = j_3$, then  $u_{\lin}(\A,x)_{j_0,j_1} = (1+ a_1 )^2 \log n$.
        \item {\bf Part 1b.} For $j_1 \neq j_3$, then  $u_{\lin}(\A,x)_{j_0,j_1} = ( 1 + a_1 ) \log n  $.
    \end{itemize}
    \item {\bf Part 2.} For $j_0 \neq j_3$
    \begin{itemize}
        \item {\bf Part 2a.} For $j_1 = j_3$, then $u_{\lin}(\A,x)_{j_0,j_1} = ( 1 + a_1 ) \log n  $
        \item {\bf Part 2b.} For $j_1 \neq j_3$, then $u_{\lin}(A,x)_{j_0,j_1} = \log n$.
    \end{itemize}
    \item {\bf Part 3.} For $j_0 = j_3$,
    \begin{itemize}
        \item {\bf Part 3a.} For $j_1 = j_3$, then  $f_{\lin}(\A,x)_{j_0,j_1} \leq \frac{1+a_1}{n}$
        \item {\bf Part 3b.} For $j_1 \neq j_3$, then  $f_{\lin}(\A,x)_{j_0,j_1} \leq \frac{1}{n}$
    \end{itemize}
    \item {\bf Part 4.} For $j_0 \neq j_3$,
    \begin{itemize}
        \item {\bf Part 4a.} For $j_1 = j_3$, then  $f_{\lin}(\A,x)_{j_0,j_1} \leq \frac{1+a_1}{n}$
        \item {\bf Part 4b.} For $j_0 \neq j_3$, then  $f_{\lin}(\A,x)_{j_0,j_1} \leq \frac{1}{n}$
    \end{itemize}
\end{itemize}

\iffalse
\begin{itemize}
    \item {\bf Part 1.} For $j_0 = j_3$, and $j_1 = j_3$, then  $u_{\lin}(\A,x)_{j_0,j_1} = 4 \log n $.
    \item {\bf Part 2.} For $j_0 \neq j_3$ and $j_1 \neq j_3$, then $u_{\lin}(A,x)_{j_0,j_1} \in [ \log n , 1.36 \log n ] $.
    \item {\bf Part 3.} For $j_0 = j_3$, and $j_1 = j_3$, then  $f_{\lin}(\A,x)_{j_0,j_1} \leq [0, 4/n]$.
    \item {\bf Part 4.} For $j_0 = j_3$, and $j_1 \neq j_3$, then  $f_{\lin}(\A,x)_{j_0,j_1} \leq [0, 4/n]$.
    \item {\bf Part 5.} For $j_0 \neq j_3$, then  $f_{\lin}(\A,x)_{j_0,j_1} \in [0, 4/n]$.
\end{itemize}
\fi
\end{lemma}
\begin{proof}

{\bf Proof of Part 1.}

{\bf Proof of Part 1a.}

For $j_0 = j_3$ and $j_1 = j_3$, by computing the circ product of two $3$-sparse vector
\begin{align*}
u_{\lin}(\A,x)_{j_0,j_1} 
= & ~  ( a_1 \sqrt{\log n} + b \sqrt{\log n} +  c \sqrt{\log n} )^2  \\
= & ~  (a_1 + b +c )^2 \log n \\
= & ~ (a_1 +1)^2 \log n
\end{align*}
where the first step  follows from Definition \ref{def:dataset_self_attention}, the second step follows from simpel algebra and the last step follows from $c+b = 1$.

{\bf Proof of Part 1b.}

For $j_0 = j_3$ and $j_1 \neq j_3$.
\begin{align*}
u_{\lin}(\A,x)_{j_0,j_1} 
= & ~  (a_1 \sqrt{\log n} + b \sqrt{\log n}  + c \sqrt{\log n} ) ( b \sqrt{\log n}  + c \sqrt{\log n} )  \\
= & ~  ( (b + c)^2 + a_1(b+c) ) \log n  \\
= & ~ ( 1 + a_1 ) \log n  
\end{align*}
where the first step  follows from Definition \ref{def:dataset_self_attention}, the second step follows from simple algebra and the last step follows from $c+b = 1$.

{\bf Proof of Part 2.}

{\bf Proof of Part 2a.}

For $j_0 \neq j_3$ and $j_1 = j_3$, this case is same as Part 1b.

{\bf Proof of Part 2b.}
For $j_0 \neq j_3$ and $j_1 \neq j_3$,

For the case that we tensor a $2$-sparse vector with another $2$-sparse vector, we have 
\begin{align*}
u_{\lin}(\A,x)_{j_0,j_1} 
= & ~  ( b \sqrt{\log n}+c \sqrt{\log n} )^2  \\
= & ~  ( b+c)^2  \log n  \\
= & ~ \log n 
\end{align*}
where the first step follows from Definition \ref{def:dataset_self_attention}, the second step follows from simple algebra and the last step follows from $b+c = 1$.

{\bf Proof of Part 3.}

{\bf Proof of Part 3a.}

We can show
\begin{align*}
f_{\lin}(\A,x)_{j_0,j_1} = & ~ \langle  u_{\lin}(\A,x)_{j_0}, {\bf 1}_n \rangle^{-1} \cdot u_{\lin}(\A,x)_{j_0,j_1}\\ 
= & ~ \frac{1}{ (a_1+1)^2 \log n + (n-1) (1+a_1) \log n } \cdot  (a_1+1)^2 \log n\\
\leq & ~ \frac{1}{ (a_1+1) \log n + (n-1) (1+a_1) \log n } \cdot  (a_1+1)^2 \log n\\
= & ~ \frac{a_1+1}{ n  }
\end{align*}
where the first step follows from the definition of $f_{\lin}$, the second step follows from  {\bf Part 1}, the third step follows from simple algebra and the last step follows from simple algebra. 

{\bf Proof of Part 3b.}
\begin{align*}
f_{\lin}(\A,x)_{j_0,j_1} = & ~ \langle  u_{\lin}(\A,x)_{j_0}, {\bf 1}_n \rangle^{-1} \cdot u_{\lin}(\A,x)_{j_0,j_1}\\ 
= & ~ \frac{1}{ (a_1+1)^2 \log n + (n-1) (1+a_1) \log n } \cdot  (1+a_1) \log n \\
= & ~ \frac{1}{ (a_1+1)  + (n-1)   }  \\
\leq & ~ 1/n
\end{align*}
where the first step follows from the definition of $f_{\lin}$, the second step follows from  {\bf Part 1}, the third step follows from simple algebra and the last step follows from simple algebra. 
{\bf Proof of Part 4.}

{\bf Proof of Part 4a.}
We can show
\begin{align*}
f_{\lin}(\A,x)_{j_0,j_1} \leq & ~ \frac{ (1+a_1) \log n }{ (1+a_1) \log n + (n-1) \cdot \log n } \\
\leq & ~ \frac{1+a_1}{n} 
\end{align*}

{\bf Proof of Part 4b.}

We can show
\begin{align*}
f_{\lin}(\A,x)_{j_0,j_1} \leq & ~ \frac{  \log n }{ (1+a_1) \log n + (n-1) \cdot \log n } \\
\leq & ~ \frac{1}{n}
\end{align*}

\end{proof}

\subsubsection{Dataset 1 Property when applying function \texorpdfstring{$c_{\lin}$}{}}
\label{sec:ds_1_c_lin}

\begin{lemma}\label{lem:self_dataset_1_c_lin}
If the following conditions hold
\begin{itemize}
    \item Let $\{A_1, A_2, A_3 \}$ from dataset ${\cal D}_1$ (see Definition~\ref{def:dataset_self_attention})
    \item Let $QK^\top = {\bf 1}_{d \times d}$
    \item Let $V = I_d$
    \item Let $n = t(d-2)$
\end{itemize}
Then for $c_{\lin}(\A,x)_{j_0,i_0} $ entry we have
\begin{itemize}
    \item {\bf Part 1.} For $j_0=j_3$ and $i_0 = 1$, we have $ c_{\lin}(\A,x)_{j_0,i_0}  \leq \frac{ (1+a_1)}{n}  \cdot a_1 \sqrt{\log n}$
    \item {\bf Part 2.} For $j_0=j_3$ and $i_0 \in \{2,\cdots,d-1\}$,
        we have $c_{\lin}(\A,x)_{j_0,i_0} \leq \frac{(1+a_1)}{n} \cdot t \cdot b \sqrt{ \log n}  $ 
    \item {\bf Part 3.} For $j_0 = j_3$ and $i_0 = d$, we have $ c_{\lin}(\A,x)_{j_0,i_0} = c \sqrt{\log n}$ 
    \item {\bf Part 4.} For $j_0 \neq j_3$ and $i_0 = 1$, we have $ c_{\lin}(\A,x)_{j_0,i_0} \leq \frac{(1+a_1)}{n} \cdot a_1 \sqrt{ \log n}$
    \item {\bf Part 5.} For $j_0\neq j_3$ and $i_0 \in \{2,\cdots,d-1\}$, we have $ c_{\lin}(\A,x)_{j_0,i_0} \leq \frac{(1+a_1)}{n} \cdot t \cdot b \sqrt{\log n}$
    \item {\bf Part 6.} For $j_0\neq j_3$ and $i_0 = d$, we have $ c_{\lin}(\A,x)_{j_0,i_0} = c \sqrt{\log n}$
\end{itemize}
\end{lemma}
\begin{proof}
{\bf Proof of Part 1.}

It follows from Part 3 of Lemma~\ref{lem:self_dataset_1_f_lin}, and Type I column in $A_3$.

{\bf Proof of Part 2.}

It follows from Part 3 of Lemma~\ref{lem:self_dataset_1_f_lin}, and Type II column in $A_3$.

We know that each entry in $f_{\lin}(\A,x)_{j_0}$ is at least $0$ and is at most $\frac{1+a_1}{n}$.

We know that each entry in $(A_3 V)_{i_0} \in \R^n$ is at least $0$ and at most $b \sqrt{ \log n}$ and it is $t$-sparse.

Thus,
\begin{align*}
\langle f_{\lin}(\A,x)_{j_0}, (A_3 V)_{i_0} \rangle \leq \frac{(1+a_1)t}{n} b \sqrt{ \log n}
\end{align*}

{\bf Proof of Part 3.}

It follows from Part 3 of Lemma~\ref{lem:self_dataset_1_f_lin}, and Type III column in $A_3$.

{\bf Proof of Part 4.}

It follows from Part 4 of Lemma~\ref{lem:self_dataset_1_f_lin}, and Type I column in $A_3$.

{\bf Proof of Part 5.}

It follows from Part 4 of Lemma~\ref{lem:self_dataset_1_f_lin}, and Type II column in $A_3$.

{\bf Proof of Part 6.}

It follows from Part 4 of Lemma~\ref{lem:self_dataset_1_f_lin}, and Type III column in $A_3$.
\end{proof}

\subsubsection{Dataset 1 Property when applying function \texorpdfstring{$c_{\lin}$}{} with random signs}
\label{sec:ds_1_y_lin}
\begin{lemma}\label{lem:self_dataset_1_random_lin}
If the following conditions hold
\begin{itemize}
    \item Let $\{A_1, A_2,A_3,\}$ be from dataset ${\cal D}_1$ (see Definition~\ref{def:dataset_self_attention})
    \item Let $t \sqrt{d} = o(n^{0.99})$ (since $t(d-2) = n$, then this implies $d = \omega(n^{0.02})$)
\end{itemize}
Then, for each random string $\sigma \in \{-1,+1\}^d$, we have
\begin{itemize}
    \item {\bf Part 1.} If $j_0 = j_3$, then $\Pr[ |\langle c_{\lin}(\A,x)_{j_0} , \sigma \rangle| < (c+0.1) \sqrt{\log n} ] \geq 1-\delta/\poly(n)$
    \item {\bf Part 2.} If $j_0 \neq j_3$, then $\Pr[ | \langle c_{\lin}(\A,x)_{j_0} , \sigma \rangle | < (c+0.1) \sqrt{\log n} ] \geq 1 - \delta/\poly(n)$
\end{itemize}
\end{lemma}
\begin{proof}
{\bf Proof of Part 1.}
It follows from Part 1,2,3 of Lemma~\ref{lem:self_dataset_1_c_lin} and Hoeffding inequality (Lemma~\ref{lem:hoeffding_bound}).

By hoeffding inequality (Lemma~\ref{lem:hoeffding_bound}), we know that
\begin{align*}
| \langle c_{\lin}(\A,x)_{j_0} , \sigma \rangle - c \sqrt{\log n} | \leq & ~ O( \sqrt{\log (n/\delta)} ) \cdot \frac{ (1+a_1) t \sqrt{d} }{ n } b \sqrt{ \log n} \\
\leq & ~ 0.1 \sqrt{\log n}
\end{align*}
with probability at least $1-\delta/\poly(n)$. Here the last step due to $t\sqrt{d} = o(n^{0.99})$, $a_1 = O(1)$, $b = O(1)$, and $\poly(\log n) \leq n^{0.01}$.

{\bf Proof of Part 2.}
It follows from Part 4,5,6 of Lemma~\ref{lem:self_dataset_1_c_lin} and Hoeffding inequality (Lemma~\ref{lem:hoeffding_bound}).

By hoeffding inequality (Lemma~\ref{lem:hoeffding_bound}), we know that
\begin{align*}
| \langle c_{\lin}(\A,x)_{j_0} , \sigma \rangle - c \sqrt{\log n} | \leq & ~ O( \sqrt{\log (n/\delta)} ) \cdot \frac{ (1+a_1) t \sqrt{d} }{ n } b \sqrt{ \log n} \\
\leq & ~ 0.1 \sqrt{\log n}
\end{align*}
with probability at least $1-\delta/\poly(n)$. Here the last step due to $t\sqrt{d} = o(n^{0.99})$, $a_1 = O(1)$, $b = O(1)$, and $\poly(\log n) \leq n^{0.01}$.
\end{proof}

\subsubsection{Dataset 1 Property when applying function \texorpdfstring{$F_{\lin}$}{}}
\label{sec:ds_1_F_lin}
 
\begin{theorem}
If the following conditions hold
\begin{itemize}
    \item Let $d \in [ \omega(n^{0.02}) , n ]$
    \item Let $\tau = (c+0.1)\sqrt{\log n}$
    \item Let $m = O(\log (n/\delta))$
    \item For any $\{A_1,A_2,A_3\}$ from ${\cal D}_1$ (Definition~\ref{def:dataset_self_attention})
    \item Let $F_{\lin}(A_1,A_2,A_3):=\phi( \sum_{j_0=1}^n \sum_{j_1=1}^m \phi_{\tau}( \langle c_{\lin}(\A,x)_{j_0} , y_{j_1} ) )$
\end{itemize}
Then we have
\begin{itemize}
    \item With high probability $1-\delta/\poly(n)$, $F_{\lin}(A_1,A_2,A_3) = 0$
\end{itemize}
\end{theorem}
\begin{proof}
It follows from using Lemma~\ref{lem:self_dataset_1_random_exp}.
\end{proof}

\subsection{Dataset 0 with \texorpdfstring{$F_{\lin}$}{}}\label{sec:app:data_0_with_lin}
 
In Section~\ref{sec:ds_0_u_f_lin} we analyse the property of dataset 0 with respect to function $u_{\lin}$ and $f_{\lin}$. In Section~\ref{sec:ds_0_c_lin} we analyse the property of dataset 0 with respect to function $c_{\lin}$. In Section~\ref{sec:ds_0_y_lin} we analyse the property of dataset 0 with respect to function $c_{\lin}$ with random signs. In Section~\ref{sec:ds_0_F_lin} we show the property of dataset 0 with respect to the output of $F_{\lin}$.

\subsubsection{Dataset 0 Property when applying function \texorpdfstring{$u_{\lin}$}{} and \texorpdfstring{$f_{\lin}$}{}}
\label{sec:ds_0_u_f_lin}

\begin{lemma}\label{lem:self_dataset_0_f_lin}
If the following conditions hold
\begin{itemize}
    \item Let $\{ A_1, A_2, A_3 \}$ from dataset ${\cal D}_0$ (see Definition~\ref{def:dataset_self_attention})
    \item Let $QK^\top = I_d $
\end{itemize} 
Then, for $u_{\lin}(\A,x)_{j_0,j_1}$ and $f_{\lin}(\A,x)_{j_0,j_1}$ entry we have

\begin{itemize}
    \item {\bf Part 1.} For $j_0 = j_3$, 
    \begin{itemize}
        \item {\bf Part 1a.} For $j_1 = j_3$, then  $u_{\lin}(\A,x)_{j_0,j_1} = (1+ a_0 )^2 \log n$.
        \item {\bf Part 1b.} For $j_1 \neq j_3$, then  $u_{\lin}(\A,x)_{j_0,j_1} = ( 1 + a_0 ) \log n  $.
    \end{itemize}
    \item {\bf Part 2.} For $j_0 \neq j_3$
    \begin{itemize}
        \item {\bf Part 2a.} For $j_1 = j_3$, then $u_{\lin}(\A,x)_{j_0,j_1} = ( 1 + a_0 ) \log n  $
        \item {\bf Part 2b.} For $j_1 \neq j_3$, then $u_{\lin}(A,x)_{j_0,j_1} = \log n$.
    \end{itemize}
    \item {\bf Part 3.} For $j_0 = j_3$,
    \begin{itemize}
        \item {\bf Part 3a.} For $j_1 = j_3$, then  $f_{\lin}(\A,x)_{j_0,j_1} \leq \frac{1+a_0}{n}$
        \item {\bf Part 3b.} For $j_1 \neq j_3$, then  $f_{\lin}(\A,x)_{j_0,j_1} \leq \frac{1}{n}$
    \end{itemize}
    \item {\bf Part 4.} For $j_0 \neq j_3$,
    \begin{itemize}
        \item {\bf Part 4a.} For $j_1 = j_3$, then  $f_{\lin}(\A,x)_{j_0,j_1} \leq \frac{1+a_0}{n}$
        \item {\bf Part 4b.} For $j_0 \neq j_3$, then  $f_{\lin}(\A,x)_{j_0,j_1} \leq \frac{1}{n}$
    \end{itemize}
\end{itemize}

\end{lemma}

\begin{proof}

{\bf Proof of Part 1.}

{\bf Proof of Part 1a.}

For $j_0 = j_3$ and $j_1 = j_3$, by computing the circ product of two $3$-sparse vector
\begin{align*}
u_{\lin}(\A,x)_{j_0,j_1} 
= & ~  ( a_0 \sqrt{\log n} + b \sqrt{\log n} +  c \sqrt{\log n} )^2  \\
= & ~  (a_0 + b +c )^2 \log n \\
= & ~ (a_0 +1)^2 \log n
\end{align*}
where the first step  follows from Definition \ref{def:dataset_self_attention}, the second step follows from simple algebra and the last step follows from $c+b = 1$.

{\bf Proof of Part 1b.}

For $j_0 = j_3$ and $j_1 \neq j_3$.
\begin{align*}
u_{\lin}(\A,x)_{j_0,j_1} 
= & ~  (a_0 \sqrt{\log n} + b \sqrt{\log n}  + c \sqrt{\log n} ) ( b \sqrt{\log n}  + c \sqrt{\log n} )  \\
= & ~  ( (b + c)^2 + a_0(b+c) ) \log n  \\
= & ~ ( 1 + a_0 ) \log n  
\end{align*}
where the first step  follows from Definition \ref{def:dataset_self_attention}, the second step follows from simple algebra and the last step follows from $c+b = 1$.

{\bf Proof of Part 2.}

{\bf Proof of Part 2a.}

For $j_0 \neq j_3$ and $j_1 = j_3$, this case is same as Part 1b.

{\bf Proof of Part 2b.}
For $j_0 \neq j_3$ and $j_1 \neq j_3$,

For the case that we tensor a $2$-sparse vector with another $2$-sparse vector, we have 
\begin{align*}
u_{\lin}(\A,x)_{j_0,j_1} 
= & ~  ( b \sqrt{\log n}+c \sqrt{\log n} )^2  \\
= & ~  ( b+c)^2  \log n  \\
= & ~ \log n 
\end{align*}
where the first step follows from Definition \ref{def:dataset_self_attention}, the second step follows from simple algebra and the last step follows from $b+c = 1$.

{\bf Proof of Part 3.}

{\bf Proof of Part 3a.}

We can show
\begin{align*}
f_{\lin}(\A,x)_{j_0,j_1} = & ~ \langle  u_{\lin}(\A,x)_{j_0}, {\bf 1}_n \rangle^{-1} \cdot u_{\lin}(\A,x)_{j_0,j_1}\\ 
= & ~ \frac{1}{ (a_0+1)^2 \log n + (n-1) (1+a_0) \log n } \cdot  (a_0+1)^2 \log n\\
\leq & ~ \frac{1}{ (a_0+1) \log n + (n-1) (1+a_0) \log n } \cdot  (a_0+1)^2 \log n\\
= & ~ \frac{a_0+1}{ n  }
\end{align*}
where the first step follows from the definition of $f_{\lin}$, the second step follows from  {\bf Part 1}, the third step follows from simple algebra and the last step follows from simple algebra. 

{\bf Proof of Part 3b.}
\begin{align*}
f_{\lin}(\A,x)_{j_0,j_1} = & ~ \langle  u_{\lin}(\A,x)_{j_0}, {\bf 1}_n \rangle^{-1} \cdot u_{\lin}(\A,x)_{j_0,j_1}\\ 
= & ~ \frac{1}{ (a_0+1)^2 \log n + (n-1) (1+a_0) \log n } \cdot  (1+a_0) \log n \\
= & ~ \frac{1}{ (a_0+1)  + (n-1)   }  \\
\leq & ~ 1/n
\end{align*}
where the first step follows from the definition of $f_{\lin}$, the second step follows from  {\bf Part 1}, the third step follows from simple algebra and the last step follows from simple algebra. 
{\bf Proof of Part 4.}

{\bf Proof of Part 4a.}
We can show
\begin{align*}
f_{\lin}(\A,x)_{j_0,j_1} \leq & ~ \frac{ (1+a_0) \log n }{ (1+a_0) \log n + (n-1) \cdot \log n } \\
\leq & ~ \frac{1+a_0}{n} 
\end{align*}

{\bf Proof of Part 4b.}

We can show
\begin{align*}
f_{\lin}(\A,x)_{j_0,j_1} \leq & ~ \frac{  \log n }{ (1+a_0) \log n + (n-1) \cdot \log n } \\
\leq & ~ \frac{1}{n}
\end{align*}

\end{proof}

\subsubsection{Dataset 0 Property when applying function \texorpdfstring{$c_{\lin}$}{}}
\label{sec:ds_0_c_lin}

\begin{lemma}\label{lem:self_dataset_0_c_lin}
If the following conditions hold
\begin{itemize}
    \item Let $\{A_1, A_2, A_3 \}$ from dataset ${\cal D}_0$ (see Definition~\ref{def:dataset_self_attention})
    \item Let $QK^\top = {\bf 1}_{d \times d}$
    \item Let $V = I_d$
    \item Let $n = t(d-2)$
\end{itemize}
Then for $c_{\lin}(\A,x)_{j_0,i_0} $ entry we have
\begin{itemize}
    \item {\bf Part 1.} For $j_0=j_3$ and $i_0 = 1$, we have $ c_{\lin}(\A,x)_{j_0,i_0}  \leq \frac{ (1+a_0)}{n}  \cdot a_0 \sqrt{\log n}$
    \item {\bf Part 2.} For $j_0=j_3$ and $i_0 \in \{2,\cdots,d-1\}$,
        we have $c_{\lin}(\A,x)_{j_0,i_0} \leq \frac{(1+a_0)}{n} \cdot t \cdot b \sqrt{ \log n}  $ 
    \item {\bf Part 3.} For $j_0 = j_3$ and $i_0 = d$, we have $ c_{\lin}(\A,x)_{j_0,i_0} = c \sqrt{\log n}$ 
    \item {\bf Part 4.} For $j_0 \neq j_3$ and $i_0 = 1$, we have $ c_{\lin}(\A,x)_{j_0,i_0} \leq \frac{(1+a_0)}{n} \cdot a_0 \sqrt{ \log n}$
    \item {\bf Part 5.} For $j_0\neq j_3$ and $i_0 \in \{2,\cdots,d-1\}$, we have $ c_{\lin}(\A,x)_{j_0,i_0} \leq \frac{(1+a_0)}{n} \cdot t \cdot b \sqrt{\log n}$
    \item {\bf Part 6.} For $j_0\neq j_3$ and $i_0 = d$, we have $ c_{\lin}(\A,x)_{j_0,i_0} = c \sqrt{\log n}$
\end{itemize}
\end{lemma}
\begin{proof}
{\bf Proof of Part 1.}

It follows from Part 3 of Lemma~\ref{lem:self_dataset_0_f_lin}, and Type I column in $A_3$.

{\bf Proof of Part 2.}

It follows from Part 3 of Lemma~\ref{lem:self_dataset_0_f_lin}, and Type II column in $A_3$.

We know that each entry in $f_{\lin}(\A,x)_{j_0}$ is at least $0$ and is at most $\frac{1+a_0}{n}$.

We know that each entry in $(A_3 V)_{i_0} \in \R^n$ is at least $0$ and at most $b \sqrt{ \log n}$ and it is $t$-sparse.

Thus,
\begin{align*}
\langle f_{\exp}(\A,x)_{j_0}, (A_3 V)_{i_0} \rangle \leq \frac{(1+a_0)t}{n} b \sqrt{ \log n}
\end{align*}

{\bf Proof of Part 3.}

It follows from Part 3 of Lemma~\ref{lem:self_dataset_0_f_lin}, and Type III column in $A_3$.

{\bf Proof of Part 4.}

It follows from Part 4 of Lemma~\ref{lem:self_dataset_0_f_lin}, and Type I column in $A_3$.

{\bf Proof of Part 5.}

It follows from Part 4 of Lemma~\ref{lem:self_dataset_0_f_lin}, and Type II column in $A_3$.

{\bf Proof of Part 6.}

It follows from Part 4 of Lemma~\ref{lem:self_dataset_0_f_lin}, and Type III column in $A_3$.
\end{proof}

\subsubsection{Dataset 0 Property when applying function \texorpdfstring{$c_{\lin}$}{} with random signs}
\label{sec:ds_0_y_lin}

\begin{lemma}\label{lem:self_dataset_0_random_lin}
If the following conditions hold
\begin{itemize}
    \item Let $\{A_1, A_2,A_3,\}$ be from dataset ${\cal D}_0$ (see Definition~\ref{def:dataset_self_attention})
    \item Let $t \sqrt{d} = o(n^{0.99})$ (since $t(d-2) = n$, then this implies $d = \omega(n^{0.02})$)
\end{itemize}
Then, for each random string $\sigma \in \{-1,+1\}^d$, we have
\begin{itemize}
    \item {\bf Part 1.} If $j_0 = j_3$, then $\Pr[ |\langle c_{\lin}(\A,x)_{j_0} , \sigma \rangle| < (c+0.1) \sqrt{\log n} ] \geq 1-\delta/\poly(n)$
    \item {\bf Part 2.} If $j_0 \neq j_3$, then $\Pr[ | \langle c_{\lin}(\A,x)_{j_0} , \sigma \rangle | < (c+0.1) \sqrt{\log n} ] \geq 1 - \delta/\poly(n)$
\end{itemize}
\end{lemma}
\begin{proof}
{\bf Proof of Part 1.}
It follows from Part 1,2,3 of Lemma~\ref{lem:self_dataset_0_c_lin} and Hoeffding inequality (Lemma~\ref{lem:hoeffding_bound}).

By hoeffding inequality (Lemma~\ref{lem:hoeffding_bound}), we know that
\begin{align*}
| \langle c_{\lin}(\A,x)_{j_0} , \sigma \rangle - c \sqrt{\log n} | \leq & ~ O( \sqrt{\log (n/\delta)} ) \cdot \frac{ (1+a_0) t \sqrt{d} }{ n } b \sqrt{ \log n} \\
\leq & ~ 0.1 \sqrt{\log n}
\end{align*}
with probability at least $1-\delta/\poly(n)$. Here the last step due to $t\sqrt{d} = o(n^{0.99})$, $a_0 = O(1)$, $b = O(1)$, and $\poly(\log n) \leq n^{0.01}$.

{\bf Proof of Part 2.}
It follows from Part 4,5,6 of Lemma~\ref{lem:self_dataset_0_c_lin} and Hoeffding inequality (Lemma~\ref{lem:hoeffding_bound}).

By hoeffding inequality (Lemma~\ref{lem:hoeffding_bound}), we know that
\begin{align*}
| \langle c_{\lin}(\A,x)_{j_0} , \sigma \rangle - c \sqrt{\log n} | \leq & ~ O( \sqrt{\log (n/\delta)} ) \cdot \frac{ (1+a_0) t \sqrt{d} }{ n } b \sqrt{ \log n} \\
\leq & ~ 0.1 \sqrt{\log n}
\end{align*}
with probability at least $1-\delta/\poly(n)$. Here the last step due to $t\sqrt{d} = o(n^{0.99})$, $a_0 = O(1)$, $b = O(1)$, and $\poly(\log n) \leq n^{0.01}$.
\end{proof}
\subsubsection{Dataset 0 Property when applying function \texorpdfstring{$F_{\exp}$}{} }
\label{sec:ds_0_F_lin}

\begin{theorem}
If the following conditions hold
\begin{itemize}
    \item Let $d \in [ \omega(n^{0.02}) , n ]$
    \item Let $\tau = (c+0.1)\sqrt{\log n}$
    \item Let $m = O(\log (n/\delta))$
    \item For any $\{A_1,A_2,A_3\}$ from ${\cal D}_0$ (Definition~\ref{def:dataset_self_attention})
    \item Let $F_{\exp}(A_1,A_2,A_3):=\phi( \sum_{j_0=1}^n \sum_{j_1=1}^m \phi_{\tau}( \langle c_{\exp}(\A,x)_{j_0} , y_{j_1} ) )$
\end{itemize}
Then we have
\begin{itemize}
    \item With high probability $1-\delta/\poly(n)$, $F_{\lin}(A_1,A_2,A_3) = 0$
\end{itemize}
\end{theorem}
\begin{proof}
It follows from using Lemma~\ref{lem:self_dataset_0_random_lin}.
\end{proof}
%\newpage
\section{Self-Attention Dataset, More discussion}
\label{sec:app_self_identity}

In this section instead of choosing $QK^\top = {\bf 1}_{d \times d}$, we choose $QK^\top = I_d$. The proofs are similar, we only provide the proofs for $F_{\exp}$ outputs $>0$ for ${\cal D}_1$. We omitted the proofs for $F_{\exp}$ outputs $0$ for ${\cal D}_0$ and $F_{\lin}$ outputs $0$ for ${\cal D}_1$ and ${\cal D}_0$. In Section~\ref{sec:ds_1_all1_F_exp} we analyse the dataset 1 with the function $F_{\exp}$.

\subsection{Dataset 1 with \texorpdfstring{$F_{\exp}$}{}}
\label{sec:ds_1_all1_F_exp}
In Section~\ref{sec:ds_1_u_f_exp_all1} we analyse the property of dataset 1 with respect to function $u_{\exp}$ and $f_{\exp}$. In Section~\ref{sec:ds_1_c_exp_all1} we analyse the property of dataset 1 with respect to function $c_{\exp}$. In Section~\ref{sec:ds_1_y_exp_all1} we analyse the property of dataset 1 with respect to function $c_{\exp}$ with random signs. In Section~\ref{sec:ds_1_F_exp_all1} we show the property of dataset 1 with respect to the output of $F_{\exp}$. 

\subsubsection{Dataset 1 Property when applying function \texorpdfstring{$u_{\exp}$}{} and \texorpdfstring{$f_{\exp}$}{}}
\label{sec:ds_1_u_f_exp_all1}

\begin{lemma}\label{lem:self_dataset_1_f_exp:part2}
If the following conditions hold
\begin{itemize} 
    \item Let $a_1 \geq 1$
    \item Let $b^2+c^2 = 1$
    \item Let $\{ A_1, A_2, A_3 \}$ from dataset ${\cal D}_1$ (see Definition~\ref{def:dataset_self_attention})
    \item Let $QK^\top = I_d $ (This is the major difference compared to Lemma~\ref{lem:self_dataset_1_f_exp})
\end{itemize} 
Then, for $u_{\exp}(\A,x)_{j_0,j_1}$ and $f_{\exp}(\A,x)_{j_0,j_1}$ entry we have
\begin{itemize}
    \item {\bf Part 1.} For $j_0 = j_3$, 
    \begin{itemize}
        \item {\bf Part 1a.} For $j_1 = j_3$, then  $u_{\exp}(\A,x)_{j_0,j_1} = n^{(1+a_1)^2 } $.
        \item {\bf Part 1b.} For $j_1 \neq j_3$, then  $u_{\exp}(\A,x)_{j_0,j_1}  \in [n^{c^2} , n^{b^2 + c^2}]$.
        \begin{itemize}
            \item There $t-1$ indices equal $n^{b^2 + c^2}$
            \item There are $n-t$ indices equal to $n^{c^2}$
        \end{itemize}
    \end{itemize}
    \item {\bf Part 2.} For $j_0 \neq j_3$, for all $j_1 \in [n]$, then $u_{\exp}(\A,x)_{j_0,j_1}  \in [n^{c^2} , n^{b^2 + c^2}]$
    \begin{itemize}
            \item There $t-1$ indices equal $n^{b^2 + c^2}$
            \item There are $n-t$ indices equal to $n^{c^2}$
    \end{itemize}
    \item {\bf Part 3.} For $j_0 = j_3$,
    \begin{itemize}
        \item {\bf Part 3a.} For $j_1 = j_3$, then  $f_{\exp}(\A,x)_{j_0,j_1} \geq 1/2$ (if $a_1 > 1$)
        \item {\bf Part 3b.} For $j_1 \neq j_3$, then  $f_{\exp}(\A,x)_{j_0,j_1} \leq 1/n^{1-b^2}$
    \end{itemize}
    \item {\bf Part 4.} For $j_0 \neq j_3$, for all $j_1 \in [n]$, $u_{\exp}(\A,x)_{j_0,j_1}$ has the following properties
    \begin{itemize}
            \item There $t-1$ indices $\leq 1/ n^{1-b^2 }$
            \item There are $n-t$ indices  $\leq 1 / n$
    \end{itemize}
\end{itemize}
\end{lemma}
\begin{proof}
 
{\bf Proof of Part 1.}

{\bf Proof of Part 1a.}
For $j_0 = j_3$ and $j_1 = j_3$, by computing the circ product of two $3$-sparse vector
\begin{align*}
u_{\exp}(\A,x)_{j_0,j_1}
= & ~ \exp( ( a_1 \sqrt{\log n} )^2 + (b \sqrt{\log n} )^2 +  ( c \sqrt{\log n} )^2 ) \\
= & ~ \exp( (a_1^2 + b^2 + c^2) \log n  )   
\end{align*}
where the first step follows from Definition \ref{def:dataset_self_attention}, the second step follows from simple algebra and the last step follows from simple algebra.

{\bf Proof of Part 1b.}
For $j_0 = j_3$ and $j_1 \neq j_3$. There are $(n-1)$ indices $j_1$ in this case.

There are $t$ indices, we have
\begin{align*}
u_{\exp}(\A,x)_{j_0,j_1} 
= & ~ \exp( (  b \sqrt{\log n}  + c \sqrt{\log n} ) ( b \sqrt{\log n}  + c \sqrt{\log n} )^2  ) \\
= & ~ \exp(  (b^2 + c^2)  \log n ) \\
= & ~ n^{b^2+c^2}
\end{align*}
where the first step follows from Definition \ref{def:dataset_self_attention}, the second step follows from simple algebra and the last step follows from simple algebra.

There are $n-t-1$ indices, we have
\begin{align*}
u_{\exp}(\A,x)_{j_0,j_1} 
= & ~ \exp( (c \sqrt{\log n} )^2  ) \\
= & ~ n^{c^2}
\end{align*}

{\bf Proof of Part 2.}

Proof is similar to Part 1b.

{\bf Proof of Part 3.}

{\bf Proof of Part 3a.}
We can show
\begin{align*}
f_{\exp}(\A,x)_{j_0,j_1} = & ~ \langle  u_{\exp}(\A,x)_{j_0}, {\bf 1}_n \rangle^{-1} \cdot u_{\exp}(\A,x)_{j_0,j_1}\\ 
= & ~ \frac{1}{ n^{(a_1^2 + b^2 +c^2)} + (n-1) n^{b^2+c^2} } \cdot n^{(a_1^2 + b^2 + c^2)}\\
\geq & ~ \frac{1}{2}
\end{align*}
where the first step follows from the definition of $f_{\exp}$, the second step follows from  {\bf Part 1} and {\bf Part 2} and the last step follows from $a_1 > 1$.

{\bf Proof of Part 3b.}
\begin{align*}
f_{\exp}(\A,x)_{j_0,j_1} = & ~ \langle  u_{\exp}(\A,x)_{j_0}, {\bf 1}_n \rangle^{-1} \cdot u_{\exp}(\A,x)_{j_0,j_1}\\ 
\leq & ~ \frac{1}{ n^{(a_1^2+b^2 + c^2)} + (n-1) n^{ c^2} } \cdot n^{(b^2 +c^2)}\\
\leq & ~\frac{1}{ n^{c^2} + (n-1) n^{ c^2} }  \cdot n^{(b^2 +c^2)} \\
\leq & ~ 1/n^{1-b^2}
\end{align*}

{\bf Proof of Part 4.}

{\bf Proof of Part 4a.}

There are $t$ indices we have:
\begin{align*}
f_{\exp}(\A,x)_{j_0,j_1} 
\leq & ~ \frac{n^{b^2+c^2}}{ t \cdot n^{b^2+c^2} + (n-t) \cdot n^{c^2} } \\
\leq & ~ \frac{n^{b^2+c^2}}{ t \cdot n^{c^2} + (n-t) \cdot n^{c^2} } \\
\leq & ~ \frac{1}{n^{1-b^2}}
\end{align*}
where the first step follows from  {\bf Part 1} and {\bf Part 2}  
and the last step follows from simple algebra.

{\bf Proof of Part 4b.}
There are $n-t$ indices we have:
\begin{align*}
f_{\exp}(\A,x)_{j_0,j_1} 
\leq & ~ \frac{n^{c^2}}{ t \cdot n^{b^2+c^2} + (n-t) \cdot n^{c^2} } \\
\leq & ~ \frac{n^{c^2}}{ t \cdot n^{c^2} + (n-t) \cdot n^{c^2} } \\
\leq & ~ \frac{1}{n }
\end{align*}
where the first step follows from  {\bf Part 1} and {\bf Part 2} and the last step follows from simple algebra.

\end{proof}

\subsubsection{Dataset 1 Property when applying function \texorpdfstring{$c_{\exp}$}{}}
\label{sec:ds_1_c_exp_all1}

\begin{lemma}\label{lem:self_dataset_1_c_exp:part2}
If the following conditions hold
\begin{itemize}
    \item Let $\{A_1, A_2, A_3 \}$ from dataset ${\cal D}_1$ (see Definition~\ref{def:dataset_self_attention})
    \item Let $QK^\top = {\bf 1}_{d \times d}$
    \item Let $V = I_d$
    \item Let $n = t(d-2)$
\end{itemize}
Then for $c_{\exp}(\A,x)_{j_0,i_0} $ entry we have
\begin{itemize}
    \item {\bf Part 1.} For $j_0=j_3$ and $i_0 = 1$, we have $ c_{\exp}(\A,x)_{j_0,i_0}  \geq \frac{1}{2} \cdot a_1 \sqrt{ \log n}$
    \item {\bf Part 2.} For $j_0=j_3$ and $i_0 \in \{2,\cdots,d-1\}$
    \begin{itemize}
        \item There is only one $i_0$, we have $c_{\exp}(\A,x)_{j_0,i_0} \geq \frac{1}{2} \cdot b \sqrt{\log n}$
        \item For the rest of $i_0$, we have $c_{\exp}(\A,x)_{j_0,i_0} \leq \frac{t}{n^{1-b^2}} \cdot b \sqrt{ \log n} $
    \end{itemize}
    \item {\bf Part 3.} For $j_0 = j_3$ and $i_0 = d$, we have $ c_{\exp}(\A,x)_{j_0,i_0} = c \sqrt{\log n}$ 
    \item {\bf Part 4.} For $j_0 \neq j_3$ and $i_0 = 1$, we have $ c_{\exp}(\A,x)_{j_0,i_0} \leq \frac{1}{n^{1-b^2}} \cdot a_1 \sqrt{ \log n}$
    \item {\bf Part 5.} For $j_0\neq j_3$ and $i_0 \in \{2,\cdots,d-1\}$, we have $ c_{\exp}(\A,x)_{j_0,i_0} \leq \frac{t}{n^{1-b^2}} \cdot b \sqrt{ \log n}$
    \item {\bf Part 6.} For $j_0\neq j_3$ and $i_0 = d$, we have $ c_{\exp}(\A,x)_{j_0,i_0} = c \sqrt{\log n}$
\end{itemize}
\end{lemma}
\begin{proof}

{\bf Proof of Part 1.}

It follows from Part 3 of Lemma~\ref{lem:self_dataset_1_f_exp:part2} and Type I column in $A_3$ (Definition~\ref{def:dataset_self_attention}).

\begin{align*}
\langle f_{\exp}(\A,x)_{j_0}, (A_3 V)_{i_0} \rangle
= & ~ \langle f_{\exp}(\A,x)_{j_0} , e_{j_3} \cdot a_1 \sqrt{\log n} \rangle \\
\geq & ~ \frac{1}{2} a_1 \sqrt{ \log n}
\end{align*}
where the first step follows from Type I column in $A_3$  (Definition~\ref{def:dataset_self_attention}) and the last step follows from Part 3 of Lemma~\ref{lem:self_dataset_1_f_exp:part2}.

{\bf Proof of Part 2.}

It follows from Part 3 of Lemma~\ref{lem:self_dataset_1_f_exp:part2} and Type II column in $A_3$ (Definition~\ref{def:dataset_self_attention}).

There are two cases we need to consider, there is one $(A_3 V)_{i_0}$'s $j_3$ coordinate is $1$, in this situation, we know
\begin{align*}
\langle f_{\exp}(\A, x)_{j_0}, (A_3 V)_{i_0} \rangle \geq \frac{1}{2} \cdot b \sqrt{ \log n}
\end{align*}

For the other case, $(A_3 V)_{i_0}$'s $j_3$ coordinate is $0$, in this case we use the property that $(A_3 V)_{i_0}$ is $t$-sparse vector. Then, we have
\begin{align*}
\langle f_{\exp}(\A, x)_{j_0}, (A_3 V)_{i_0} \rangle
\leq & ~ t \cdot \frac{1}{n^{1-b^2}} \cdot b \sqrt{ \log n} \\
\leq & ~ \frac{t}{n^{1-b^2}} \cdot b \sqrt{\log n}
\end{align*}

{\bf Proof of Part 3.}

It follows from the fact that $f_{\exp}(\A,x)_{j_0}$ is a normalized vector and Type III column in $A_3$ (Definition~\ref{def:dataset_self_attention}).

We can show
\begin{align*}
\langle f_{\exp}(\A,x)_{j_0}, (A_3 V)_{i_0} \rangle = & ~ \langle f_{\exp}(\A,x)_{j_0}, c\sqrt{\log n} \cdot {\bf 1}_n \rangle \\
= & ~ \langle f_{\exp}(\A,x)_{j_0}, {\bf 1}_n \rangle \cdot c \sqrt{\log n} \\
= & ~ c \sqrt{\log n}
\end{align*}

{\bf Proof of Part 4.}

It follows from Part 4 of Lemma~\ref{lem:self_dataset_1_f_exp:part2} and Type I column in $A_3$ (Definition~\ref{def:dataset_self_attention}).
\begin{align*}
\langle f_{\exp}(\A,x)_{j_0}, (A_3 V)_{i_0} \rangle = & ~ \langle f_{\exp}(\A,x)_{j_0}, e_{j_3} \cdot a_1 \sqrt{\log n} \rangle \\
\leq & ~ \frac{1}{n^{1-b^2}} \cdot a_1 \sqrt{ \log n}
\end{align*}
where the first step follows from Type I column in $A_3$ and the last step follows from, for $j_0 \neq j_3$, then  $f_{\exp}(\A,x)_{j_0,j_1} \leq 1/n^{1-a_1}$ (Part 5 of Lemma~\ref{lem:self_dataset_1_f_exp:part2}).

{\bf Proof of Part 5.}

It follows from Part 4 of Lemma~\ref{lem:self_dataset_1_f_exp:part2} and Type II column in $A_3$ (Definition~\ref{def:dataset_self_attention}).
\begin{align*}
\langle f_{\exp}(\A,x)_{j_0}, (A_3 V)_{i_0} \rangle = & ~  t b \sqrt{ \log n} \cdot \cdot \frac{1}{n} \\
\leq & ~ \frac{t}{n^{1-b^2}} b \sqrt{\log n}
\end{align*}
where the first step follows from Type II column in $A_3$ and the last step follows from, for $j_0 \neq j_3$, then  $f_{\exp}(\A,x)_{j_0,j_1} \leq 1/n^{1-b^2}$ (Part 4 of Lemma~\ref{lem:self_dataset_1_f_exp:part2}).

{\bf Proof of Part 6.}

It follows from the fact that $f_{\exp}(\A,x)_{j_0}$ is a normalized vector and Type III column in $A_3$ (Definition~\ref{def:dataset_self_attention}). 

We can show
\begin{align*}
\langle f_{\exp}(\A,x)_{j_0}, (A_3 V)_{i_0} \rangle = & ~ \langle f_{\exp}(\A,x)_{j_0}, c \sqrt{\log n} \cdot {\bf 1}_n \rangle \\
= & ~ \langle f_{\exp}(\A,x)_{j_0}, {\bf 1}_n \rangle \cdot c \sqrt{\log n} \\
= & ~ c \sqrt{\log n}
\end{align*}

where the first step follows from Type III column in $A_3$ (Definition~\ref{def:dataset_self_attention}), the second step follows from simple algebra and the last step follows from the fact that $f_{\exp}(\A,x)_{j_0}$ is a normalized vector.
\end{proof}

\subsubsection{Dataset 1 Property when applying function \texorpdfstring{$c_{\exp}$}{} with random signs}
\label{sec:ds_1_y_exp_all1}

\begin{lemma}\label{lem:self_dataset_1_random_exp:part2}
If the following conditions hold
\begin{itemize}
    \item Let $\{A_1, A_2,A_3\}$ be from dataset ${\cal D}_1$ (see Definition~\ref{def:dataset_self_attention})
    \item Let $t \sqrt{d} = o(n^{1-b^2-0.01})$ (since $t(d-2) = n$, then we know $\sqrt{d} = \omega(n^{b^2+0.01} )$, this implies $d= \omega(n^{2b^2+0.02})$) .  
\end{itemize}
Then, for each random string $\sigma \in \{-1,+1\}^d$, we have
\begin{itemize}
    \item {\bf Part 1.} If $j_0 = j_3$, then $\Pr[ \langle c_{\exp}(\A,x)_{j_0} , \sigma \rangle \geq (c+0.1) \sqrt{\log n} ] \geq 1/10$
    \item {\bf Part 2.} If $j_0 \neq j_3$, then $\Pr[ \langle c_{\exp}(\A,x)_{j_0} , \sigma \rangle < (c+0.1) \sqrt{\log n} ] \geq 1 - \delta/\poly(n)$
\end{itemize}
\end{lemma}
\begin{proof}
{\bf Proof of Part 1.}

For $j_0 = j_3$, following  Part 1,2,3 of Lemma~\ref{lem:self_dataset_1_c_exp:part2}, there are three cases for $c_{\exp}(\A,x)_{j_0,i_0}$.

For $i_0 = 1$, we have $c_{\exp}(\A,x)_{j_0,i_0} \geq \frac{1}{2} a_1 \sqrt{ \log n}$

For $i_0 \in \{ 2, \cdots, d-1 \}$, there is one $i_0$ such that $c_{\exp}(\A,x)_{j_0,i_0} \geq \frac{1}{2} b \sqrt{\log n}$. For the rest of $i_0$, we have $c_{\exp}(\A,x)_{j_0,i_0} \leq \frac{t}{n^{1-b^2} } b \sqrt{\log n}$.

For $i_0 = d$, we have $c_{\exp}(\A,x)_{j_0,i_0} = c\sqrt{\log n}$.

Let $S =\{1, i_0' ,d\}$ denote a set of three indices, and $i_0'$ denote that special index.

By hoeffding inequality (Lemma~\ref{lem:hoeffding_bound}), we know that
\begin{align*}
| \langle c_{\exp}(\A,x)_{j_0,[d] \backslash S} , \sigma \rangle  | \leq & ~ O( \sqrt{\log (n/\delta)} ) \cdot \frac{ t \sqrt{d-3} }{ n^{1-b^2} } \cdot b\sqrt{ \log n} \\
\leq & ~ 0.1 \sqrt{\log n}
\end{align*}
with probability at least $1-\delta/\poly(n)$. Here the last step due to $t\sqrt{d} = o(n^{1-b^2-0.01})$ and $\poly(\log n) \leq n^{0.01}$.

It is obvious that, with probability at least $1/8$, we have 
\begin{align*}
    \langle c_{\exp}(\A,x)_{j_0,S} , \sigma \rangle \geq  \frac{1}{2}a_1 \sqrt{ \log n} + \frac{1}{2} b \sqrt{\log n} + c \sqrt{\log n} \geq (c+0.2) \cdot \sqrt{\log n}
\end{align*}  
Since the probability that $\sigma_{i_0} = 1$ for all three cases is $\frac{1}{8}$.

Hence, combining above two events, we have
\begin{align*}
    \Pr[ \langle c_{\exp}(\A,x)_{j_0} , \sigma \rangle \geq (c + 0.1)\sqrt{\log n} ] \geq 1/8 - \delta/\poly(n) \geq 1/10
\end{align*}

{\bf Proof of Part 2.}
It follows from Part 4,5,6 of Lemma~\ref{lem:self_dataset_1_c_exp:part2} and Hoeffding inequality (Lemma~\ref{lem:hoeffding_bound}).

Let $S =\{1,d\}$. 

By hoeffding inequality (Lemma~\ref{lem:hoeffding_bound}), we know that
\begin{align*}
| \langle c_{\exp}(\A,x)_{j_0, [d]\backslash S} , \sigma \rangle  | \leq & ~ O( \sqrt{\log (n/\delta)} ) \cdot \frac{ t \sqrt{d} }{ n^{1-b^2} } \cdot b\sqrt{ \log n} \\
\leq & ~ 0.0.5 \sqrt{\log n}
\end{align*}
with probability at least $1-\delta/\poly(n)$. Here the last step due to $t\sqrt{d} = o(n^{1-b^2-0.01})$ and $\poly(\log n) \leq n^{0.01}$.

It is obvious that 
\begin{align*}
|c_{\exp}(\A,x)_{j_0,S} , \sigma \rangle| \leq c\sqrt{\log n} + 0.05 \sqrt{\log n}
\end{align*}

Thus, 
\begin{align*}
    \Pr[ \langle c_{\exp}(\A,x)_{j_0} , \sigma \rangle \leq (c + 0.1)\sqrt{\log n} ] \geq 1- \delta/ \poly(n) %1/10
\end{align*}
Now, we complete the proof.
\end{proof}

\subsubsection{Dataset 1 Property when applying function \texorpdfstring{$F_{\exp}$}{} }
\label{sec:ds_1_F_exp_all1}

\begin{theorem}
If the following conditions hold
\begin{itemize}
    \item Let $a_1 \geq 1$
    \item Let $b^2 + c^2 = 1$
    \item Let $b \in (0,0.2)$
    \item Let $d \in [ \omega(n^{2b^2+0.02}) , n ]$
    \item Let $\tau = (c+0.1)\sqrt{\log n}$
    \item Let $m = O(\log (n/\delta))$
    \item For any $\{A_1,A_2,A_3\}$ from ${\cal D}_1$ (Definition~\ref{def:dataset_self_attention})
    \item Let $QK^\top = I_d$
    \item Let $F_{\exp}(A_1,A_2,A_3):=\phi( \sum_{j_0=1}^n \sum_{j_1=1}^m \phi_{\tau}( \langle c_{\exp}(\A,x)_{j_0} , y_{j_1} \rangle) )$
\end{itemize}
Then we have
\begin{itemize}
    \item With high probability $1-\delta/\poly(n)$, $F_{\exp}(A_1,A_2,A_3) > 0$
\end{itemize}
\end{theorem}
\begin{proof}
It follows from using Lemma~\ref{lem:self_dataset_1_random_exp:part2}.
\end{proof}
%\newpage
\section{Cross-Attention Dataset}
\label{sec:app_cross_attention}
 In Section \ref{sec:cross_dataset}, we give the definition of the dataset used in the following parts for the cross-attention..
In Section \ref{sec:app:data1_exp_cross}, we show that the output of the $F_{\exp}$ is greater than $0$ with high probability.
In Section \ref{sec:app:data0_exp_cross}, we show that the output of $F_{\exp}$ is equal to $0$ with high probability. 
In Section \ref{sec:app:data1_lin_cross}, we show that the output of the $F_{\lin}$ is equal to $0$ with high probability.
In Section \ref{sec:app:data0_lin_cross}, we show that the output of $F_{\lin}$ is equal to $0$ with high probability. 
 
\subsection{Definition of Dataset}
\label{sec:cross_dataset}

\begin{definition}\label{def:dataset_cross_attention}
Assume the follow parameters,
\begin{itemize}
    \item Let $a_0 \in (0, 0.1)$
    \item Let $a_1 > 1$
    %\item Let $b$
\end{itemize}
Given two sets of datasets ${\cal D}_0$ and ${\cal D}_1$.
\begin{itemize}
    \item For each $\{ A_1, A_2, A_3 \} \in {\cal D}_0$, we have
    \begin{itemize}
        \item $A_1 \neq A_2 = A_3 \in \R^{n \times d}$
        \item Assume $n = (d-1)t$ where $t$ is a positive integer
        \item There are two special index $j_2 \in [n]$ and $j_3 \in [n]$
        \item The first column of $A_1$ is, $j_2$-th location is $a_0 \log n$ and rest locations are $0$
        \item For the second column to $d-1$ columns in $A_1$, all the entries are $\log n$ %it is 
       % $\begin{bmatrix} I_{d-1} \\ I_{d-1} \\ \vdots \\ I_{d-1} \end{bmatrix} \cdot \log n$
        
        \item the first column of $A_2$ is $e_{j_3}$ (this is a one-sparse vector where only $j_3$-th location is $1$)
        \item the second column to $d-1$ columns in $A_2$ are $\begin{bmatrix} I_{d-1} \\ I_{d-1} \\ \vdots \\ I_{d-1} \end{bmatrix}$
    \end{itemize}
    \item For each $\{ A_1, A_2, A_3 \} \in {\cal D}_1$ , we have
    \begin{itemize}
        \item $A_1 \neq A_2 = A_3 \in \R^{n \times d}$
        \item Assume $n = (d-1)t$ where $t$ is a positive integer
        \item There are two special index $j_2 \in [n]$ and $j_3 \in [n]$
        \item The first column of $A_1$ is, $j_2$-th location is $a_1 \cdot \log n$ and rest locations are $0$
        \item For the second column to $d-1$ columns in $A_1$, all the entries are $\log n$ %it is 
        %$\begin{bmatrix} I_{d-1} \\ I_{d-1} \\ \vdots \\ I_{d-1} \end{bmatrix} \cdot \log n$
        \item the first column of $A_2$ is $e_{j_3} \in \R^n$ (this is a one-sparse vector where only $j_3$-th location is $1$)
        \item the second column to $d-1$ columns in $A_2$ are $\begin{bmatrix} I_{d-1} \\ I_{d-1} \\ \vdots \\ I_{d-1} \end{bmatrix}$
    \end{itemize}
\end{itemize}
\end{definition}

\subsection{Dataset 1 with \texorpdfstring{$F_{\exp}$}{}}\label{sec:app:data1_exp_cross}
In Section~\ref{sec:ds_1_u_f_exp_cross} we analyse the property of dataset 1 with respect to function $u_{\exp}$ and $f_{\exp}$. In Section~\ref{sec:ds_1_c_exp_cross} we analyse the property of dataset 1 with respect to function $c_{\exp}$. In Section~\ref{sec:ds_1_y_exp_cross} we analyse the property of dataset 1 with respect to function $c_{\exp}$ with random signs. In Section~\ref{sec:ds_1_F_exp_cross} we show the property of dataset 1 with respect to the output of $F_{\exp}$.

\subsubsection{Dataset 1 Property when applying function \texorpdfstring{$u_{\exp}$}{} and \texorpdfstring{$f_{\exp}$}{}}
\label{sec:ds_1_u_f_exp_cross}

\begin{lemma}\label{lem:cross_dataset_1_f_exp}
If the following conditions hold
\begin{itemize}
    \item Let $\{ A_1, A_2, A_3 \}$ from dataset ${\cal D}_1$ (see Definition~\ref{def:dataset_cross_attention})
    \item Let $QK^\top = I_d $
\end{itemize} 
Then, for $u_{\exp}(\A,x)_{j_0,j_1}$ and $f_{\exp}(\A,x)_{j_0,j_1}$ entry we have
\begin{itemize}
    \item {\bf Part 1.} For $j_0 = j_2$, 
    \begin{itemize}
        \item {\bf Part 1a.} For $j_1 = j_3$, then  $u_{\exp}(\A,x)_{j_0,j_1} = n^{1+a_1} $.
        \item {\bf Part 1b.} For $j_1 \neq j_3$, then  $u_{\exp}(\A,x)_{j_0,j_1} = n $, 
    \end{itemize}
    \item {\bf Part 2.} For $j_0 \neq j_2$
    \begin{itemize}
        \item {\bf Part 2a.} For $j_1 = j_3$, then $u_{\exp}(A,x)_{j_0,j_1} = n $.
        \item {\bf Part 2b.} For $j_1 \neq j_3$, then $u_{\exp}(A,x)_{j_0,j_1} = n$.
    \end{itemize}
    \item {\bf Part 3.} For $j_0 = j_2$, 
    \begin{itemize}
    \item {\bf Part 3a.} For $j_1 = j_3$, then  $f_{\exp}(\A,x)_{j_0,j_1} \geq 1/2$ (if $a_1 > 1$)
    \item {\bf Part 3b.} For $j_1 \neq j_3$, then  $f_{\exp}(\A,x)_{j_0,j_1} \leq 1/n$
    \end{itemize}
    \item {\bf Part 4.}For $j_0 = j_2$, 
    \begin{itemize}
        \item {\bf Part 4a.} For $j_0 = j_2$, then  $f_{\exp}(\A,x)_{j_0,j_1} = 1/n$
        \item {\bf Part 4b.} For $j_0 \neq j_2$, then  $f_{\exp}(\A,x)_{j_0,j_1} = 1/n$
    \end{itemize}
\end{itemize}
\end{lemma}
\begin{proof}

{\bf Proof of Part 1.}
{\bf Proof of Part 1a.}

By computing the inner product we know  
\begin{align*}
u_{\exp}(\A,x)_{j_0,j_1} 
= & ~ \exp( a_1 \log n + \log n ) \\
= & ~ \exp( (1+a_1) \log n )  \\
= & ~ n^{1+a_1}
\end{align*}

{\bf Proof of Part 1b.}

We have
\begin{align*}
u_{\exp}(\A,x)_{j_0,j_1} 
= & ~ \exp(\log n) \\
= & ~ n
\end{align*} 

{\bf Proof of Part 2.}
Proofs are same as Part 1b.

{\bf Proof of Part 3.}

{\bf Proof of Part 3a.}

We can show
\begin{align*}
f_{\exp}(\A,x)_{j_0,j_1} \geq & ~ \frac{n^{1+a_1}}{ n^{1+a_1} + (n-1) n } \\
\geq & ~ \frac{1}{2}
\end{align*}

{\bf Proof of Part 3b.}

We can show
\begin{align*}
f_{\exp}(\A,x)_{j_0,j_1} \leq & ~ \frac{n}{ n^{1+a_1} + (n-1) \cdot n } \\
\leq & ~ \frac{n}{n^2} \\
\leq & ~ \frac{1}{n}
\end{align*}

{\bf Proof of Part 4.}

We can show
\begin{align*}
f_{\exp}(\A,x)_{j_0,j_1} = & ~ \frac{n}{ n \cdot n } \\
= & ~ \frac{1}{n}
\end{align*}

\end{proof}

\subsubsection{Dataset 1 Property when applying function \texorpdfstring{$c_{\exp}$}{}}
\label{sec:ds_1_c_exp_cross}

\begin{lemma}\label{lem:cross_dataset_1_c_exp}
If the following conditions hold
\begin{itemize}
    \item Let $\{A_1, A_2, A_3 \}$ from dataset ${\cal D}_1$ (see Definition~\ref{def:dataset_cross_attention})
    \item Let $QK^\top = I_d$
    \item Let $V = I_d$
    \item Let $n = t(d-1)$
\end{itemize}
Then for $c_{\exp}(\A,x)_{j_0,i_0} $ entry we have
\begin{itemize}
    \item {\bf Part 1.} For $j_0=j_2$ and $i_0 = 1$, we have $ c_{\exp}(\A,x)_{j_0,i_0}  \geq \frac{1}{2} $
    \item {\bf Part 2.} For $j_0=j_2$ and $i_0 \in \{2,\cdots,d\}$
    \begin{itemize}
        \item There is only one $i_0$, we have $c_{\exp}(\A,x)_{j_0,i_0} \geq \frac{1}{2}$
        \item For the rest of $i_0$, we have $c_{\exp}(\A,x)_{j_0,i_0} \leq \frac{t}{n } $
    \end{itemize} 
    \item {\bf Part 3.} For $j_0 \neq j_2$ and $i_0 = 1$, we have $ c_{\exp}(\A,x)_{j_0,i_0} \leq \frac{1}{n}  $
    \item {\bf Part 4.} For $j_0\neq j_2$ and $i_0 \in \{2,\cdots,d\}$, we have $ c_{\exp}(\A,x)_{j_0,i_0} \leq \frac{t}{n}  $ 
\end{itemize}
\end{lemma}
\begin{proof}

{\bf Proof of Part 1.}

It follows from Part 3 of Lemma~\ref{lem:cross_dataset_1_f_exp} and Type I column in $A_3$ (Definition~\ref{def:dataset_cross_attention}).

{\bf Proof of Part 2.}

It follows from Part 3 of Lemma~\ref{lem:cross_dataset_1_f_exp} and Type II column in $A_3$ (Definition~\ref{def:dataset_cross_attention}).

{\bf Proof of Part 3.}

It follows from Part 4 of Lemma~\ref{lem:cross_dataset_1_f_exp} and Type I column in $A_3$ (Definition~\ref{def:dataset_cross_attention}).

{\bf Proof of Part 4.}

It follows from Part 4 of Lemma~\ref{lem:cross_dataset_1_f_exp} and Type II column in $A_3$ (Definition~\ref{def:dataset_cross_attention}).

\end{proof}

\subsubsection{Dataset 1 Property when applying function \texorpdfstring{$c_{\exp}$}{} with random signs}
\label{sec:ds_1_y_exp_cross}

\begin{lemma}\label{lem:cross_dataset_1_random_exp}
If the following conditions hold
\begin{itemize}
    \item Let $\{A_1, A_2,A_3,\}$ be from dataset ${\cal D}_1$ (see Definition~\ref{def:dataset_cross_attention})
    \item Let $t \sqrt{d} = o(n^{0.99})$ (since $t(d-1) = n$, then this implies $d = \omega(n^{0.02})$)
\end{itemize}
Then, for each random string $\sigma \in \{-1,+1\}^d$, we have
\begin{itemize}
    \item {\bf Part 1.} If $j_0 = j_3$, then $\Pr[ \langle c_{\exp}(\A,x)_{j_0} , \sigma \rangle \geq 1 - 0.1 ] = 1/10$
    \item {\bf Part 2.} If $j_0 \neq j_3$, then $\Pr[ |\langle c_{\exp}(\A,x)_{j_0} , \sigma \rangle| < 0.5 ] = 1 - \delta/\poly(n)$
\end{itemize}
\end{lemma}
\begin{proof}
{\bf Proof of Part 1.}
It follows from Part 1,2,3 of Lemma~\ref{lem:cross_dataset_1_c_exp}, random sign distribution.

 There are two large coordinates, both of them multiplying with a positive sign, the chance of that is probability $1/4$.

Thus,
 \begin{align*}
 \Pr[ \langle c_{\exp}(\A,x)_{j_0} , \sigma \rangle \geq \frac{1}{2} + \frac{1}{2} - 0.1 ] \geq 1/4 - \delta/\poly(n) \geq 1/10
 \end{align*}

{\bf Proof of Part 2.}
It follows from Part 4,5,6 of Lemma~\ref{lem:cross_dataset_1_c_exp} and Hoeffding inequality (Lemma~\ref{lem:hoeffding_bound}).

By hoeffding inequality (Lemma~\ref{lem:hoeffding_bound}), we know that
\begin{align*}
| \langle c_{\exp}(\A,x)_{j_0} , \sigma \rangle  | \leq & ~ O( \sqrt{\log (n/\delta)} ) \cdot \frac{ t \sqrt{d} }{ n }   \\
\leq & ~ 0.1
\end{align*}
with probability at least $1-\delta/\poly(n)$. Here the last step due to $t\sqrt{d} = o(n^{0.99})$ and $\poly(\log n) \leq n^{0.01}$.
\end{proof}

\subsubsection{Dataset 1 Property when applying function \texorpdfstring{$F_{\exp}$}{} }
\label{sec:ds_1_F_exp_cross}

\begin{theorem}
If the following conditions hold
\begin{itemize}
    \item Let $d \in [ \omega(n^{0.02}) , n ]$
    \item Let $\tau = 0.9$
    \item Let $m = O(\log (n/\delta))$
    \item For any $\{A_1,A_2,A_3\}$ from ${\cal D}_1$ (Definition~\ref{def:dataset_cross_attention})
    \item Let $F_{\exp}(A_1,A_2,A_3):=\phi( \sum_{j_0=1}^n \sum_{j_1=1}^m \phi_{\tau}( \langle c_{\exp}(\A,x) , y_{j_1} ) )$
\end{itemize}
Then we have
\begin{itemize}
    \item With high probability $1-\delta/\poly(n)$, $F_{\exp}(A_1,A_2,A_3) > 0$
\end{itemize}
\end{theorem}
\begin{proof}
It follows from using Lemma~\ref{lem:cross_dataset_1_random_exp}.
\end{proof}

%\newpage
\subsection{Dataset 0 with \texorpdfstring{$F_{\exp}$}{}}\label{sec:app:data0_exp_cross}
In Section~\ref{sec:ds_0_u_f_exp_cross} we analyse the property of dataset 0 with respect to function $u_{\exp}$ and $f_{\exp}$. In Section~\ref{sec:ds_0_c_exp_cross} we analyse the property of dataset 0 with respect to function $c_{\exp}$. In Section~\ref{sec:ds_0_y_exp_cross} we analyse the property of dataset 0 with respect to function $c_{\exp}$ with random signs. In Section~\ref{sec:ds_0_F_exp_cross} we show the property of dataset 0 with respect to the output of $F_{\exp}$. 

\subsubsection{Dataset 0 Property when applying function \texorpdfstring{$u_{\exp}$}{} and \texorpdfstring{$f_{\exp}$}{}}
\label{sec:ds_0_u_f_exp_cross}

\begin{lemma}\label{lem:cross_dataset_0_f_exp}
If the following conditions hold
\begin{itemize}
    \item Let $\{ A_1, A_2, A_3 \}$ from dataset ${\cal D}_1$ (see Definition~\ref{def:dataset_cross_attention})
    \item Let $QK^\top = I_d $
\end{itemize} 
Then, for $u_{\exp}(\A,x)_{j_0,j_1}$ and $f_{\exp}(\A,x)_{j_0,j_1}$ entry we have
\begin{itemize}
     \item {\bf Part 1.} For $j_0 = j_2$, 
    \begin{itemize}
        \item {\bf Part 1a.} For $j_1 = j_3$, then  $u_{\exp}(\A,x)_{j_0,j_1} = n^{1+a_0} $.
        \item {\bf Part 1b.} For $j_1 \neq j_3$, then  $u_{\exp}(\A,x)_{j_0,j_1} = n $, 
    \end{itemize}
    \item {\bf Part 2.} For $j_0 \neq j_2$
    \begin{itemize}
        \item {\bf Part 2a.} For $j_1 = j_3$, then $u_{\exp}(A,x)_{j_0,j_1} = n $.
        \item {\bf Part 2b.} For $j_1 \neq j_3$, then $u_{\exp}(A,x)_{j_0,j_1} = n$.
    \end{itemize}
    \item {\bf Part 3.} For $j_0 = j_2$, 
    \begin{itemize}
    \item {\bf Part 3a.} For $j_1 = j_3$, then  $f_{\exp}(\A,x)_{j_0,j_1} \leq 1/n^{1-a_0}$ (if $a_0 < 1$)
    \item {\bf Part 3b.} For $j_1 \neq j_3$, then  $f_{\exp}(\A,x)_{j_0,j_1} \leq 1/n$
    \end{itemize}
    \item {\bf Part 4.}
    \begin{itemize}
        \item {\bf Part 4a.} For $j_0 = j_2$, then  $f_{\exp}(\A,x)_{j_0,j_1} = 1/n$
        \item {\bf Part 4b.} For $j_0 \neq j_2$, then  $f_{\exp}(\A,x)_{j_0,j_1} = 1/n$
    \end{itemize}
\end{itemize}
\end{lemma}
\begin{proof}

{\bf Proof of Part 1.}

{\bf Proof of Part 1a.}

By computing the inner product we know  
\begin{align*}
u_{\exp}(\A,x)_{j_0,j_1} 
= & ~ \exp( a_0 \log n + \log n ) \\
= & ~ \exp( (1+a_0) \log n )  \\
= & ~ n^{1+a_0}
\end{align*}

{\bf Proof of Part 1b.} 

We can show
\begin{align*}
u_{\exp}(\A,x)_{j_0,j_1} 
= & ~ \exp(\log n) \\
= & ~ n
\end{align*}

{\bf Proof of Part 2.} 

We can show
\begin{align*}
u_{\exp}(\A,x)_{j_0,j_1} 
= & ~ \exp(\log n) \\
= & ~ n
\end{align*}

{\bf Proof of Part 3.}

{\bf Proof of Part 3a.}
We can show
\begin{align*}
f_{\exp}(\A,x)_{j_0,j_1} \leq & ~ \frac{n^{1+a_0}}{ n^{1+a_0} + (n-1) n } \\
\leq & ~ \frac{ n^{1+a_0} }{n^2} \\
\leq & ~ \frac{1}{1-a_0}
\end{align*}

{\bf Proof of Part 3b.}

We can show
\begin{align*}
f_{\exp}(\A,x)_{j_0,j_1} \leq & ~ \frac{n}{ n^{1+a_0} + (n-1) \cdot n } \\
\leq & ~ \frac{n}{n^2} \\\
= & ~ \frac{1}{n}
\end{align*}

{\bf Proof of Part 4.}

We can show
\begin{align*}
f_{\exp}(\A,x)_{j_0,j_1} = & ~ \frac{n }{ n \cdot n } \\
= & ~  1/n
\end{align*}

\end{proof}

\subsubsection{Dataset 0 Property when applying function \texorpdfstring{$c_{\exp}$}{}}
\label{sec:ds_0_c_exp_cross}

\begin{lemma}\label{lem:cross_dataset_0_c_exp}
If the following conditions hold
\begin{itemize}
    \item Let $\{A_1, A_2, A_3 \}$ from dataset ${\cal D}_0$ (see Definition~\ref{def:dataset_cross_attention})
    \item Let $QK^\top = I_d$
    \item Let $V = I_d$
    \item Let $n = t(d-1)$
\end{itemize}
Then for $c_{\exp}(\A,x)_{j_0,i_0} $ entry we have
\begin{itemize}
    \item {\bf Part 1.} For $j_0=j_3$ and $i_0 = 1$, we have $ c_{\exp}(\A,x)_{j_0,i_0}  \leq \frac{1}{n^{1-a_0}}  $
    \item {\bf Part 2.} For $j_0=j_3$ and $i_0 \in \{2,\cdots,d\}$,
        we have $c_{\exp}(\A,x)_{j_0,i_0} \leq \frac{t}{n^{1-a_0}}    $  
    \item {\bf Part 4.} For $j_0 \neq j_3$ and $i_0 = 1$, we have $ c_{\exp}(\A,x)_{j_0,i_0} \leq \frac{1}{n}  $
    \item {\bf Part 5.} For $j_0\neq j_3$ and $i_0 \in \{2,\cdots,d\}$, we have $ c_{\exp}(\A,x)_{j_0,i_0} \leq \frac{t}{n } $ 
\end{itemize}
\end{lemma}
\begin{proof}
{\bf Proof of Part 1.}

It follows from Part 3 of Lemma~\ref{lem:cross_dataset_0_f_exp}, and Type I column in $A_3$.

{\bf Proof of Part 2.}

It follows from Part 3 of Lemma~\ref{lem:cross_dataset_0_f_exp}, and Type II column in $A_3$.

We know that each entry in $f_{\exp}(\A,x)_{j_0}$ is at least $0$ and is at most $\frac{1}{n^{1-a_0}}$.

We know that each entry in $(A_3 V)_{i_0} \in \R^n$ is at least $0$ and at most $1$ and it is $t$-sparse.

Thus,
\begin{align*}
\langle f_{\exp}(\A,x)_{j_0}, (A_3 V)_{i_0} \rangle \leq \frac{1}{n^{1-a_0}}  
\end{align*}

{\bf Proof of Part 3.}

It follows from Part 4 of Lemma~\ref{lem:cross_dataset_0_f_exp}, and Type I column in $A_3$.

{\bf Proof of Part 4.}

It follows from Part 4 of Lemma~\ref{lem:cross_dataset_0_f_exp}, and Type II column in $A_3$.

\end{proof}

\subsubsection{Dataset 0 Property when applying function \texorpdfstring{$c_{\exp}$}{} with random signs}
\label{sec:ds_0_y_exp_cross}

\begin{lemma}\label{lem:cross_dataset_0_random_exp}
If the following conditions hold
\begin{itemize}
    \item Let $\{A_1, A_2,A_3,\}$ be from dataset ${\cal D}_0$ (see Definition~\ref{def:dataset_cross_attention})
    \item Let $t \sqrt{d} = o(n^{1-a_0-0.01})$ (since $t(d-1) = n$, then this implies $d = \omega(n^{2a_0 + 0.02})$)
\end{itemize}
Then, for each random string $\sigma \in \{-1,+1\}^d$, we have
\begin{itemize}
    \item {\bf Part 1.} If $j_0 = j_3$, then $\Pr[ |\langle c_{\exp}(\A,x)_{j_0} , \sigma \rangle| < 0.5 ] = 1-\delta/\poly(n)$
    \item {\bf Part 2.} If $j_0 \neq j_3$, then $\Pr[ | \langle c_{\exp}(\A,x)_{j_0} , \sigma \rangle | < 0.5 ] = 1 - \delta/\poly(n)$
\end{itemize}
\end{lemma}
\begin{proof}
{\bf Proof of Part 1.}
It follows from Part 1,2,3 of Lemma~\ref{lem:cross_dataset_1_c_exp}, random sign distribution.

{\bf Proof of Part 2.}
It follows from Part 4,5,6 of Lemma~\ref{lem:cross_dataset_1_c_exp} and Hoeffding inequality.

By hoeffding inequality, we know that
\begin{align*}
| \langle c_{\exp}(\A,x)_{j_0} , \sigma \rangle  | \leq & ~ O( \sqrt{\log (n/\delta)} ) \cdot \frac{ t \sqrt{d} }{ n^{1-a_0} }  \\
\leq & ~ 0.5 
\end{align*}
with probability at least $1-\delta/\poly(n)$. Here the last step due to $t\sqrt{d} = o(n^{1-a_0-0.01})$ and and $\poly(\log n) \leq n^{0.01}$.
\end{proof}

\subsubsection{Dataset 0 Property when applying function \texorpdfstring{$F_{\exp}$}{} }
\label{sec:ds_0_F_exp_cross}

\begin{theorem}
If the following conditions hold
\begin{itemize}
    \item Let $d \in [ \omega(n^{2a_0+0.02}) , n ]$
    \item Let $a_0 \in (0,0.1)$
    \item Let $\tau = 0.9$
    \item Let $m = O(\log (n/\delta))$
    \item For any $\{A_1,A_2,A_3\}$ from ${\cal D}_1$ (Definition~\ref{def:dataset_cross_attention})
    \item Let $F_{\exp}(A_1,A_2,A_3):=\phi( \sum_{j_0=1}^n \sum_{j_1=1}^m \phi_{\tau}( \langle c_{\exp}(\A,x) , y_{j_1} ) )$
\end{itemize}
Then we have
\begin{itemize}
    \item With high probability $1-\delta/\poly(n)$, $F_{\exp}(A_1,A_2,A_3) = 0$
\end{itemize}
\end{theorem}
\begin{proof}
It follows from using Lemma~\ref{lem:cross_dataset_1_random_exp}.
\end{proof}

%\newpage

\subsection{Dataset 1 with \texorpdfstring{$F_{\lin}$}{}}\label{sec:app:data1_lin_cross}
In Section~\ref{sec:ds_1_u_f_lin_cross} we analyse the property of dataset 1 with respect to function $u_{\lin}$ and $f_{\lin}$. In Section~\ref{sec:ds_1_c_lin_cross} we analyse the property of dataset 1 with respect to function $c_{\lin}$. In Section~\ref{sec:ds_1_y_lin_cross} we analyse the property of dataset 1 with respect to function $c_{\lin}$ with random signs. In Section~\ref{sec:ds_1_F_lin_cross} we show the property of dataset 1 with respect to the output of $F_{\lin}$. 

\subsubsection{Dataset 1 Property when applying function \texorpdfstring{$u_{\lin}$}{} and \texorpdfstring{$f_{\lin}$}{}}
\label{sec:ds_1_u_f_lin_cross}

\begin{lemma}\label{lem:cross_dataset_1_f_lin}
If the following conditions hold
\begin{itemize}
    \item Let $\{ A_1, A_2, A_3 \}$ from dataset ${\cal D}_1$ (see Definition~\ref{def:dataset_cross_attention})
    \item Let $QK^\top = I_d $
\end{itemize} 
Then, for $u_{\lin}(\A,x)_{j_0,j_1}$ and $f_{\lin}(\A,x)_{j_0,j_1}$ entry we have
\begin{itemize}
    \item {\bf Part 1.} For $j_0 = j_2$, and $j_1 = j_3$, then  $u_{\lin}(\A,x)_{j_0,j_1} = (1+a_1) \log n $.
    \item {\bf Part 2.} For $j_0 \neq j_2$ or $j_1 \neq j_3$, then $u_{\lin}(A,x)_{j_0,j_1} = \log n $.
    \item {\bf Part 3.} For $j_0 = j_2$, and $j_1 = j_3$, then  $f_{\lin}(\A,x)_{j_0,j_1} \leq  (1+a_1)/n$.
    \item {\bf Part 4.} For $j_0 = j_2$, and $j_1 \neq j_3$, then  $f_{\lin}(\A,x)_{j_0,j_1} \leq 1/n$.
    \item {\bf Part 5.} For $j_0 \neq j_2$, then  $f_{\lin}(\A,x)_{j_0,j_1} = 1/n$.
\end{itemize}
\end{lemma}
\begin{proof}

{\bf Proof of Part 1.}

By computing the inner product we know 
\begin{align*}
u_{\lin}(\A,x)_{j_0,j_1} 
= & ~ a_1 \log n + \log n  \\
= & ~ (1+a_1) \log n  
\end{align*}
where the last step follows from simple algebra.

{\bf Proof of Part 2.}

We can show
\begin{align*}
u_{\lin}(\A,x)_{j_0,j_1} 
=  \log n
\end{align*}

{\bf Proof of Part 3.}

We can show
\begin{align*}
f_{\lin}(\A,x)_{j_0,j_1} \leq & ~ \frac{ (1+a_1) \log n}{ (1+a_1) \log n + (n-1) \cdot \log n } \\
\leq & ~ \frac{ 1+a_1 }{n}
\end{align*}

{\bf Proof of Part 4.}

We can show
\begin{align*}
f_{\lin}(\A,x)_{j_0,j_1} \leq & ~ \frac{ \log n }{ (1+a_1) \log n + (n-1) \cdot \log n } \\
\leq & ~ \frac{1}{n} 
\end{align*}

{\bf Proof of Part 5.}

We can show
\begin{align*}
f_{\lin}(\A,x)_{j_0,j_1} = & ~ \frac{ \log n }{ n \cdot \log n } \\
= & ~ \frac{1}{n}
\end{align*}

\end{proof}

\subsubsection{Dataset 1 Property when applying function \texorpdfstring{$c_{\lin}$}{}}
\label{sec:ds_1_c_lin_cross}

\begin{lemma}\label{lem:cross_dataset_1_c_lin}
If the following conditions hold
\begin{itemize}
    \item Let $\{A_1, A_2, A_3 \}$ from dataset ${\cal D}_0$ (see Definition~\ref{def:dataset_cross_attention})
    \item Let $QK^\top = I_d$
    \item Let $V = I_d$
    \item Let $n = t(d-1)$
\end{itemize}
Then for $c_{\lin}(\A,x)_{j_0,i_0} $ entry we have
\begin{itemize}
    \item {\bf Part 1.} For $j_0=j_3$ and $i_0 = 1$, we have $ c_{\lin}(\A,x)_{j_0,i_0}  \leq \frac{1+a_1}{n}  $
    \item {\bf Part 2.} For $j_0=j_3$ and $i_0 \in \{2,\cdots,d\}$,
        we have $c_{\lin}(\A,x)_{j_0,i_0} \leq \frac{(1+a_1)t}{n}   $ 
      
    \item {\bf Part 3.} For $j_0 \neq j_3$ and $i_0 = 1$, we have $ c_{\lin}(\A,x)_{j_0,i_0} \leq \frac{1}{n}  $
    \item {\bf Part 4.} For $j_0\neq j_3$ and $i_0 \in \{2,\cdots,d\}$, we have $ c_{\lin}(\A,x)_{j_0,i_0} \leq \frac{t}{n} $ 
\end{itemize}
\end{lemma}
\begin{proof}
{\bf Proof of Part 1.}

It follows from Part 3 and Part 4 of Lemma~\ref{lem:cross_dataset_1_f_lin}, and Type I column in $A_3$.

{\bf Proof of Part 2.}

It follows from Part 3 and Part 4 of Lemma~\ref{lem:cross_dataset_1_f_lin}, and Type II column in $A_3$.

We know that each entry in $f_{\lin}(\A,x)_{j_0}$ is at least $0$ and is at most $\frac{(1+a_1)}{n}$.

We know that each entry in $(A_3 V)_{i_0} \in \R^n$ is at least $0$ and at most $1$ and it is $t$-sparse.

Thus,
\begin{align*}
\langle f_{\lin}(\A,x)_{j_0}, (A_3 V)_{i_0} \rangle \leq \frac{(1+a_1)t}{n}  
\end{align*}

{\bf Proof of Part 3.}

It follows from Part 5 of Lemma~\ref{lem:cross_dataset_1_f_lin}, and Type I column in $A_3$.

{\bf Proof of Part 4.}

It follows from Part 5 of Lemma~\ref{lem:cross_dataset_1_f_lin}, and Type II column in $A_3$.

\end{proof}

\subsubsection{Dataset 1 Property when applying function \texorpdfstring{$c_{\lin}$}{} with random signs}
\label{sec:ds_1_y_lin_cross}

\begin{lemma}\label{lem:cross_dataset_1_random_lin}
If the following conditions hold
\begin{itemize}
    \item Let $\{A_1, A_2,A_3,\}$ be from dataset ${\cal D}_0$ (see Definition~\ref{def:dataset_cross_attention})
    \item Let $t \sqrt{d} = o(n^{0.99})$ (since $t(d-1) = n$, then this implies $d = \omega(n^{0.02})$)
\end{itemize}
Then, for each random string $\sigma \in \{-1,+1\}^d$, we have
\begin{itemize}
    \item {\bf Part 1.} If $j_0 = j_3$, then $\Pr[ |\langle c_{\lin}(\A,x)_{j_0} , \sigma \rangle| < 0.5 ] = 1-\delta/\poly(n)$
    \item {\bf Part 2.} If $j_0 \neq j_3$, then $\Pr[ | \langle c_{\lin}(\A,x)_{j_0} , \sigma \rangle | < 0.5 ] = 1 - \delta/\poly(n)$
\end{itemize}
\end{lemma}
\begin{proof}
{\bf Proof of Part 1.}
It follows from Part 1,2,3 of Lemma~\ref{lem:cross_dataset_1_c_lin}, random sign distribution.

By hoeffding inequality (Lemma~\ref{lem:hoeffding_bound}), we know that
\begin{align*}
| \langle c_{\lin}(\A,x)_{j_0} , \sigma \rangle   | \leq & ~ O( \sqrt{\log (n/\delta)} ) \cdot \frac{ (1+a_1) t \sqrt{d} }{ n }   \\
\leq & ~ 0.5  
\end{align*}
with probability at least $1-\delta/\poly(n)$.  Here $a_1 = O(1)$.

{\bf Proof of Part 2.}
It follows from Part 4,5,6 of Lemma~\ref{lem:cross_dataset_1_c_lin} and Hoeffding inequality (Lemma~\ref{lem:hoeffding_bound}).

By hoeffding inequality (Lemma~\ref{lem:hoeffding_bound}), we know that
\begin{align*}
| \langle c_{\lin}(\A,x)_{j_0} , \sigma \rangle   | \leq & ~ O( \sqrt{\log (n/\delta)} ) \cdot \frac{ 2 t \sqrt{d} }{ n }   \\
\leq & ~ 0.5  
\end{align*}
with probability at least $1-\delta/\poly(n)$. Here the last step due to $t\sqrt{d} = o(n^{0.99})$ and and $\poly(\log n) \leq n^{0.01}$.
\end{proof}

\subsubsection{Dataset 1 Property when applying function \texorpdfstring{$F_{\lin}$}{} }
\label{sec:ds_1_F_lin_cross}

\begin{theorem}
If the following conditions hold
\begin{itemize}
    \item Let $d \in [ \omega(n^{0.02}) , n ]$
    \item Let $\tau = 0.9$
    \item Let $a_1 = O(1)$
    \item Let $m = O(\log (n/\delta))$
    \item For any $\{A_1,A_2,A_3\}$ from ${\cal D}_1$ (Definition~\ref{def:dataset_cross_attention})
    \item Let $F_{\lin}(A_1,A_2,A_3):=\phi( \sum_{j_0=1}^n \sum_{j_1=1}^m \phi_{\tau}( \langle c_{\lin}(\A,x) , y_{j_1} ) )$
\end{itemize}
Then we have
\begin{itemize}
    \item With high probability $1-\delta/\poly(n)$, $F_{\lin}(A_1,A_2,A_3) = 0$
\end{itemize}
\end{theorem}
\begin{proof}
It follows from using Lemma~\ref{lem:cross_dataset_1_random_exp}.
\end{proof}

%\newpage
\subsection{Dataset 0 with \texorpdfstring{$F_{\lin}$}{}}\label{sec:app:data0_lin_cross}
In Section~\ref{sec:ds_0_u_f_lin_cross} we analyse the property of dataset 0 with respect to function $u_{\lin}$ and $f_{\lin}$. In Section~\ref{sec:ds_0_c_lin_cross} we analyse the property of dataset 0 with respect to function $c_{\lin}$. In Section~\ref{sec:ds_0_y_lin_cross} we analyse the property of dataset 0 with respect to function $c_{\lin}$ with random signs. In Section~\ref{sec:ds_0_F_lin_cross} we show the property of dataset 0 with respect to the output of $F_{\lin}$.

\subsubsection{Dataset 0 Property when applying function \texorpdfstring{$u_{\lin}$}{} and \texorpdfstring{$f_{\lin}$}{}}
\label{sec:ds_0_u_f_lin_cross}

\begin{lemma}\label{lem:cross_dataset_0_f_lin}
If the following conditions hold
\begin{itemize}
    \item Let $\{ A_1, A_2, A_3 \}$ from dataset ${\cal D}_1$ (see Definition~\ref{def:dataset_cross_attention})
    \item Let $QK^\top = I_d $
\end{itemize} 
Then, for $u_{\lin}(\A,x)_{j_0,j_1}$ and $f_{\lin}(\A,x)_{j_0,j_1}$ entry we have
\begin{itemize}
    \item {\bf Part 1.} For $j_0 = j_3$, and $j_1 = j_3$, then  $u_{\lin}(\A,x)_{j_0,j_1} = (1+a_0) \log n $.
    \item {\bf Part 2.} For $j_0 \neq j_3$ and $j_1 \neq j_3$, then $u_{\lin}(A,x)_{j_0,j_1} =  \log n  $.
    \item {\bf Part 3.} For $j_0 = j_3$, and $j_1 = j_3$, then  $f_{\lin}(\A,x)_{j_0,j_1} \leq (1+a_0)/n$.
    \item {\bf Part 4.} For $j_0 = j_3$, and $j_1 \neq j_3$, then  $f_{\lin}(\A,x)_{j_0,j_1} \leq 1/n$.
    \item {\bf Part 5.} For $j_0 \neq j_3$, for all $j_1 \in [n]$ then  $f_{\lin}(\A,x)_{j_0,j_1} = 1/n$.
\end{itemize}
\end{lemma}
\begin{proof}

{\bf Proof of Part 1.}

By computing the inner product we know 
\begin{align*}
u_{\lin}(\A,x)_{j_0,j_1} 
= & ~  a_0 \log n + \log n  \\
= & ~  (1+a_0) \log n  
\end{align*}
where the last step follows from simple algebra.

{\bf Proof of Part 2.}

We have
\begin{align*}
u_{\lin}(\A,x)_{j_0,j_1} 
=  \log n
\end{align*} 

{\bf Proof of Part 3.}

We can show
\begin{align*}
f_{\lin}(\A,x)_{j_0,j_1} \leq & ~ \frac{(1+a_0) \log n}{ (1+a_0) \log n + (n-1) \cdot \log n } \\
\leq & ~ \frac{1+a_0}{n}
\end{align*}

{\bf Proof of Part 4.}

We can show
\begin{align*}
f_{\lin}(\A,x)_{j_0,j_1} \leq & ~ \frac{ \log n }{ (1+a_0) \log n + (n-1) \cdot \log n } \\
\leq & ~ \frac{1}{n} 
\end{align*}

{\bf Proof of Part 5.}

We can show
\begin{align*}
f_{\lin}(\A,x)_{j_0,j_1} = & ~ \frac{   \log n }{ n \cdot \log n } \\
= & ~ \frac{1}{n}
\end{align*}

\end{proof}

\subsubsection{Dataset 0 Property when applying function \texorpdfstring{$c_{\lin}$}{}}
\label{sec:ds_0_c_lin_cross}

\begin{lemma}\label{lem:cross_dataset_0_c_lin}
If the following conditions hold
\begin{itemize}
    \item Let $\{A_1, A_2, A_3 \}$ from dataset ${\cal D}_0$ (see Definition~\ref{def:dataset_cross_attention})
    \item Let $QK^\top = I_d$
    \item Let $V = I_d$
    \item Let $n = t(d-2)$
\end{itemize}
Then for $c_{\lin}(\A,x)_{j_0,i_0} $ entry we have
\begin{itemize}
    \item {\bf Part 1.} For $j_0=j_2$ and $i_0 = 1$, we have $ c_{\lin}(\A,x)_{j_0,i_0}  \leq \frac{1+a_0}{n}  $
    \item {\bf Part 2.} For $j_0=j_2$ and $i_0 \in \{2,\cdots,d\}$,
        we have $c_{\lin}(\A,x)_{j_0,i_0} \leq \frac{(1+a_0)t}{n}   $ 
    \item {\bf Part 3.} For $j_0 \neq j_2$ and $i_0 = 1$, we have $ c_{\lin}(\A,x)_{j_0,i_0} \leq \frac{1}{n}  $
    \item {\bf Part 4.} For $j_0\neq j_2$ and $i_0 \in \{2,\cdots,d\}$, we have $ c_{\lin}(\A,x)_{j_0,i_0} \leq \frac{t}{n}  $
    
\end{itemize}
\end{lemma}
\begin{proof}
{\bf Proof of Part 1.}

It follows from Part 3 and Part 4 of Lemma~\ref{lem:cross_dataset_0_f_lin}, and Type I column in $A_3$.

{\bf Proof of Part 2.}

It follows from Part 3 and Part 4 of Lemma~\ref{lem:cross_dataset_0_f_lin}, and Type II column in $A_3$.

We know that each entry in $f_{\lin}(\A,x)_{j_0}$ is at least $0$ and is at most $\frac{1+a_0}{n}$.

We know that each entry in $(A_3 V)_{i_0} \in \R^n$ is at least $0$ and at most $1$ and it is $t$-sparse.

Thus,
\begin{align*}
\langle f_{\lin}(\A,x)_{j_0}, (A_3 V)_{i_0} \rangle \leq \frac{(1+a_0)t}{n}  
\end{align*}

{\bf Proof of Part 3.}

It follows from Part 5 of Lemma~\ref{lem:cross_dataset_0_f_lin}, and Type I column in $A_3$.

{\bf Proof of Part 4.}

It follows from Part 5 of Lemma~\ref{lem:cross_dataset_0_f_lin}, and Type II column in $A_3$.

\end{proof}

\subsubsection{Dataset 0 Property when applying function \texorpdfstring{$c_{\lin}$}{} with random signs}
\label{sec:ds_0_y_lin_cross}

\begin{lemma}\label{lem:cross_dataset_0_random_lin}
If the following conditions hold
\begin{itemize}
    \item Let $\{A_1, A_2,A_3,\}$ be from dataset ${\cal D}_0$ (see Definition~\ref{def:dataset_cross_attention})
    \item Let $t \sqrt{d} = o(n^{0.99})$ (since $t(d-1) = n$, then this implies $d = \omega(n^{0.02})$)
\end{itemize}
Then, for each random string $\sigma \in \{-1,+1\}^d$, we have
\begin{itemize}
    \item {\bf Part 1.} If $j_0 = j_3$, then $\Pr[ |\langle c_{\lin}(\A,x)_{j_0} , \sigma \rangle| < 0.5 ] = 1-\delta/\poly(n)$
    \item {\bf Part 2.} If $j_0 \neq j_3$, then $\Pr[ | \langle c_{\lin}(\A,x)_{j_0} , \sigma \rangle | < 0.5 ] = 1 - \delta/\poly(n)$
\end{itemize}
\end{lemma}
\begin{proof}
{\bf Proof of Part 1.}
It follows from Part 1,2,3 of Lemma~\ref{lem:cross_dataset_0_c_lin}, random sign distribution.

By hoeffding inequality (Lemma~\ref{lem:hoeffding_bound}), we know that
\begin{align*}
| \langle c_{\lin}(\A,x)_{j_0} , \sigma \rangle   | \leq & ~ O( \sqrt{\log (n/\delta)} ) \cdot \frac{ (1+a_0)  t \sqrt{d} }{ n }   \\
\leq & ~ 0.5  
\end{align*}
with probability at least $1-\delta/\poly(n)$. Here $a_0 = O(1)$

{\bf Proof of Part 2.}
It follows from Part 4,5,6 of Lemma~\ref{lem:cross_dataset_0_c_lin} and Hoeffding inequality (Lemma~\ref{lem:hoeffding_bound}).

By hoeffding inequality (Lemma~\ref{lem:hoeffding_bound}), we know that
\begin{align*}
| \langle c_{\lin}(\A,x)_{j_0} , \sigma \rangle   | \leq & ~ O( \sqrt{\log (n/\delta)} ) \cdot \frac{  t \sqrt{d} }{ n }   \\
\leq & ~ 0.5  
\end{align*}
with probability at least $1-\delta/\poly(n)$. Here the last step due to $t\sqrt{d} = o(n^{0.99})$ and and $\poly(\log n) \leq n^{0.01}$.
\end{proof}

\subsubsection{Dataset 0 Property when applying function \texorpdfstring{$F_{\lin}$}{} } 
\label{sec:ds_0_F_lin_cross}

\begin{theorem}
If the following conditions hold
\begin{itemize}
    \item Let $d \in [ \omega(n^{0.02}) , n ]$
    \item Let $\tau = 0.9$
    \item Let $a_0 \in (0,0.1)$
    \item Let $m = O(\log (n/\delta))$
    \item For any $\{A_1,A_2,A_3\}$ from ${\cal D}_0$ (Definition~\ref{def:dataset_cross_attention})
    \item Let $F_{\lin}(A_1,A_2,A_3):=\phi( \sum_{j_0=1}^n \sum_{j_1=1}^m \phi_{\tau}( \langle c_{\lin}(\A,x) , y_{j_1} ) )$
\end{itemize}
Then we have
\begin{itemize}
    \item With high probability $1-\delta/\poly(n)$, $F_{\lin}(A_1,A_2,A_3) = 0$
\end{itemize}
\end{theorem}
\begin{proof}
It follows from using Lemma~\ref{lem:cross_dataset_1_random_exp}.
\end{proof}

\subsection{Experiments for Cross-Attention}

\begin{figure*}[!ht]
    \centering
    \subfloat[Cross-Attention $n$]{\includegraphics[width=0.32\textwidth]{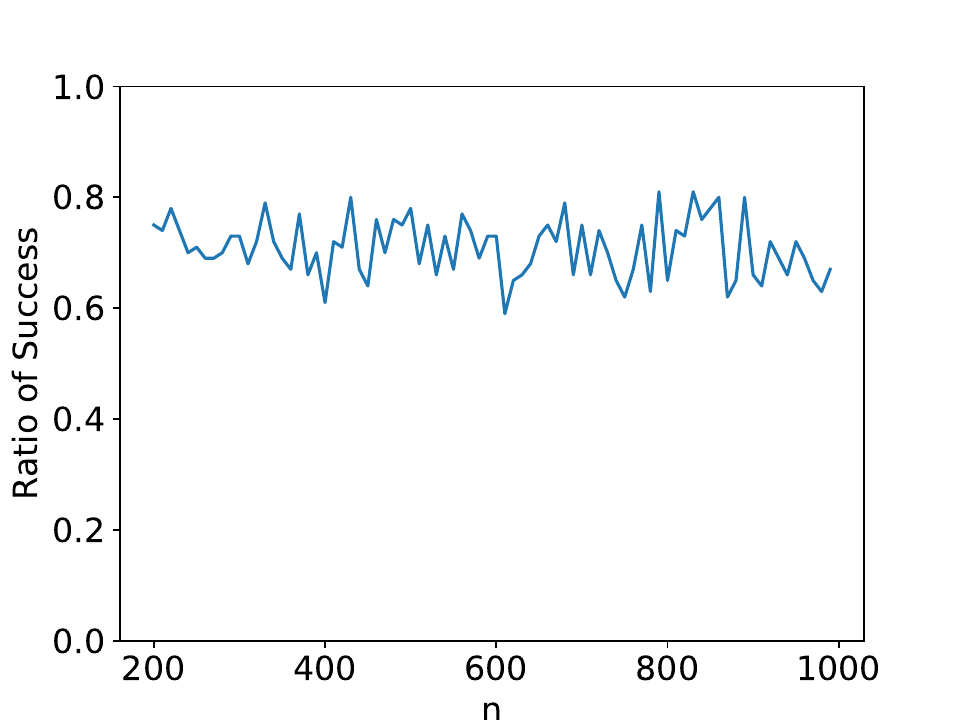}}\label{fig:cross_n}
    \subfloat[Cross-Attention $d$]  {\includegraphics[width=0.32\textwidth]{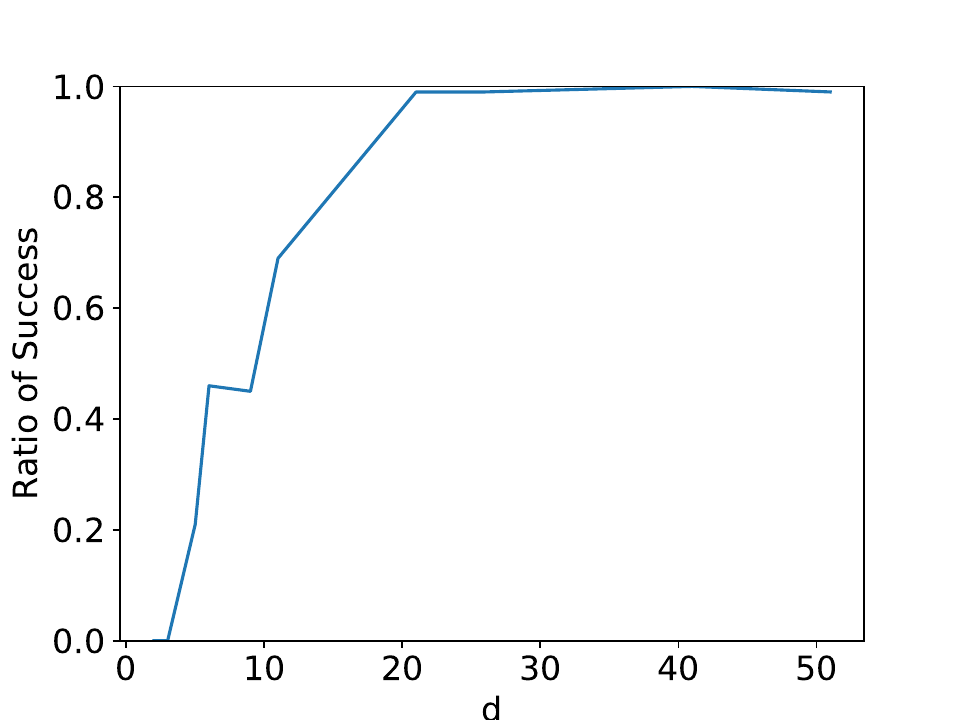}} \label{fig:cross_d}
    \subfloat[Cross-Attention $m$]  {\includegraphics[width=0.32\textwidth]{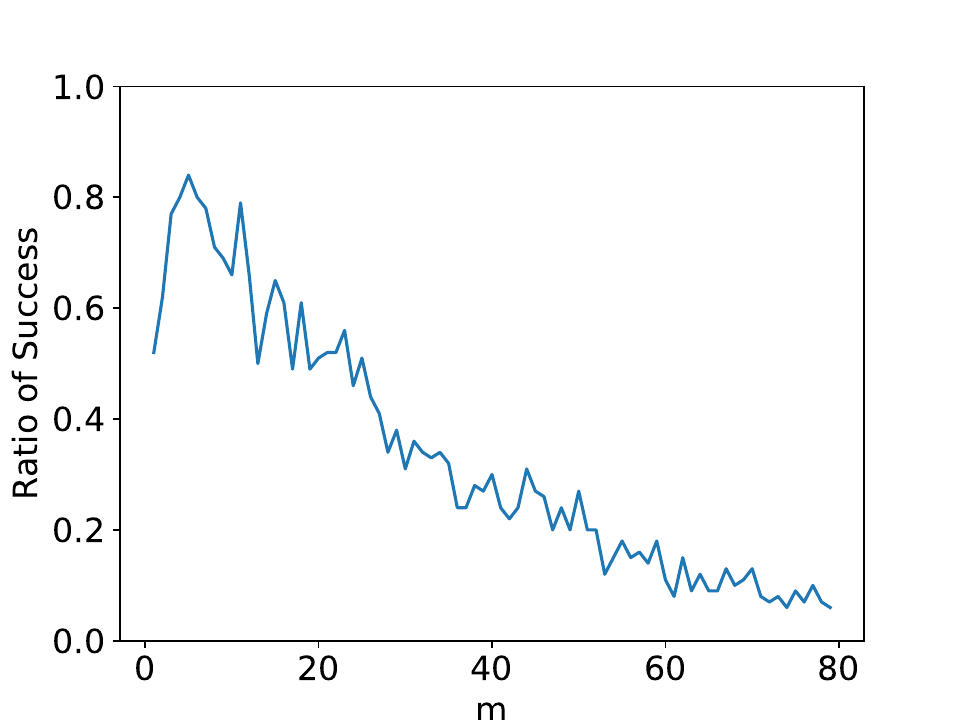}} \label{fig:cross_m}
    \subfloat[Cross-Attention $a_1$]  {\includegraphics[width=0.32\textwidth]{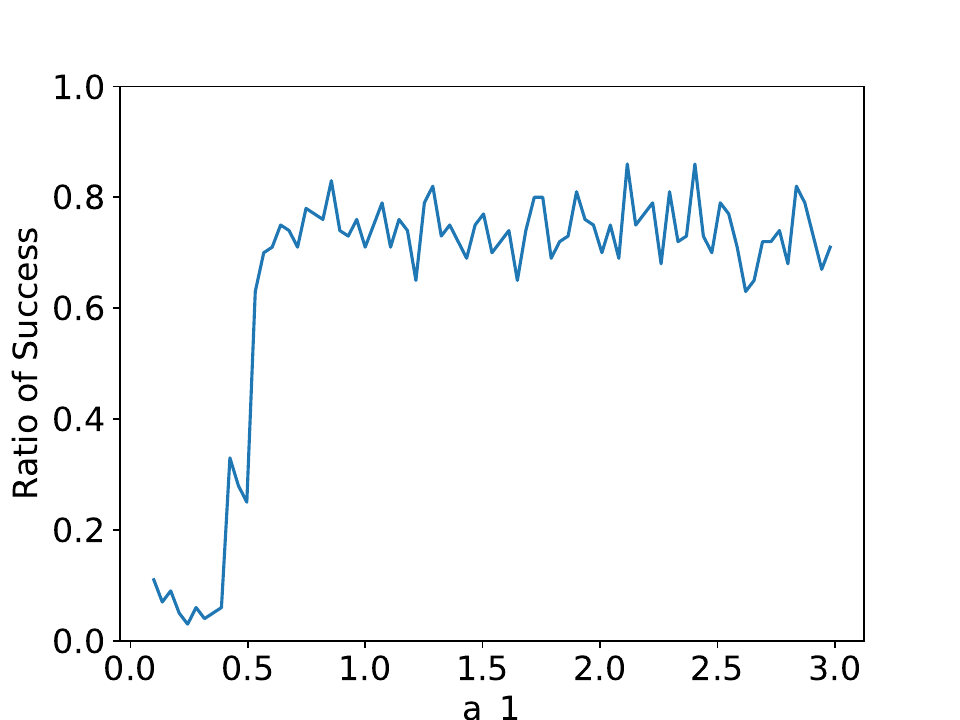}} \label{fig:cross_a1}
    \subfloat[Cross-Attention $a_0$]  {\includegraphics[width=0.32\textwidth]{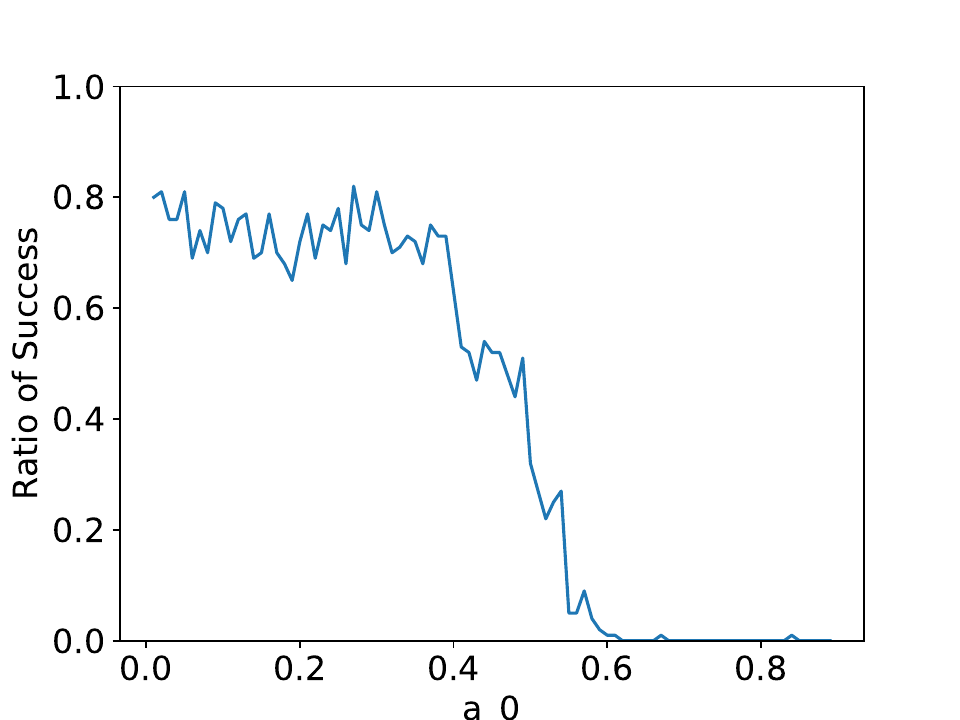}} \label{fig:cross_a0}
    \caption{Cross-Attention} 
    \label{fig:cross}
\end{figure*}

We set the dataset as described in Section~\ref{sec:cross_dataset}. For a pair of inputs $(A_{01}, A_{02}, A_{03}) \in \mathcal{D}_0$ and $(A_{11}, A_{12}, A_{13}) \in \mathcal{D}_1$, we define the event $E_{\mathrm{success}}$ as
\begin{align*}
            E_{\mathrm{success}}
    :=   &~ F_{\exp}(A_{11}, A_{12}, A_{13}) > 0 \\
    \land&~ F_{\lin}(A_{11}, A_{12}, A_{13}) = 0 \\
    \land&~ F_{\exp}(A_{01}, A_{02}, A_{03}) = 0 \\
    \land&~ F_{\lin}(A_{01}, A_{02}, A_{03}) = 0. 
\end{align*}

We first fix $X = \mathrm{vec}(I_d)$, and deploy the numerical experiments for parameters $n$, $d$, $m$, $a_0$ and $a_1$. To be specific, 
\begin{itemize}
    \item We deploy experiments for $n \in [200, 1000]$ with a step size of $10$. For each $n$, we set $d = 11$, $\delta = 0.01$, $m=\log (n/\delta)$, $a_0 = 0.01$, $a_1 = 3$. For each set of parameters, we iteratively generated models 100 times and recorded the occurrences of successful events denoted as $E_{\mathrm{success}}$. The result can be found in Figure (a) in Figure~\ref{fig:cross}. 
    
    \item We deploy experiments for $d \in \{2, 3, 5, 6, 9, 11, 21, 26, 41, 51\}$. For each $d$, we set $n = 200$, $\delta = 0.01$, $m=\log (n/\delta)$, $a_0 = 0.01$, $a_1 = 3$. For each set of parameters, we iteratively generated models 100 times and recorded the occurrences of successful events denoted as $E_{\mathrm{success}}$. The result can be found in Figure (b) in Figure~\ref{fig:cross}. 
    
    \item We deploy experiments for $m \in [1, 80]$ with a step size of $1$. For each $m$, we set $n = 200$, $d = 11$, $\delta = 0.01$, $m=\log (n/\delta)$, $a_0 = 0.01$, $a_1 = 3$. For each set of parameters, we iteratively generated models 100 times and recorded the occurrences of successful events denoted as $E_{\mathrm{success}}$. The result can be found in Figure (c) in Figure~\ref{fig:cross}. 
    
    \item We deploy experiments for $a_1 \in [0.1, 3]$ with a step size of $0.036$. For each $a_1$, we set $n = 200$, $d=12$, $\delta = 0.01$, $m=\log (n/\delta)$, $a_0 = 0.01$. For each set of parameters, we iteratively generated models 100 times and recorded the occurrences of successful events denoted as $E_{\mathrm{success}}$. The result can be found in Figure (d) in Figure~\ref{fig:cross}. 
    
    \item We deploy experiments for $a_0 \in [0.01, 0.9]$ with a step size of $0.01$. For each $a_0$, we set $n = 200$, $d=12$, $\delta = 0.01$, $m=\log (n/\delta)$, $a_0 = 0.01$. For each set of parameters, we iteratively generated models 100 times and recorded the occurrences of successful events denoted as $E_{\mathrm{success}}$. The result can be found in Figure (e) in Figure~\ref{fig:cross}. 
\end{itemize}

\ifdefined\isarxiv
%\section*{Acknowledgments}
\bibliographystyle{alpha}
\bibliography{ref}
\else

\fi

%%%% Cut-line between first 10 pages and appendix

%%% some writing rules

%% Writing rule for creating tags.
%% Tags :
%% Theorem    \ref{thm:bla_bla}
%% Lemma      \ref{lem:bla_bla}
%% Claim      \ref{cla:bla_bla}
%% Corollary  \ref{cor:bla_bla}
%% Fact       \ref{fac:bla_bla}
%% Definition \ref{def:bla_bla}
%% Section    \ref{sec:bla_bla}
%% Subsection \ref{sub:bla_bla}
%% Equation   \ref{eq:bla_bla}

\end{document}